\documentclass[10pt]{article}

\let\articleaddcontentsline\addcontentsline
\usepackage[preprint]{tmlr}
\let\addcontentsline\articleaddcontentsline

\usepackage{amsmath,amsfonts,bm}

\def\1{\bm{1}}

\DeclareMathAlphabet{\mathsfit}{\encodingdefault}{\sfdefault}{m}{sl}
\SetMathAlphabet{\mathsfit}{bold}{\encodingdefault}{\sfdefault}{bx}{n}

\usepackage{amsmath}
\usepackage{amssymb}
\usepackage{amsthm}
\usepackage{booktabs}
\usepackage{longtable}
\usepackage{array}
\usepackage{float}
\usepackage{placeins}
\usepackage{graphicx}
\usepackage{wrapfig}
\usepackage{needspace}
\usepackage{microtype}
\usepackage{hyperref}
\usepackage{url}
\usepackage{doi}
\usepackage{csquotes}
\usepackage[capitalize]{cleveref}

\theoremstyle{definition}
\newtheorem{definition}{Definition}[section]
\newtheorem{question}{Question}
\crefname{question}{Question}{Questions}
\Crefname{question}{Question}{Questions}
\usepackage{enumitem}
\setlist{leftmargin=*}

  \crefname{appendix}{Appendix}{Appendices}
  \Crefname{appendix}{Appendix}{Appendices}

\makeatletter
\renewcommand{\paragraph}{%
  \@startsection{paragraph}{4}{\z@}%
  {0pt}%
  {-1em}%
  {\normalfont\normalsize\bfseries}%
}
\makeatother

\title{Do Higher-Order Models Win for Higher-Order Reasons? Rethinking Performance Gains in Hypergraph Learning%
\thanks{Generative AI tools assisted with feedback on software implementation and debugging. The authors reviewed and tested AI-assisted code. The authors take responsibility for the final content, claims, and artifacts.}}

\newcommand{\preprintauthor}[4]{%
  \begin{minipage}[t]{0.32\textwidth}\centering%
    \href{#3}{\textbf{#1}}\textsuperscript{#2}\\[-1pt]
    {\small\normalfont\href{mailto:#4}{\nolinkurl{#4}}}%
  \end{minipage}%
}
\author{%
  \begin{center}
  \begin{tabular}{@{}c@{\hspace{0.02\textwidth}}c@{\hspace{0.02\textwidth}}c@{}}
\preprintauthor{Fanchen Bu}{1}{https://bokveizen.github.io/}{boqvezen97@kaist.ac.kr} & \preprintauthor{Fan Li}{2}{https://coco-hut.github.io/}{fan.li8@unsw.edu.au} & \preprintauthor{Geon Lee}{1}{https://geonlee0325.github.io/}{geonlee0325@kaist.ac.kr} \\[1.8ex]\noalign{\vskip 1.2ex}
\preprintauthor{Sunwoo Kim}{1}{https://sites.google.com/view/sunwoo97}{kswoo97@kaist.ac.kr} & \preprintauthor{Xiaoyang Wang}{2}{https://research.unsw.edu.au/people/dr-xiaoyang-wang}{xiaoyang.wang1@unsw.edu.au} & \preprintauthor{Renaud Lambiotte}{3}{https://www.lambiotte.be/}{renaud.lambiotte@maths.ox.ac.uk} \\[1.8ex]\noalign{\vskip 1.2ex}
\multicolumn{3}{c}{\preprintauthor{Kijung Shin}{1}{https://kijungs.github.io/}{kijungs@kaist.ac.kr}} \\[1.2ex]
  \end{tabular}\\[1.2ex]
  {\small\normalfont\textsuperscript{1}KAIST\quad
   \textsuperscript{2}UNSW Sydney\quad
   \textsuperscript{3}University of Oxford}
  \end{center}
}
\hypersetup{
  hidelinks,
  pdftitle={Do Higher-Order Models Win for Higher-Order Reasons? Rethinking Performance Gains in Hypergraph Learning},
  pdfauthor={Fanchen Bu, Fan Li, Geon Lee, Sunwoo Kim, Xiaoyang Wang, Renaud Lambiotte, Kijung Shin}
}

\begin{document}

\addtocontents{toc}{\protect\setcounter{tocdepth}{-1}}

\maketitle

\begin{abstract}
Higher-order models (e.g., hypergraph neural networks) often outperform lower-order baselines on hypergraph learning benchmarks, and their advantages are commonly attributed to their ability to exploit higher-order information.
However, better performance alone does not establish this explanation. We therefore ask: \emph{Do higher-order models win for higher-order reasons?}
To investigate this question, we introduce a controlled performance-attribution framework that perturbs higher-order information while preserving the lower-order, i.e., pairwise, information.
Across 25 commonly used hypergraph learning benchmarks spanning three tasks, we frequently observe an intriguing pattern: higher-order models originally outperform lower-order baselines, yet retain most of their advantage after perturbation.
This suggests that much of the observed advantage remains achievable without the higher-order information.
We then investigate potential lower-order explanations for these remaining gaps.
We find that simple additions to a lower-order baseline, e.g., richer pairwise weighting, more steps of pairwise feature propagation, and normalization, reduce the remaining performance gaps, supporting lower-order explanations for part of the observed advantage.
Our analysis calls for the hypergraph learning community to rethink performance attribution by distinguishing performance gains from their explanations, adopt stronger lower-order baselines, and use suitable benchmarks that better test the value of higher-order information.
\end{abstract}

\section{Introduction}
\label{sec:introduction}
Hypergraphs generalize graphs by allowing a single hyperedge to connect more than two entities, making them a natural mathematical representation of higher-order relations~\citep{bick2023higher}.
Higher-order learning models, especially hypergraph neural networks (HNNs), are correspondingly designed to exploit hypergraph structure, commonly by propagating and aggregating information through hyperedges rather than only between pairs~\citep{gao2022hypergraph,kim2024survey}.

Empirically, higher-order models often outperform lower-order alternatives, such as graph neural networks (GNNs), on hypergraph learning tasks~\citep{li2026dhgbench}, and their advantage is commonly interpreted as evidence that these models benefit from their capacity to exploit higher-order relations~\citep{feng2019hypergraph,bai2021hypergraph,xu2023hypehr,li2026dhgbench}.\footnote{See \cref{app:motivating-interpretations} for the detailed statements from the cited literature.}
This interpretation seems natural, but conventional empirical comparisons do not provide sufficient evidence that the observed advantage is explained by higher-order information, i.e., the information about group relations beyond what the pairwise representation preserves (see \cref{def:information-order}).
Attributing their advantage specifically to such higher-order information requires separate evidence.
We therefore ask: \emph{Do higher-order models actually win for higher-order reasons?}
In other words, does their advantage arise from exploiting information about group relations beyond what the pairwise representation preserves?

To answer this question, we propose a \emph{controlled intervention} on the data representation.
For each hypergraph, we construct a perturbed counterpart that preserves its lower-order information while perturbing its higher-order information.
We then evaluate higher-order models and lower-order baselines on both the original and perturbed views.\footnote{We intentionally keep it abstract here and defer the specific definitions of lower- and higher-order information considered in this work to~\cref{sec:problem_formulation}, and the construction of the perturbed view to~\cref{sec:framework}.}
With that intervention, our \emph{key intuition} is simple: if the higher-order information is indeed a major source of a higher-order model's performance advantage over lower-order baselines, then the performance advantage should shrink substantially under the perturbed view where such higher-order information is disturbed by the intervention.

\begin{figure}[t]
  \centering
  \includegraphics[width=\textwidth]{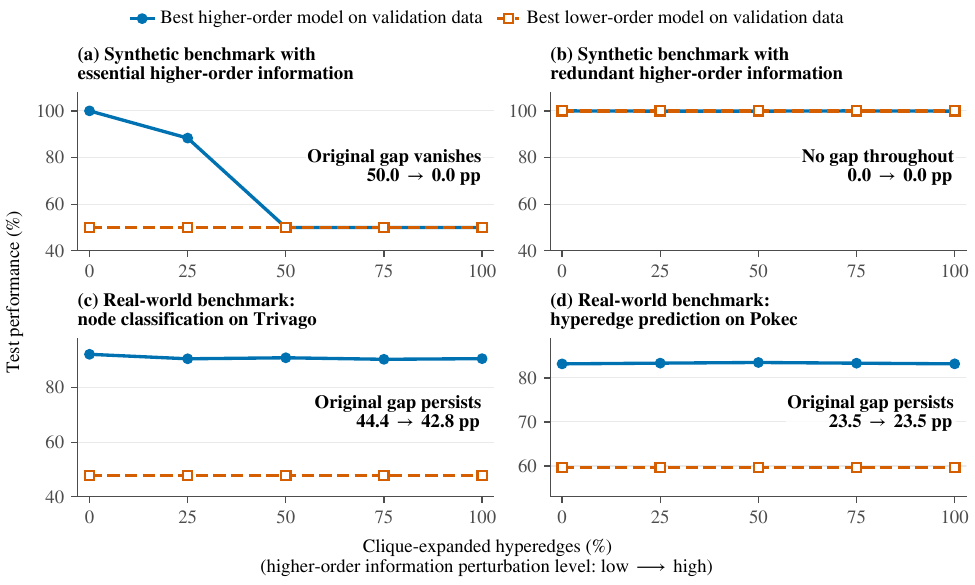}
  \vspace{-8mm}
  \caption{\textbf{Large performance advantages of higher-order models can survive perturbation of higher-order information.}  
  Synthetic benchmarks show (a) a vanishing advantage of the higher-order model when
  higher-order information is essential and (b) equally good performance by higher- and lower-order
  models throughout the intervention when higher-order information is redundant.
  In contrast, (c--d) two real-world benchmarks exhibit a distinct pattern:
  higher-order models retain most of their advantages despite the perturbation.
  See \cref{sec:framework} for details of the perturbation and
  \cref{sec:empirical_analysis} for the synthetic benchmarks and real-world results.}
  \vspace{-5mm}
  \label{fig:opening-attribution}
\end{figure}

To concretize and validate our intuition, we first construct two synthetic benchmarks in which the role of higher-order
information is known by construction.\footnote{In this work, we use \emph{benchmark} to mean a dataset--task pair; see \cref{sec:framework:prelim} for more details. \cref{sec:synthetic} presents those two synthetic benchmarks and \cref{app:synthetic} gives their full construction.}
As shown in \cref{fig:opening-attribution}(a), 
when the higher-order information is essential, a capable higher-order model
achieves perfect performance while the lower-order baseline remains at chance; once the higher-order information is perturbed, the higher-order model also falls to chance, and its entire advantage vanishes.
When the higher-order information is redundant, 
both higher-order and lower-order models can achieve perfect performance before and after the
intervention (\cref{fig:opening-attribution}(b)).

However, when we investigate commonly used real-world hypergraph learning benchmarks, we observe that many of them exhibit a third, more intriguing pattern.
The higher-order model wins on the original hypergraph, yet its advantage largely survives after the candidate higher-order information is perturbed, as illustrated by the two representative examples in
\cref{fig:opening-attribution}(c) and \cref{fig:opening-attribution}(d).
Across the benchmarks we study, higher-order models often retain most
of their performance advantage even after higher-order information is perturbed.
This suggests that much of their advantage can be achieved without the higher-order information. See \cref{sec:real_world} for detailed results.

What, then, sustains these remaining advantages?
We investigate whether some lower-order reasons, e.g., differences in how models process the preserved lower-order information, can potentially account for them.
Our case studies show that several retained gaps can be substantially
reduced through richer pairwise weighting, more steps of pairwise feature propagation, and normalization.
Motivated by these findings, we test the same simple additions to a lower-order baseline across benchmarks, without further hyperparameter tuning.
We observe that such additions are overall helpful, supporting lower-order explanations for part of the observed advantage. See \cref{sec:remaining_gaps} for more details.

In summary, our contributions are threefold:
\begin{itemize}[nosep]
    \item \textbf{Performance-attribution question and empirical evidence.}
    We try to distinguish higher-order models' performance advantages from evidence
    that higher-order information explains those advantages, and formalize
    this distinction as a performance-attribution problem
    (\cref{sec:problem_formulation}).
    We challenge the common interpretation that higher-order models win
    because they exploit higher-order information: across 25 commonly used
    hypergraph learning benchmarks spanning three tasks, many advantages
    largely remain after higher-order information is perturbed
    (\cref{sec:real_world}).

    \item \textbf{A principled evaluation protocol with controlled perturbation.}
    We introduce a controlled performance-attribution framework that perturbs higher-order information while preserving lower-order information, enabling us to measure how much advantage disappears or survives after such perturbation (\cref{sec:framework}).
    Controlled synthetic cases illustrate the expected responses when higher-order information is essential or redundant by construction (\cref{sec:synthetic}).

    \item \textbf{Investigation of source of performance gain beyond higher-order information.}
    We investigate whether lower-order designs and mechanisms can potentially explain some of the observed performance advantages of higher-order models. Guided by these findings, we show that
    simple additions, e.g., richer pairwise weighting, more steps of pairwise feature propagation, and normalization, can strengthen a lower-order baseline (\cref{sec:remaining_gaps}).
\end{itemize}
Together, our findings highlight the need to distinguish a model's
performance advantage from the information that explains it.
Our findings highlight the need for evaluations that test whether an advantage survives a controlled change to the claimed information source, investigate what sustains the remaining gap, and assess whether the benchmark tests the capability it is intended to measure (\cref{sec:benchmark_audit}).

\paragraph{Reproducibility.} The code and datasets are available online~\citep{ho_perf_attribution}.\footnote{\url{https://anonymous.4open.science/r/HO-perf-attribution/}}

\section{Related Work}
\label{sec:related_work}
\paragraph{Hypergraph learning and benchmark evidence.}
Hypergraph learning encompasses spectral methods, incidence-based neural
propagation, graph reductions, and learnable multiset functions
\citep{zhou2006learning,feng2019hypergraph,yadati2019hypergcn,chien2022allset}.
See \citet{gao2022hypergraph,kim2024survey} for broader surveys.
Recent benchmarks compare HNNs with pairwise-graph and feature-only baselines
across multiple learning tasks \citep{li2026dhgbench}.
Such comparisons establish predictive differences between models, but do not
by themselves identify the information responsible for those differences.

\paragraph{Predictive utility of higher-order information.}
Prior work studies the predictive value of higher-order information by varying
the available interaction order: \citet{yoon2020higher} progressively add
higher-order projections, while \citet{bian2025stepwise} progressively reduce
higher-order interactions.
We instead study contemporary HNNs and ask an attribution question: when an HNN
outperforms a lower-order model, does that advantage actually depend on
higher-order information?

\paragraph{Projection fidelity and higher-order reducibility.}
Prior work studies information loss, reconstruction ambiguity, and
non-uniqueness when hypergraphs are reduced to pairwise representations
\citep{kang2024low,wang2024graphs,larock2025nonuniqueness,lee2025marioh}.
Related work asks when higher-order mechanisms or dynamics are reducible to
pairwise ones
\citep{rosas2022mechanisms,lucas2026reducibility,llabres2026reducibility}.
These works concern structural recovery or dynamical reducibility; our focus is
the predictive role of higher-order information in HNN performance.

\section{The Performance-Attribution Question}
\label{sec:problem_formulation}
In this section, we formalize the central problem studied in this work.

\subsection{Preliminaries}
\label{sec:framework:prelim}

A \emph{hypergraph} is a tuple
$\mathcal H=(\mathcal V,\mathcal E,\mathbf w,\mathbf X)$, where $\mathcal V$
is the node set, $\mathcal E\subseteq\{e\subseteq\mathcal V:|e|\ge2\}$
is a set of unique hyperedges, $\mathbf w\colon\mathcal E\to
\mathbb N_{>0}$ assigns each hyperedge a weight representing its
multiplicity, and $\mathbf X$ denotes the node features (if any).\footnote{For notational simplicity, $\mathcal H$ denotes the
non-singleton part of the input hypergraph. Singleton hyperedges,
when present in the data, are retained unchanged and treated as
fixed background input.}
For a hyperedge $e$, let $\operatorname{clique}(e)=\bigl\{\{u,v\}:u,v\in e,\ u\ne v\bigr\}$
denote the set of unordered node pairs induced by $e$.
The \emph{weighted clique expansion} of $\mathcal H$ is the graph
$\operatorname{WCE}(\mathcal H)=(\mathcal V,\mathcal E_2,
\mathbf w_{\mathcal H},\mathbf X)$,
where $\mathcal E_2=\bigcup_{e\in\mathcal E}\operatorname{clique}(e)$ is a set of unique pairwise edges.
Their weights are computed by $\mathbf w_{\mathcal H}(u,v)=\sum_{e\in\mathcal E}  \mathbf w(e)\,\mathbf 1\!\left[\{u,v\}\subseteq e\right]$.

Clique expansion is a standard way to reduce a hypergraph to a
graph~\citep{bick2023higher,yadati2019hypergcn,gao2022hypergraph}.
However, this reduction is generally lossy: distinct hypergraphs can induce
the same weighted clique expansion, so the original hyperedges cannot,
in general, be uniquely reconstructed from the expanded
graph~\citep{wang2024graphs,larock2025nonuniqueness}.

In this paper, a \emph{dataset} $\mathcal D$ means the underlying hypergraph
data: one hypergraph for node classification or hyperedge prediction, and a
collection of hypergraphs for hypergraph classification.  A \emph{task}
$\tau$ specifies the prediction problem, and a \emph{benchmark}
$\mathcal B=(\mathcal D,\tau)$ is a dataset paired with a task.  Thus, using
the same dataset for different tasks defines different benchmarks.

\subsection{Higher- and lower-order information and models}
\label{sec:framework:higher_lower_order}

\begin{definition}[Higher- and lower-order information]
\label{def:information-order}
The \emph{lower-order information} carried by a hypergraph $\mathcal H$ is
its weighted clique expansion $\operatorname{WCE}(\mathcal H)$.
Two hypergraphs $\mathcal H_1,\mathcal H_2$ are
\emph{lower-order equivalent}, written
$\mathcal H_1\equiv_{\mathrm{LO}}\mathcal H_2$, if
$\operatorname{WCE}(\mathcal H_1)
  =\operatorname{WCE}(\mathcal H_2)$.
The \emph{higher-order information} carried by $\mathcal H$ is the information that distinguishes it from other hypergraphs that are lower-order equivalent to it.
\end{definition}

The projection weights count pair co-occurrences.
The choice is motivated by incidence-based HNNs, in which a pair can exchange information through each instance of a shared hyperedge~\citep{feng2019hypergraph}.
Our attribution statements are conditional on this choice of lower-order information. 
\cref{sec:limitations} discusses alternative interventions that preserve
the same lower-order information.

\begin{definition}[Higher- and lower-order models]
\label{def:model-order}
A parameterized model (e.g., HNN or GNN)
$M=\{M_\theta\}_{\theta\in\Theta}$ is \emph{higher-order} if at least one of
its parameterized functions depends on higher-order information, i.e.,
$
  \exists\,\theta\in\Theta,\ \exists\,\mathcal H_1,\mathcal H_2:
  \quad \mathcal H_1\equiv_{\mathrm{LO}}\mathcal H_2
  \quad\text{and}\quad
  M_\theta(\mathcal H_1)\ne M_\theta(\mathcal H_2)$.
Conversely, $M$ is \emph{lower-order} if every parameterized function is determined by lower-order information, i.e.,
$
  \forall\,\theta\in\Theta,\ \forall\,\mathcal H_1,\mathcal H_2:
  \quad \mathcal H_1\equiv_{\mathrm{LO}}\mathcal H_2
  \quad\Longrightarrow\quad
  M_\theta(\mathcal H_1)=M_\theta(\mathcal H_2)$.
\end{definition}

This definition concerns information access rather than architecture.
For example, graph neural networks augmented with hypergraph-derived
encodings can exploit higher-order information
\citep{pellegrin2025higher}; a graph architecture therefore does not
automatically qualify as a lower-order baseline.

This definition of a higher-order model family is deliberately permissive: it
describes the family's \emph{capacity} to use information beyond the weighted
clique expansion, while a particular hyperparameter configuration or fitted instance may use only lower-order information.

\subsection{The performance-attribution question}
\label{sec:framework:question}

The preceding definitions distinguish a model family's \emph{capacity} to use
higher-order information from whether that information explains its empirical
performance.  A higher-order model is capable of distinguishing
hypergraphs with the same weighted clique expansion, yet its advantage on a particular
benchmark may instead arise from other differences between the models.

\begin{question}[Performance attribution]
\label{question:performance-attribution}
Given a benchmark $\mathcal B=(\mathcal D,\tau)$, a higher-order model $M_H$,
and a lower-order model $M_L$, suppose that $M_H$ outperforms $M_L$ on
$\mathcal B$.  To what extent is this performance advantage attributable to
the higher-order information in the benchmark?
\end{question}

\section{A Controlled Performance-Attribution Framework}
\label{sec:framework}
We aim to answer the performance-attribution question (\cref{question:performance-attribution})
with a controlled intervention that changes higher-order information while preserving the lower-order information.

\paragraph{Original and perturbed views.}
For a hypergraph $\mathcal H$, we call
$\mathcal H^{(0)}=\mathcal H$ and
$\mathcal H^{(1)}=\operatorname{WCE}(\mathcal H)$ the \emph{original view}
and the (fully) \emph{perturbed view}, respectively.  The intervention
preserves the lower-order information exactly because
$\operatorname{WCE}(\mathcal H^{(0)})
  =\operatorname{WCE}(\mathcal H^{(1)})
  =\mathcal H^{(1)}$,  
while replacing every non-pairwise hyperedge by its weighted pairwise
expansion. 
Given a benchmark $\mathcal B=(\mathcal D,\tau)$, 
applying this transformation to the hypergraph(s) in
$\mathcal D$ yields two benchmark views
$\mathcal B^{(r)}=(\mathcal D^{(r)},\tau)$ for $r\in\{0,1\}$; for hypergraph
classification, it is applied to each hypergraph in the dataset.
The task $\tau$ is unchanged.

\paragraph{Key intuition.}
Our key intuition is that, if higher-order information is a major
source of the higher-order model $M_H$'s advantage on the original
benchmark $\mathcal B=\mathcal B^{(0)}$, that advantage should
shrink substantially on the fully perturbed benchmark
$\mathcal B^{(1)}$.
Here, the intervention replaces higher-order groupings with
weighted pairwise connections while preserving lower-order
information, allowing us to test how much of the original
advantage depends on those groupings.

\paragraph{Performance attribution.}
Let $P^{(r)}(M;\mathcal B)$ denote the performance of model $M$ on $\mathcal B^{(r)}$, where higher values indicate better performance, 
and $\mathcal B$ can be omitted when the context is clear.
By \cref{def:model-order}, the lower-order model $M_L$ receives the same information before and after the intervention.
Hence $P^{(0)}(M_L) = P^{(1)}(M_L)$; we denote this common value by $P(M_L)$.\footnote{See \cref{app:baseline-invariance} for the empirical validation of this invariance.}
Then we define the \emph{performance gaps} on the original and perturbed views as
$\Delta^{(r)}(M_H;M_L)=P^{(r)}(M_H)-P(M_L)$ for $r = 0$ and $r = 1$ respectively,
where $M_H$ and $M_L$ can be omitted when the context is clear.
Then the original advantage of the higher-order model can be decomposed as
\begin{equation}
  \Delta^{(0)}
  =
  \underbrace{\bigl(\Delta^{(0)}-\Delta^{(1)}\bigr)}
              _{\text{gap removed by the intervention; denoted as }
                \Delta_{\mathrm{attr}}}
  +
  \underbrace{\Delta^{(1)}}_{\text{gap surviving the intervention}}.
  \label{eq:gap-decomposition}
\end{equation}
A positive $\Delta_{\mathrm{attr}} := \Delta^{(0)}-\Delta^{(1)}$ indicates that the advantage is sensitive to the intervention.
A positive $\Delta^{(1)}$ shows that an advantage remains achievable without the perturbed higher-order information.

\paragraph{Partial perturbations as additional diagnostics.}
To diagnose how a model's response develops between the two endpoints $\mathcal B^{(0)}$ and $\mathcal B^{(1)}$, we additionally consider \emph{partially perturbed views} at finer-grained levels $r\in(0,1)$.
At level $r$, a proportion $r$ of the hyperedges of size larger than $2$ is expanded into pairwise ones, while the remaining hyperedges are left unchanged. 
Unlike the two endpoints $\mathcal B^{(0)}$ and $\mathcal B^{(1)}$, an intermediate level does not specify a unique view because many subsets can realize the same proportion.
We therefore sample multiple partial views at each level and average their
test performance.
This extends the notations $P^{(r)}$ and $\Delta^{(r)}$ above to a continuous spectrum $r\in[0,1]$.
More details of the construction and sampling procedure for partial perturbations are given in \cref{app:partial-perturbation-construction}.

\section{Empirical Analysis on Real-World Benchmarks}
\label{sec:empirical_analysis}
We now use the framework in \cref{sec:framework} to study \cref{question:performance-attribution} empirically.

\subsection{Experimental setup}
\label{sec:empirical_analysis:setup}

\paragraph{Tasks.} We consider three major hypergraph-learning
tasks: node classification, hyperedge prediction, and hypergraph
classification.

\paragraph{Datasets.} For node classification, we use Actor, Cora, Cora-CA,
DBLP-CA, Pokec, Pubmed, Gamers, ModelNet40, NTU2012, Ratings, Trivago,
Walmart, and Yelp.
For hyperedge prediction, we use Actor, Cora, Cora-CA, DBLP-CA, Pokec, and Pubmed.
For hypergraph classification, we use IMDB-Dir-Form, IMDB-Dir-Genre, RHG-10,
RHG-3, Steam-Player, and Twitter-Friend.
This gives 25 real-world benchmarks (i.e., dataset--task pairs) in total.\footnote{{More precisely, RHG-3 and RHG-10 are not completely \emph{real-world} datasets: they are collections of structural hypergraph classes, but they are commonly used hypergraph-learning benchmarks.}}

\paragraph{Models.} We evaluate 17
representative HNN models:
HGNN~\citep{feng2019hypergraph},
HyperGCN~\citep{yadati2019hypergcn},
HCHA~\citep{bai2021hypergraph},
LEGCN~\citep{yang2022legcn},
HyperND~\citep{prokopchik2022hypernd},
PhenomNN~\citep{wang2023phenomnn},
SheafHyperGNN~\citep{duta2023sheafhypergnn},
HJRL~\citep{yan2024hjrl},
DPHGNN~\citep{saxena2024dphgnn},
TF-HNN~\citep{tang2025tfhnn},
HNHN~\citep{dong2020hnhn},
UniGCNII~\citep{huang2021unignn},
AllSetTransformer~\citep{chien2022allset},
ED-HNN~\citep{wang2023edhnn},
HyperGT~\citep{liu2024hypergt},
EHNN~\citep{kim2022ehnn}, and
T-HyperGNN~\citep{wang2025thypergnn}.
We also evaluate a multilayer perceptron (MLP) using only node features, and
the clique-expansion graph convolutional and attention models
(CEGCN and CEGAT)~\citep{chien2022allset}.
By \cref{def:model-order}, the 17 HNN
models are higher-order, whereas MLP, CEGCN, and CEGAT are lower-order.

\paragraph{Model selection across perturbation levels.}
Unless otherwise stated, on each benchmark we independently select the
higher-order model and the lower-order model with the highest mean validation
performance on the original view $\mathcal B^{(0)}$.  
We use the same selected model configurations and hyperparameters
at every perturbation level, training each model separately
on each view.
See \cref{app:real-world-results} for results on other models.

\paragraph{Evaluation metrics.} Node classification is evaluated by accuracy, hyperedge prediction by average precision (AP) and area under the receiver operating
characteristic curve (AUROC), and hypergraph classification by both accuracy and macro-averaged F1 (Macro-F1).
Unless stated otherwise, the main text reports accuracy for node and hypergraph classification and AP for hyperedge prediction, with secondary-metric results provided in the appendix (see, e.g., \cref{app:secondary-gaps,app:edge_pred,app:hg_cls,app:cegcn-additions}).

\paragraph{Additional details.}
\cref{app:experimental_setup} provides more details of 
each dataset and the data preprocessing,
the models and their implementations, 
and the hyperparameter settings.

\subsection{Synthetic calibration: Two controlled cases}
\label{sec:synthetic}

We first use two controlled synthetic cases as concrete instances to validate
our key intuition in \cref{sec:framework}.  

\paragraph{Construction.}
We consider hypergraph classification using two controlled datasets.
We provide a high-level description of the construction below; see
\cref{app:synthetic} for the details.
We use two motif templates $\mathcal T_0$ and $\mathcal T_1$,
each specifying a small hypergraph pattern on six nodes.
The two patterns differ in their hyperedges, but have identical weighted clique expansions
(\cref{def:information-order}).
Each hypergraph is a disjoint union of 21 motif instances drawn
from these two templates.
The label indicates which template occurs more often, and the two
classes are balanced.
In the case with \emph{essential} higher-order information, we provide
no additional signal. All hypergraphs therefore have the same
lower-order representation. Lower-order models cannot perform beyond
chance, whereas a capable higher-order model can distinguish the templates.
In the case with \emph{redundant} higher-order information, we add a
two-node hyperedge specific to each template to each motif instance.
These hyperedges distinguish the templates in their weighted clique
expansions, making the label recoverable from lower-order information alone.

\paragraph{Expected patterns.}
In the essential case, the higher-order model should initially outperform the lower-order model, but this advantage should shrink as higher-order information is perturbed and vanish at the fully perturbed endpoint.
Using the notation in \cref{sec:framework},
we therefore expect a substantially positive $\Delta^{(0)}$,
decreasing $\Delta^{(r)}$ as $r$ increases, and eventually
$\Delta^{(1)}\simeq0$ with
$\Delta_{\mathrm{attr}}\simeq\Delta^{(0)}$.
In the redundant case, both model families can recover the label
from the preserved pairwise information, so neither should have
a performance advantage before or after perturbation; we expect $\Delta^{(r)}\simeq0$ for every $r\in[0,1]$.

\paragraph{Results.}
In the essential case (\cref{fig:opening-attribution}(a)), the selected
higher-order model achieves perfect test accuracy on the original view while
the selected lower-order model remains at the $50\%$ chance level.
Once the higher-order information is fully perturbed, the higher-order model also falls
to chance and its entire $50$-point advantage vanishes, yielding $\Delta^{(0)}=\Delta_{\mathrm{attr}}=50$ points and $\Delta^{(1)}=0$.
In the redundant case (\cref{fig:opening-attribution}(b)), both selected models maintain
perfect performance regardless of the intervention, so $\Delta^{(r)}\simeq0$ throughout.
These two cases match our expectation and validate the intuition underlying our performance-attribution framework.

\paragraph{Additional details.}
\cref{app:synthetic} provides more details of 
the construction of the two controlled cases and
the evaluation procedure.

\subsection{Real-world results: Advantages largely survive perturbation}
\label{sec:real_world}

\begin{figure}[!htbp]
  \centering
  \includegraphics[width=0.60\textwidth]{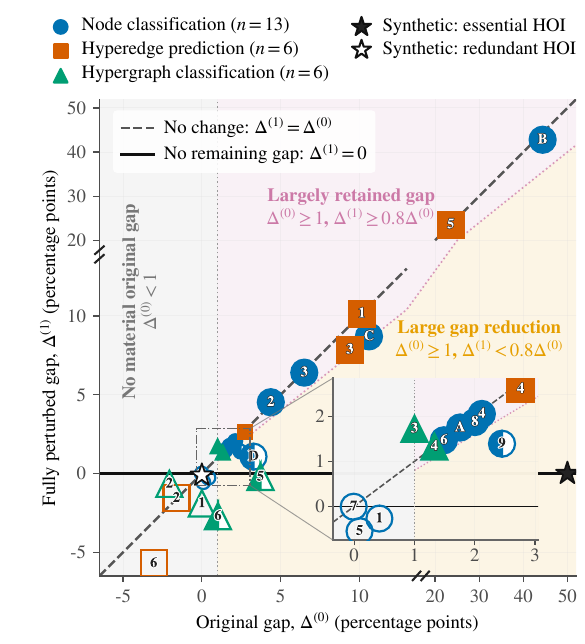}
  \par\smallskip
  \begingroup
  \definecolor{figtasknode}{HTML}{0072B2}
  \definecolor{figtaskedge}{HTML}{D55E00}
  \definecolor{figtaskhypergraph}{HTML}{009E73}
  \newcommand{\nodekey}[1]{\textcolor{figtasknode}{\(\bullet\,#1\)}}
  \newcommand{\edgekey}[1]{\textcolor{figtaskedge}{\(\blacksquare\,#1\)}}
  \newcommand{\hypergraphkey}[1]{\textcolor{figtaskhypergraph}{\(\blacktriangle\,#1\)}}
  \fontsize{7}{8}\selectfont
  \setlength{\tabcolsep}{2pt}
  \renewcommand{\arraystretch}{0.94}  
  \begin{tabular}{@{}llll@{}}
    \multicolumn{2}{@{}l}{\textbf{\textcolor{figtasknode}{Node cls.}}}
      & \textbf{\textcolor{figtaskedge}{Hyperedge pred.}}
      & \textbf{\textcolor{figtaskhypergraph}{Hypergraph cls.}} \\
    \nodekey{1}\ Actor   & \nodekey{8}\ ModelNet40 & \edgekey{1}\ Actor   & \hypergraphkey{1}\ IMDB-Dir-Form \\
    \nodekey{2}\ Cora    & \nodekey{9}\ NTU2012    & \edgekey{2}\ Cora    & \hypergraphkey{2}\ IMDB-Dir-Genre \\
    \nodekey{3}\ Cora-CA & \nodekey{A}\ Ratings    & \edgekey{3}\ Cora-CA & \hypergraphkey{3}\ RHG-10 \\
    \nodekey{4}\ DBLP-CA & \nodekey{B}\ Trivago    & \edgekey{4}\ DBLP-CA & \hypergraphkey{4}\ RHG-3 \\
    \nodekey{5}\ Pokec   & \nodekey{C}\ Walmart    & \edgekey{5}\ Pokec   & \hypergraphkey{5}\ Steam-Player \\
    \nodekey{6}\ Pubmed  & \nodekey{D}\ Yelp$^*$   & \edgekey{6}\ Pubmed  & \hypergraphkey{6}\ Twitter-Friend \\
    \nodekey{7}\ Gamers  &                        &                     &
  \end{tabular}\par
  \endgroup
  \caption{{\textbf{Most performance advantages of higher-order models largely persist even after the perturbation of higher-order information.} Original versus fully perturbed performance gaps between higher-
  and lower-order models across 25 benchmarks. Of the 18 with a material
  original gap of at least one point, 14 retain at least 80\% after perturbation.
  For benchmark markers, filled and half-filled symbols indicate
  that at least 80\% or less than 80\% of a material original gap
  remains, respectively; hollow symbols indicate an original gap
  below one percentage point.
  Stars denote synthetic cases with essential or redundant higher-order information (HOI).}}
  \label{fig:real-gap-map}
\end{figure}

We have seen the effect of the intervention in settings where the role of
higher-order information is known by construction.
Now, we turn to real-world benchmarks.
On each real-world benchmark, which of the two controlled patterns appears?
The answer is often \emph{neither}.  Across the 25 commonly used hypergraph learning benchmarks (\cref{sec:empirical_analysis:setup}), we frequently observe a third pattern:
The higher-order model initially outperforms the lower-order
baseline by at least one percentage point ($\Delta^{(0)}\ge1$),
yet most of this advantage remains on the fully perturbed view.
The remaining gap $\Delta^{(1)}$ is therefore close to the original
gap $\Delta^{(0)}$, so the portion removed by the intervention,
$\Delta_{\mathrm{attr}}$, is small relative to the original advantage.
We describe this as a \emph{largely retained gap}.

\paragraph{Representative largely retained gaps.}
\cref{fig:opening-attribution}(c) and \cref{fig:opening-attribution}(d) show the two selected real-world comparisons with the
largest original gaps $\Delta^{(0)}$.
For node classification on Trivago (\cref{fig:opening-attribution}(c)), 
the selected higher-order model's original advantage $\Delta^{(0)}=44.4$ decreases only to
$\Delta^{(1)}=42.8$ accuracy points, giving $\Delta_{\mathrm{attr}}=1.6$ points
(3.6\% of $\Delta^{(0)}$).
For hyperedge prediction on Pokec (\cref{fig:opening-attribution}(d)), 
the advantage remains essentially unchanged at $23.5$ AP points, so
$\Delta^{(0)}\simeq\Delta^{(1)}$ and $\Delta_{\mathrm{attr}}\simeq0$.
Thus, even for the two benchmarks with the largest original advantages, almost all of $\Delta^{(0)}$ remains achievable after perturbing higher-order information.

\paragraph{Benchmark-wide analysis.}
In \cref{fig:real-gap-map}, for each benchmark, we show the performance gaps
on the original and perturbed views. 
Blue circles, orange squares, and green triangles denote node classification, hyperedge prediction, and hypergraph
classification, respectively.
Within each task, the identifiers follow the dataset order in \cref{sec:empirical_analysis:setup}; the six datasets shared
by node classification and hyperedge prediction have matching identifiers.
For Yelp,  the higher-order model with the best validation performance
exceeds the available memory, so we use the highest-ranked model by
validation performance that fits within the memory budget.
The stars show the essential and redundant synthetic references from \cref{fig:opening-attribution}. 

We organize the endpoint comparisons into three categories.
Benchmarks with
$\Delta^{(0)}<1$ point have \emph{no material original gap}.
Among those with $\Delta^{(0)}\geq1$, we call the gap \emph{largely retained} when at least 80\% of the original advantage remains, $\Delta^{(1)}\geq0.8\Delta^{(0)}$, and call it \emph{large gap reduction} otherwise.
Of the 25 benchmarks, seven have no material original gap.
On these benchmarks, the original comparison provides no material
advantage of the higher-order model to attribute in the first place.

\emph{Observation 1: Most performance advantages of higher-order
models over lower-order baselines largely persist even after
higher-order information is perturbed.}
Among the remaining 18 benchmarks, 14 exhibit largely retained gaps.
A majority of these advantages remains largely intact even under
stricter criteria: 12 of 18 retain at least 90\% of their original
gap, and 10 of 18 retain at least 95\%.
Median retention among these 18 comparisons is 96.7\%.
The retained advantages include both large and smaller original gaps.
For the two largest gaps, node classification on Trivago and
hyperedge prediction on Pokec, almost the entire advantage remains
after perturbation, as illustrated in
\cref{fig:opening-attribution}(c) and \cref{fig:opening-attribution}(d).
Thus, even a large performance improvement over a lower-order
baseline need not provide strong evidence that higher-order
information explains that improvement.
The broader pattern in \cref{fig:real-gap-map} shows that this
mismatch is not confined to these two examples:
most material advantages remain largely achievable using the
preserved pairwise representation.

\emph{Observation 2: Largely retained gaps occur across tasks.}
Among original advantages of at least one percentage point, largely
retained gaps occur for 8 of 10 node-classification benchmarks, all four
hyperedge-prediction benchmarks, and 2 of 4 hypergraph-classification
benchmarks.
The pattern therefore extends across prediction at the node,
hyperedge, and hypergraph levels, although it is not equally
prevalent in every task.
For node classification and hyperedge prediction, most or all
material advantages survive largely intact.
Hypergraph classification presents a more mixed picture:
the advantages on RHG-10 and RHG-3 largely remain, whereas those
on Steam-Player and Twitter-Friend disappear after perturbation
(\cref{app:gap-retention}).
These contrasting outcomes show why attribution must be evaluated
for each benchmark: performance gains within the same task can
differ in their dependence on higher-order information.
Large gap reductions also need not eliminate the original advantage.
For node classification on NTU2012 and Yelp, the gaps decrease
from $2.45$ to $1.39$ and from $3.28$ to $1.07$ accuracy points,
respectively (\cref{app:gap-retention}).
These cases illustrate that the intervention can remove a substantial
part of an advantage while leaving another part to be explained.

\emph{Overall: Performance attribution changes the conclusion suggested by the original gap.}
Based only on the original performance gap $\Delta^{(0)}$, one might interpret a large advantage of the higher-order model as evidence that higher-order information is highly useful. Our framework separates that model-level advantage from its attribution to higher-order information.

\paragraph{Additional results.}
\cref{app:real-world-results} reports additional results.
\cref{app:gap-retention} reports detailed benchmark-wise results.
\cref{app:secondary-gaps} reports secondary-metric results using AUROC for hyperedge prediction and Macro-F1 for hypergraph classification.
\cref{app:node_cls,app:edge_pred,app:hg_cls} extend the analysis beyond the single selected model pair, reporting results on other models.

\paragraph{Remark.}
A largely retained gap does not imply that the higher-order model is unnecessary.
It supports the narrower conclusion that most of its advantage over the selected lower-order comparator remains achievable without the perturbed higher-order information.

\subsection{Investigating lower-order explanations for retained gaps}
\label{sec:remaining_gaps}

\begin{figure}[!t]
  \centering  
  \includegraphics[width=\textwidth]{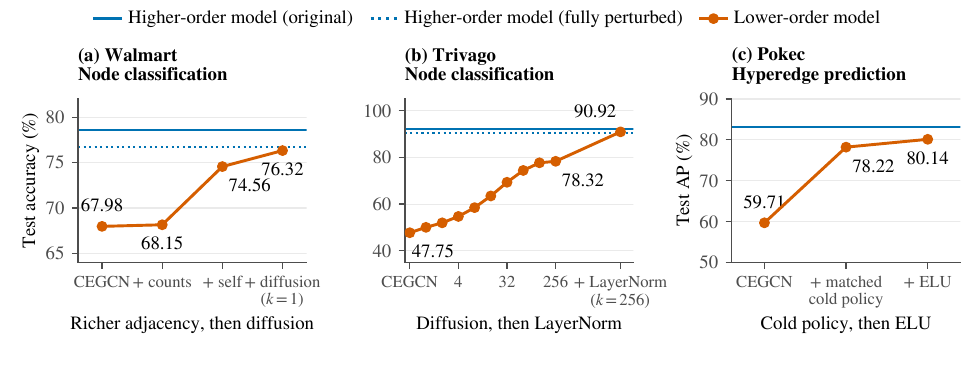}
  \vspace{-15mm}
  \caption{\textbf{Lower-order mechanisms such as stronger pairwise processing and matched model choices can substantially narrow the performance advantages that higher-order models retain after perturbation.}
  Representative case studies examine the effect of
  (a) incorporating pair multiplicities and structural self weights, then
  propagating input features through pairwise connections;
  (b) increasing the number of feature-propagation steps and normalizing
  hidden features with LayerNorm; and
  (c) matching HGNN by removing feature-dependent messages from nodes
  absent from the training structure and using ELU between learned layers. 
  }
  \label{fig:remaining-gap-case-studies}
  \vspace{-5mm}
\end{figure}

If most of a higher-order model's advantage remains even after perturbing the higher-order information, what sustains it?
In this section, we investigate potential lower-order explanations
in the benchmarks with largely retained gaps shown in
\cref{fig:real-gap-map}.
We first examine three representative cases in
\cref{fig:remaining-gap-case-studies} to investigate whether pairwise
processing, feature propagation, and normalization can explain the
remaining advantages. For hyperedge prediction, we also examine how
models handle nodes absent from the training structure.
Motivated by these case studies, we then augment CEGCN with pair
multiplicities and structural self weights, input diffusion that combines original and propagated features,
and LayerNorm, and evaluate these additions across benchmarks.
This tests whether simple lower-order improvements can narrow the
remaining gaps beyond the individual cases that motivated them.

\paragraph{Case study 1: Structural self weights and input diffusion (node classification on Walmart).}
Standard CEGCN uses a binary operator that records whether two
nodes are connected, but does not account for how often they
occur together in hyperedges (pair multiplicities) or the weights
on a node's own features derived from these pair counts
(structural self weights).
We therefore enrich CEGCN's pairwise operator and examine the resulting
performance changes. 
As shown in \cref{fig:remaining-gap-case-studies}(a), on
the fully perturbed view, standard CEGCN achieves $67.98\%$ accuracy,
compared with $76.68\%$ for the higher-order model.
Adding pair multiplicities alone yields a small improvement to $68.15\%$.
Further adding structural self weights raises accuracy to $74.56\%$.
Finally, propagating the input features once through the weighted
pairwise connections before CEGCN's learned layers
(one input-diffusion step) further raises accuracy to $76.32\%$.
Here, richer pairwise processing can recover most of the higher-order model's advantage.

\paragraph{Case study 2: Longer-range diffusion and classifier normalization (node classification on Trivago).}
Standard CEGCN uses two learned propagation layers, which may underuse the information available through longer-range pairwise diffusion.
We therefore add input diffusion before these layers and examine its effect on performance.
As shown in \cref{fig:remaining-gap-case-studies}(b), on
the fully perturbed view, standard CEGCN achieves $47.75\%$ accuracy, compared with $90.54\%$ for the higher-order model.
Propagating the input features through pairwise connections
for $k=256$ steps before the learned layers raises accuracy
to $78.32\%$.
Finally, keeping $k=256$ and adding LayerNorm to normalize
each node's hidden feature representation between CEGCN's
two learned layers raises accuracy to $90.92\%$.
Here, pairwise diffusion and classifier normalization can nearly reproduce the higher-order model's fully perturbed performance.

\begin{figure}[!t]
  \centering  
  \includegraphics[width=\textwidth]{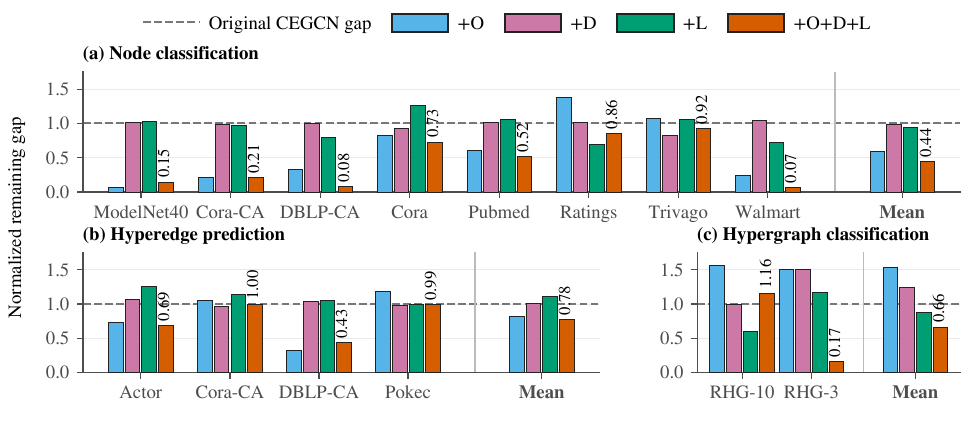}
  \vspace{-12mm}
  \caption{\textbf{Simple additions to a lower-order baseline reduce its
  performance gap to higher-order models on fully perturbed hypergraphs, especially in node classification.}
  For each benchmark with a largely retained gap, four bars show
  the gap to the higher-order reference evaluated on the fully
  perturbed view, after adding O, D, L, or O+D+L to CEGCN.
  O incorporates pair multiplicities and structural self weights;
  D combines the original input features with features propagated
  through pairwise connections; L applies LayerNorm to hidden features.
  Value labels report the O+D+L ratios.
  Each gap is divided by the gap over unmodified CEGCN:
  $1$ (the dashed line) means no change in that gap, and $0$
  means matching the higher-order reference.
  Each task mean averages its benchmark ratios equally.}
  \label{fig:stronger-endpoint-gaps}
  \vspace{-5mm}
\end{figure}

\paragraph{Case study 3: Cold-node handling and activation choice (hyperedge prediction on Pokec).}
For hyperedge prediction on Pokec, standard CEGCN and the higher-order reference, HGNN, differ in how they handle ``cold'' nodes absent from the training structure.
CEGCN's identity policy retains these nodes' transformed features through
self loops, whereas HGNN's zero-message policy removes their
feature-dependent messages at each propagation layer.
We therefore match CEGCN's cold-node policy to HGNN's and examine the
resulting performance change.
As shown in \cref{fig:remaining-gap-case-studies}(c), on the fully perturbed view, standard CEGCN achieves $59.71\%$ AP, compared with $83.19\%$ for the higher-order reference. 
CEGCN with the matched zero-message policy reaches $78.22\%$ AP.
Finally, replacing the hidden ReLU activation between CEGCN's two
propagation layers with ELU, as used by HGNN, raises AP to $80.14\%$.
Here, matching the cold-node policy and activation function can recover most of the higher-order model's advantage.

\paragraph{Evidence beyond the three case studies.}
Additional analyses identify other lower-order explanations for
advantages that survive perturbation.
For hyperedge prediction on Actor, the selected TFHNN configuration
makes predictions from node features without using hypergraph
structure, so its advantage cannot be attributed to higher-order
information.
For hypergraph classification on RHG-10 and RHG-3, models using
only weighted pairwise connections recover all or much of the
higher-order models' remaining advantage.
These results further show that an advantage surviving perturbation
can arise from feature processing or pairwise structure alone
(\cref{app:remaining-gap-analysis}).

\paragraph{Lower-order baselines motivated by the case studies.}
Motivated by these cases, we test the effect of three simple additions to standard CEGCN:
\textbf{O} replaces the binary propagation operator with the operator using pair counts and structural self weights, with symmetric
normalization to scale messages by the total connection weights of
their sending and receiving nodes (\cref{app:cegcn-additions});
\textbf{D} combines the original input features with features
propagated through pairwise connections for different numbers
of steps;
and \textbf{L} inserts LayerNorm.
Addition \textbf{O} is inspired by \cref{fig:remaining-gap-case-studies}(a), and \textbf{D} and \textbf{L} are inspired by \cref{fig:remaining-gap-case-studies}(b).
For hyperedge prediction, we also compare two policies for cold
nodes: retaining their transformed features (the identity policy)
or removing their feature-dependent messages (the zero-message
policy), motivated by \cref{fig:remaining-gap-case-studies}(c).
We inherit official CEGCN configurations and compare CEGCN
with +O, +D, +L, and +O+D+L.
\cref{fig:stronger-endpoint-gaps} compares gaps on the fully perturbed
view for the 14 benchmarks with largely retained gaps before and after strengthening CEGCN with O, D, L, or their combination. 
We express each gap relative to the gap between the higher-order
reference and unmodified CEGCN, both evaluated on the fully
perturbed view.
A ratio of $1$ means that the original CEGCN gap remains unchanged,
whereas $0$ means that the improved CEGCN matches the higher-order
reference.

\emph{Observation 1: Simple lower-order additions narrow retained gaps, especially in node classification.}
O+D+L reduces the mean remaining-gap ratio to $0.44$ for node
classification, $0.78$ for hyperedge prediction using the identity
policy for cold nodes,
and $0.66$ for hypergraph classification.
These improvements extend beyond the eight node-classification
benchmarks with largely retained gaps.
Across all 13 node-classification benchmarks, O and O+D+L improve
accuracy by an average of $2.96$ and $3.69$ points, respectively
(\cref{app:cegcn-additions}).
Decomposing O on the original views further shows that pair
multiplicities alone (O$_1$) yield an average gain of only
$0.38$ points across these 13 benchmarks.
With pair multiplicities already included, adding structural self
weights and adjusting the associated normalization (O$_2$)
yields a further $2.58$ points on average
(\cref{tab:node-cls-o1o2}). Thus, much of the improvement comes from how pairwise information is
normalized and combined with self information.

\begin{figure}[!htbp]
  \centering  
  \includegraphics[width=0.60\textwidth]{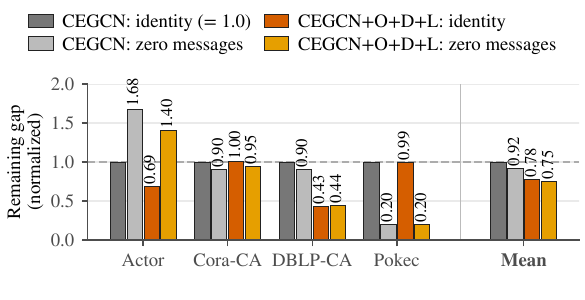}
  \caption{\textbf{Cold-node handling has benchmark-specific effects.}
  CEGCN and CEGCN+O+D+L are compared on four hyperedge-prediction benchmarks
  with largely retained gaps. For nodes absent from the training structure,
  the identity policy retains transformed features; the zero-message policy
  removes feature-dependent messages. All gaps use the same higher-order
  reference on the fully perturbed view and are divided by its gap over
  CEGCN using the identity policy ($1$ means this gap is unchanged;
  $0$ means matching the reference). The task mean averages the four
  benchmark ratios equally.}
  \label{fig:edge-cold-policy-gaps}
\end{figure}

\emph{Observation 2: Benefits vary across tasks and datasets.}
The additions leave substantial gaps on some benchmarks, and the effect
of cold-node handling varies across hyperedge-prediction benchmarks
(\cref{fig:edge-cold-policy-gaps}).
The gains also do not extend consistently to hypergraph classification
beyond the benchmarks with largely retained gaps: across all six benchmarks,
O+D+L reduces accuracy on five, with a mean change of $-0.95$ points
(\cref{app:cegcn-additions}).

\emph{Overall: Stronger lower-order comparisons can help assess
performance attribution.}
These experiments show that simple lower-order additions can
narrow some advantages that survive perturbation, although their
benefits vary across benchmarks.
They motivate evaluating stronger lower-order alternatives before
interpreting a performance gap as evidence for higher-order information.

\paragraph{Additional results.}
\cref{app:remaining-gap-cases} reports detailed results for individual
benchmarks, the O/D/L additions, and secondary metrics.

\section{Broader Implications for the Hypergraph-Learning Community}
\label{sec:benchmark_audit}
Our findings motivate three changes in how the hypergraph learning community 
evaluates hypergraph learning, regarding evaluation protocol, baselines, and benchmarks.

\paragraph{Evaluation protocols.}
A leaderboard comparison can establish whether one model outperforms another, but it does not by itself identify the source of that advantage.
Our results in \cref{sec:real_world} make this distinction concrete: many advantages of higher-order models largely remain even after higher-order information is perturbed.
We therefore recommend complementing original-view performance comparisons with
controlled interventions before attributing a gain to a particular
information source.
Specifically, for higher-order attribution, an evaluation should explicitly declare the
lower-order information being preserved, verify that the intervention preserves
it, and report both the portion of the original gap removed by the intervention
and the portion that remains.

\paragraph{Evaluation baselines.}
A weak lower-order comparator cannot establish that the information available
to lower-order models is itself insufficient.
We illustrate this distinction in \cref{sec:remaining_gaps}:
pair multiplicities, structural self-weights, diffusion, normalization, and
protocol choices can substantially change how effectively lower-order
information is used.
Attribution therefore requires stronger lower-order comparisons.
Evaluations should include competitive feature-only models and richer pairwise
models, with validation-based tuning opportunities comparable to those given to
higher-order models.
This recommendation is consistent with graph-learning studies in which stronger
baselines or improved hyperparameter choices substantially changed previously
reported architectural gaps~\citep{errica2020fair,tonshoff2024gap,katsman2026revisiting}.

\paragraph{Evaluation benchmarks.}
In a hypergraph benchmark, the existence of higher-order structure does not by itself mean that
higher-order information is needed to predict the target.
A relation may genuinely arise from a group interaction while the target remains predictable from features or pairwise statistics. 
Claims about higher-order necessity should therefore be made at the
benchmark (i.e., dataset--task) level rather than inferred from the fact that the data admit a
hypergraph representation.
Benchmarks designed to test specific information requirements can
make this distinction more testable.
Our synthetic controls in \cref{sec:synthetic} illustrate one approach:
include cases in which higher-order information is necessary by construction
alongside cases in which lower-order information is sufficient.
For real benchmarks, documentation should explain how hyperedges
and prediction targets are constructed, helping readers assess
whether the task tests higher-order information or admits
plausible feature-only or pairwise explanations.
This perspective is related to data-centric approaches such as RINGS, which
evaluate whether graph datasets distinguish the information sources they are
intended to test \citep{coupette2025rings}, and to capability-targeted suites
such as the Long Range Graph Benchmark \citep{dwivedi2022longrange}.
\color{black}

\section{Conclusion}
\label{sec:conclusion}
Existing work on hypergraph learning often motivates higher-order models by their ability to exploit non-pairwise structure, and empirical improvements over lower-order baselines are commonly interpreted as evidence for this
benefit. In this paper, we separate the observation of a performance advantage from its attribution and ask: \emph{do higher-order models actually win for higher-order reasons?} (\cref{sec:problem_formulation}).
We introduce a controlled performance-attribution framework that perturbs higher-order information while preserving the lower-order, i.e., pairwise, information, allowing the original advantage of the higher-order model to be decomposed into the portion removed by the intervention and the portion that remains (\cref{sec:framework}). 
Our synthetic controls recover the expected behaviors when higher-order information is either essential or redundant by
construction (\cref{sec:synthetic}). 
Across real-world benchmarks, however, we frequently observe a different pattern: a higher-order model has an original advantage of at least one percentage point, yet most of that advantage survives after the higher-order information is perturbed (\cref{sec:real_world}). 
Thus, the original performance gap alone can substantially overstate the evidence that the perturbed higher-order information is responsible for the advantage.
We then investigate what may sustain these retained gaps. Representative case studies identify plausible lower-order explanations, and simple additions to a lower-order baseline improve its performance (\cref{sec:remaining_gaps}). 
Together, our findings motivate attribution-aware evaluation in hypergraph learning.
This perspective calls for controlled interventions, stronger
lower-order comparators, and benchmarks designed to distinguish the capabilities they are intended to test (\cref{sec:benchmark_audit}).

\section{Limitations and Future Directions}
\label{sec:limitations}
\paragraph{Preserved information and alternative interventions.}
Our study tests whether performance gains depend on information beyond
the multiplicity-weighted clique expansion (\cref{def:information-order}).
The intervention preserves this representation exactly while changing
how nodes are grouped into hyperedges.
Our attribution statements therefore depend on this choice of preserved
lower-order information. A natural extension is to compare other ways
of constructing perturbed hypergraphs that preserve the same information.

\paragraph{From information sufficiency to model mechanisms.}
Our controls address what information is sufficient to achieve a given
performance. 
A complementary question is how the original trained model uses that information.
When a model using only lower-order information matches the
higher-order model's performance, it shows that higher-order
information is not needed to achieve that performance in the
tested comparison (\cref{app:remaining-gap-analysis}).
However, this does not show that the original higher-order model
achieved its performance in the same way.
Combining the present interventions with controlled changes to individual
architectural components would help identify how the original model
uses the information available to it.

\paragraph{Benchmarks with known information requirements.}
Our real-world analysis and synthetic controls provide complementary starting points for future benchmark development (\cref{sec:real_world,sec:synthetic,app:synthetic}).
Real-world benchmarks offer breadth and realistic heterogeneity, but
we generally do not know in advance which information is needed to
achieve a given level of predictive performance. 
The synthetic controls establish the role of higher-order information by construction, while simplifying that heterogeneity.
Semi-synthetic benchmarks could bridge these settings by retaining realistic features and relational structure while introducing controlled prediction targets whose dependence on higher-order information is known.

\bibliography{references}
\bibliographystyle{tmlr}

\clearpage
\appendix
\crefalias{section}{appendix}
\crefalias{subsection}{appendix}
\crefalias{subsubsection}{appendix}
\addtocontents{toc}{\protect\setcounter{tocdepth}{2}}
\renewcommand{\contentsname}{Appendix Contents}
\tableofcontents
\clearpage
\section{Details of Quotes from Prior Work (Supplementing \cref{sec:introduction})}
\label{app:motivating-interpretations}

To document the empirical interpretation motivating our study, we quote
several representative statements from prior work. 
After reporting improvements over graph convolutional networks,
\citet[p.~3564]{feng2019hypergraph} write that
\enquote{The better performance can be dedicated to the employed hypergraph
structure}.
Similarly, \citet[Sec.~4.1.3]{bai2021hypergraph} interpret their comparison as
demonstrating
\enquote{the benefit of considering higher-order models \ldots{} to exploit
non-pairwise relationships}.
A similar interpretation appears in more recent benchmark and application
work: \citet[Sec.~4.1.1]{li2026dhgbench} suggest that clique expansion
\enquote{disrupts high-order structures and degrades predictive performance},
while \citet[p.~588]{xu2023hypehr} conclude that
\enquote{modeling higher-order interactions beyond pairwise contributes to
performance gain}.

\section{Details of Partial Perturbations (Supplementing \cref{sec:framework})}
\label{app:partial-perturbation-construction}

For a hypergraph $\mathcal H$, define the multiset of eligible hyperedge instances containing three or more nodes
as
\begin{equation}
  \mathcal I_{>2}(\mathcal H)
  =\biguplus_{\substack{e\in\mathcal E\\ |e|>2}}
    \bigl\{(e,j):1\leq j\leq \mathbf w(e)\bigr\}.
  \label{eq:eligible-hyperedge-instances}
\end{equation}
Thus, each unit of hyperedge weight is one eligible instance, whereas
pairwise hyperedge instances are never selected for expansion.
Singleton hyperedges retained as background input are also never selected
for expansion and remain unchanged throughout the intervention.
Equivalently, $\mathcal I_{>2}(\mathcal H)$ is a list of length
\begin{equation}
  N(\mathcal H)
  =\bigl|\mathcal I_{>2}(\mathcal H)\bigr|
  =\sum_{\substack{e\in\mathcal E\\ |e|>2}}\mathbf w(e).
  \label{eq:number-eligible-hyperedge-instances}
\end{equation}

We independently draw $k$ random permutations
$\pi_1,\ldots,\pi_k$ of this list.  Each permutation defines one nested
perturbation trajectory.  At level $r\in[0,1]$, let
\begin{equation}
  m_r=\min\!\left\{N(\mathcal H),
    \left\lfloor rN(\mathcal H)+\tfrac12\right\rfloor\right\}
  \quad\text{and}\quad
  \mathcal S_i^{(r)}
  =\{\pi_i(1),\ldots,\pi_i(m_r)\}.
  \label{eq:partial-trajectory-prefix}
\end{equation}
Thus, trajectory $i$ clique-expands the first $m_r$ instances in
$\pi_i$ and leaves all other instances unchanged.  Using prefixes makes the
selected instances nested as $r$ increases.  It also gives
$\mathcal S_i^{(0)}=\varnothing$ and
$\mathcal S_i^{(1)}=\mathcal I_{>2}(\mathcal H)$, so all trajectories share
the same two endpoints; only the intermediate views depend on the random
permutation.

The resulting view $\mathcal H_i^{(r)}$ keeps every unselected hyperedge
instance, each containing two or more nodes, and replaces each selected instance $(e,j)$ by
$\operatorname{clique}(e)$; its edge multiset is therefore the union of the
unselected instances and these clique expansions.  When clique expansions
produce the same pair, their multiplicities add.
At $r=1$, the non-singleton structure contains only weighted pairs
and equals $\operatorname{WCE}(\mathcal H)$.
Singleton hyperedges present in the input are retained unchanged
at every perturbation level, including this endpoint.
Thus, the intervention preserves weighted pairwise co-occurrences
and singleton incidences while removing the original grouping
into hyperedges containing three or more nodes.
In particular, the weight
of any pair $p$ in the resulting view is
\begin{equation}
  \mathbf w_i^{(r)}(p)
  =\mathbf w(p)
   +\sum_{(e,j)\in\mathcal S_i^{(r)}}
      \mathbf 1[p\in\operatorname{clique}(e)],
  \qquad |p|=2,
  \label{eq:partial-pair-weights}
\end{equation}
where $\mathbf w(p)=0$ if $p$ is not an original hyperedge.  Each replacement
preserves weighted pairwise co-occurrences exactly, so
$\operatorname{WCE}(\mathcal H_i^{(r)})=\operatorname{WCE}(\mathcal H)$ for
every $i$ and $r$; the partial views therefore differ only in information
beyond the declared weighted clique expansion.

Replacing the intervened structure in $\mathcal D$ by $\mathcal H_i^{(r)}$ produces
$\mathcal D_i^{(r)}$ and the benchmark view
$\mathcal B_i^{(r)}=(\mathcal D_i^{(r)},\tau)$.  For hypergraph-classification
datasets, the construction is applied to each hypergraph independently.  The
task $\tau$ remains fixed.  At $r=0$ and $r=1$, the views are independent of
the sampled trajectory, so we omit $i$ and recover $\mathcal B^{(0)}$ and
$\mathcal B^{(1)}$.

Let $P_i^{(r)}(M)$ denote the test performance of model $M$ on the view
$\mathcal B_i^{(r)}$.  For the $k$ sampled trajectories, the performance and
gap between the higher- and lower-order models reported at level $r$ are
\begin{align}
  P^{(r)}(M)
  &=\frac{1}{k}\sum_{i=1}^{k}P_i^{(r)}(M),\nonumber\\
  \Delta^{(r)}(M_H;M_L)
  &=P^{(r)}(M_H)-P(M_L).
  \label{eq:trajectory-gaps}
\end{align}
At $r=0$ and $r=1$, the sampled views are identical across trajectories, so
these quantities recover the endpoint notation in \cref{sec:framework}.

\section{Details of Experimental Setup (Supplementing \cref{sec:empirical_analysis:setup})}
\label{app:experimental_setup}

\subsection{Benchmarks: Datasets and tasks}
\label{app:dataset_details}
We inherit the benchmarks from DHG-Bench~\citep{li2026dhgbench}.  
Node classification and hyperedge prediction each operate on one attributed hypergraph, while hypergraph classification operates on a collection of attributed hypergraphs.
The full inventory in \cref{sec:empirical_analysis:setup} contains 13 node-classification, six hyperedge-prediction, and six hypergraph-classification benchmarks, formed from 19 distinct datasets.
Below, we summarize what the entities, hyperedges, features, and labels represent.

\subsubsection{Datasets for node classification and hyperedge prediction}
All 13 datasets below are used for node classification.  Cora, Pubmed,
Cora-CA, DBLP-CA, Actor, and Pokec are additionally paired with the
hyperedge-prediction task, producing six separate benchmarks.

\paragraph{Cora.}
Cora is a co-citation hypergraph: nodes are papers, and a hyperedge groups
papers cited together.  Bag-of-words document vectors serve as node features,
and paper topics are the node labels.  We use the hypergraph construction
released with HyperGCN~\citep{yadati2019hypergcn}.

\paragraph{Pubmed.}
Pubmed uses the same co-citation construction as Cora, with papers as nodes,
jointly cited papers as hyperedges, bag-of-words features, and research topics
as labels~\citep{yadati2019hypergcn}.

\paragraph{Cora-CA.}
Cora-CA is the co-authorship view of Cora.  Nodes are papers, and each author
induces a hyperedge containing that author's papers; features and labels again
encode document words and paper topics, respectively~\citep{yadati2019hypergcn}.

\paragraph{DBLP-CA.}
DBLP-CA is a larger co-authorship hypergraph constructed analogously to
Cora-CA: papers are nodes, an author's papers form a hyperedge, and the task is
to predict paper categories from bag-of-words features and the hypergraph
structure~\citep{yadati2019hypergcn}.

\paragraph{NTU2012.}
NTU2012 contains 3D shapes from 67 categories.  Each shape is a node;
groups of a fixed number of nearest neighbors in the visual-feature
space form hyperedges, and the node features are derived from MVCNN and GVCNN shape
representations.  We use the processed hypergraph released with
HGNN~\citep{feng2019hypergraph}.

\paragraph{ModelNet40.}
ModelNet40 contains 3D objects from 40 categories and is processed in the same
way as NTU2012: objects are nodes, nearby objects in the learned visual-feature
space form hyperedges, and the object category is the prediction
target~\citep{feng2019hypergraph}.

\paragraph{Walmart.}
In Walmart, nodes are products and each shopping trip induces a hyperedge of
co-purchased products.  Labels are 11 broad product departments.  Following
AllSet, the node features are noisy 100-dimensional encodings of these
labels~\citep{amburg2020clustering,chien2022allset}.

\paragraph{Trivago.}
Trivago is a hotel-search hypergraph: hotels are nodes and a user's browsing
or click-out activity groups hotels into a hyperedge.  The node label is the
country in which the hotel is located~\citep{kim2023largescale}.

\paragraph{Actor.}
Actor is derived from the AMiner movie--actor--director--writer network.  Nodes
are people involved in film production, each film induces a hyperedge joining
its participants, Wikipedia keywords provide node attributes, and the label
identifies a person's role~\citep{li2025heterophily}.

\paragraph{Ratings.}
Ratings, called Amazon-ratings in its source, is derived from Amazon
co-purchasing metadata in the SNAP collection.  Products are nodes,
frequently co-purchased products are grouped by hyperedges, bag-of-words
vectors of product descriptions are the features, and the target is a
product's average review rating~\citep{leskovec2014snap,li2025heterophily}.

\paragraph{Gamers.}
Gamers, called Twitch-gamers in its source, represents Twitch accounts as
nodes and forms hyperedges from reciprocal follow relations within temporal
windows.  Account metadata provide the node features, and the binary label
records whether a channel hosts explicit content
\citep{rozemberczki2021twitch,li2025heterophily}.

\paragraph{Pokec.}
Pokec represents users of the Slovak social network as nodes.  Each user's
friend set forms a hyperedge; profile fields such as age, interests,
education, and region provide features, and user-reported gender is the
label~\citep{takac2012pokec,li2025heterophily}.

\paragraph{Yelp.}
Yelp represents restaurants as nodes and groups restaurants reviewed by the
same user into hyperedges.  The label is the restaurant's average star rating;
features combine geographic coordinates, location encodings, and a
bag-of-words representation of restaurant names~\citep{chien2022allset}.

\subsubsection{Datasets for hypergraph classification}
The six graph-level datasets were introduced or assembled for hypergraph
classification by~\citet{feng2024isomorphism}.  Here, each dataset contains a
collection of hypergraphs and each entire hypergraph is one prediction
instance.

\paragraph{RHG-10.}
RHG-10 is a synthetic collection with ten structural classes: Hyper Flower,
Hyper Pyramid, Hyper Checked Table, Hyper Wheel, Hyper Lattice, Hyper Windmill,
Hyper Firm Pyramid, Hyper RChecked Table, Hyper Cycle, and Hyper Fern.

\paragraph{RHG-3.}
RHG-3 is a focused synthetic collection with three structural classes---Hyper
Pyramid, Hyper Checked Table, and Hyper Wheel---generated with variation
within each class.

\paragraph{IMDB-Dir-Form.}
IMDB-Dir-Form contains movie-centered hypergraphs constructed from IMDB
co-director relations.  The graph label is the movie form, with classes such
as animation, documentary, and drama.

\paragraph{IMDB-Dir-Genre.}
IMDB-Dir-Genre uses the same co-director construction but labels the
hypergraphs by movie genre, with classes such as adventure, crime, and family.

\paragraph{Steam-Player.}
Each Steam-Player instance represents one player's game collection.  Games
are vertices, games sharing tags are linked by hyperedges, and the binary
target distinguishes a preference for single-player or multiplayer games.

\paragraph{Twitter-Friend.}
Each Twitter-Friend instance represents the friendship neighborhood of a
focal Twitter user, with hyperedges derived from friendship relations.  The
label distinguishes whether the focal user's post concerns ``National Dog
Day'' or ``Respect Tyler Joseph.''

\subsubsection{Other details}

\paragraph{Benchmark construction.}

We canonicalize each source hypergraph by collapsing repeated memberships within an edge and removing empty hyperedges.
Repeated hyperedges remain as integer multiplicities.
For node classification, we retain the full source node set, including
nodes incident only to singleton hyperedges and nodes with no incident
hyperedge, and retain singleton hyperedges when present.
We use label-stratified 50/25/25 train/validation/test splits.
For example, the Actor node-classification inputs contain 156 singleton
hyperedge instances, which remain unchanged at every perturbation level.
For hyperedge prediction, preprocessing removes singleton hyperedges
and restricts the node set to nodes incident to an edge containing at
least two nodes before splitting the positive hyperedges.
This node universe remains fixed even when a node is absent from the
training support. We use fixed 60/20/20 train/validation/test partitions.
Hypergraph classification canonicalizes each graph independently,
removes singleton hyperedges, retains nodes incident to edges of size
two or larger, and uses fixed 80/10/10 graph-level splits.
Across all three tasks, the intervention preserves node features, labels, prediction targets, graph order, and split membership.

\paragraph{Split details for hyperedge prediction.}
\label{app:edge_support_protocol}

Hyperedge prediction requires an observed relational structure from which the encoder can compute node representations.
We partition the distinct positive hyperedge types into 60\% training, 20\% validation, and 20\% test sets, while retaining their multiplicities.
The training positives serve both as supervised training queries and as the only hypergraph structure available to the encoder, which we call the \emph{support graph}.
Validation and test positives are used only as prediction queries and never enter message passing.
We construct sets of negative prediction queries matched to the sizes of the positive queries for each split and keep all positive queries, negative queries, and the node universe fixed across perturbation levels.
Thus, for hyperedge prediction, the intervention changes only the support graph derived from the training hyperedges.

\subsection{Models}
\label{app:model_details}

\paragraph{Spectral HNNs.}
\textbf{HGNN} applies convolution derived from a normalized hypergraph
Laplacian~\citep{feng2019hypergraph}.
\textbf{HyperGCN} replaces each hyperedge with representative pairwise links
before applying graph convolution~\citep{yadati2019hypergcn}.
\textbf{HCHA} provides trainable hypergraph-convolution and
hypergraph-attention operators~\citep{bai2021hypergraph}.
\textbf{LEGCN} constructs a lossless line expansion whose vertices represent
node--hyperedge memberships (incidences)~\citep{yang2022legcn}.
\textbf{HyperND} computes node representations through nonlinear diffusion on
the hypergraph~\citep{prokopchik2022hypernd}.
\textbf{PhenomNN} derives propagation from parameterized
hypergraph-regularized energy functions~\citep{wang2023phenomnn}.
\textbf{SheafHyperGNN} augments hypergraph Laplacians with learned sheaf
restriction maps~\citep{duta2023sheafhypergnn}.
\textbf{HJRL} jointly represents nodes and hyperedges through a cross
expansion and a reconstruction objective~\citep{yan2024hjrl}.
\textbf{DPHGNN} combines static topology-derived features with dynamic
hypergraph message passing~\citep{saxena2024dphgnn}.
\textbf{TF-HNN} separates an incidence-derived diffusion step from its
trainable prediction module~\citep{tang2025tfhnn}.

\paragraph{Spatial HNNs.}
\textbf{HNHN} alternates nonlinear node-to-hyperedge and hyperedge-to-node
updates with degree-based normalization~\citep{dong2020hnhn}.
\textbf{UniGNN} generalizes GNN aggregation to two-stage hypergraph message
passing; following DHG-Bench, we evaluate its UniGCNII
variant~\citep{huang2021unignn}.
\textbf{AllSetTransformer} parameterizes both stages of hypergraph propagation
as learned multiset functions~\citep{chien2022allset}.
\textbf{ED-HNN} approximates equivariant hypergraph diffusion through message
passing on the star expansion, a bipartite graph connecting nodes
to the hyperedges that contain them~\citep{wang2023edhnn}.
\textbf{HyperGT} combines global Transformer interactions with
incidence-derived positional encodings and structural
regularization~\citep{liu2024hypergt}.

\paragraph{Tensor-based HNNs.}
\textbf{EHNN} constructs equivariant layers by connecting sparse hypergraphs
to fixed-order tensors and sharing parameters across hyperedge
orders~\citep{kim2022ehnn}.
\textbf{T-HyperGNN} represents hypergraph structure with multiplicative
interaction tensors and performs tensor message passing~\citep{wang2025thypergnn}.

\paragraph{Lower-order baselines.}
\textbf{MLP} uses only node features.  \textbf{CEGCN} and \textbf{CEGAT}
apply GCN and GAT, respectively, to the binary (unweighted) clique expansion
\citep{chien2022allset}.

\subsection{Implementations and hyperparameter tuning}
We use the official implementations from
DHG-Bench~\citep{li2026dhgbench}.
The benchmark pool begins with the 20 released DHG-Bench methods: 17 HNNs and the MLP, CEGCN, and CEGAT baselines.
For hyperparameter tuning, we use the released YAML configuration for each dataset, task, and method provided by~\citet{li2026dhgbench}.
For completeness, \cref{tab:dhgb-pairs} gives the selected model pair on each
primary benchmark comparison. It is followed by one table per model,
with benchmarks as rows and varying hyperparameters as columns.

\begingroup\setlength{\tabcolsep}{5pt}
\setlength{\LTcapwidth}{\textwidth}
\begin{longtable}{@{}llll@{}}
\caption{Selected model pairs for the 25 primary benchmark comparisons. NC: node classification; EP: hyperedge prediction; HC: hypergraph classification. The Yelp pair includes the resource exception described in the results.}\label{tab:dhgb-pairs}\\
\toprule Task & Benchmark & Higher-order model & Lower-order model\\\midrule
\endfirsthead\toprule Task & Benchmark & Higher-order model & Lower-order model\\\midrule\endhead
EP & Actor & TF-HNN & MLP\\
EP & Cora-CA & UniGCNII & CEGAT\\
EP & DBLP-CA & PhenomNN & CEGAT\\
EP & Cora & PhenomNN & CEGAT\\
EP & Pubmed & EHNN & CEGCN\\
EP & Pokec & HGNN & CEGCN\\
HC & IMDB-Dir-Form & HCHA & CEGCN\\
HC & IMDB-Dir-Genre & HNHN & CEGCN\\
HC & RHG-10 & HNHN & CEGAT\\
HC & RHG-3 & HNHN & CEGCN\\
HC & Steam-Player & HNHN & CEGAT\\
HC & Twitter-Friend & HJRL & MLP\\
NC & ModelNet40 & ED-HNN & MLP\\
NC & NTU2012 & LEGCN & MLP\\
NC & Actor & EHNN & MLP\\
NC & Cora-CA & TF-HNN & CEGCN\\
NC & DBLP-CA & PhenomNN & CEGCN\\
NC & Cora & TF-HNN & CEGCN\\
NC & Pubmed & LEGCN & MLP\\
NC & Gamers & HCHA & MLP\\
NC & Pokec & TF-HNN & MLP\\
NC & Ratings & LEGCN & MLP\\
NC & Trivago & TF-HNN & CEGCN\\
NC & Walmart & ED-HNN & CEGCN\\
NC & Yelp & TF-HNN & MLP\\
\bottomrule\end{longtable}\endgroup
\paragraph{Model-wise YAML settings.}
Each table lists benchmarks as rows and varying YAML fields as columns. Settings shared by every listed row appear once below the table. Column symbols $L,E,H,M,T,p,\eta,\lambda$ denote \texttt{All\_num\_layers}, \texttt{embedding\_hidden}, \texttt{MLP\_hidden}, \texttt{MLP\_num\_layers}, \texttt{epochs}, \texttt{dropout}, \texttt{lr}, and \texttt{wd}, respectively; additional columns are defined below each table. NC/EP/HC have the same meanings as above.

\begingroup
\setlength{\tabcolsep}{3pt}
\setlength{\LTcapwidth}{\textwidth}
\renewcommand{\arraystretch}{1.08}
\begin{longtable}{@{}llrrrrrrr@{}}
\caption{DHG-Bench YAML settings for MLP.}\label{tab:yaml-mlp}\\
\toprule Task & Benchmark & $L$ & $H$ & $T$ & $p$ & $\eta$ & $q_{1}$ & $q_{2}$\\\midrule
\endfirsthead
\toprule Task & Benchmark & $L$ & $H$ & $T$ & $p$ & $\eta$ & $q_{1}$ & $q_{2}$\\\midrule
\endhead
EP & Actor & 2 & 128 & 200 & 0.8 & 0.01 & -- & --\\
EP & Cora-CA & 2 & 128 & 100 & 0.8 & 0.001 & -- & --\\
EP & DBLP-CA & 3 & 128 & 100 & 0.8 & 0.001 & -- & --\\
EP & Cora & 3 & 128 & 100 & 0.8 & 0.001 & -- & --\\
EP & Pubmed & 3 & 128 & 100 & 0.8 & 0.001 & -- & --\\
EP & Pokec & 2 & 256 & 100 & 0 & 0.01 & -- & --\\
HC & IMDB-Dir-Form & 2 & 128 & 100 & 0.8 & 0.01 & mean & true\\
HC & IMDB-Dir-Genre & 2 & 128 & 200 & 0.8 & 0.001 & mean & true\\
HC & RHG-10 & 2 & 128 & 200 & 0.8 & 0.01 & mean & true\\
HC & RHG-3 & 2 & 128 & 200 & 0.8 & 0.01 & mean & true\\
HC & Steam-Player & 2 & 128 & 200 & 0.8 & 0.001 & mean & true\\
HC & Twitter-Friend & 2 & 128 & 200 & 0.8 & 0.001 & mean & true\\
NC & ModelNet40 & 2 & 128 & 400 & 0.8 & 0.001 & -- & --\\
NC & NTU2012 & 2 & 128 & 400 & 0.8 & 0.001 & -- & --\\
NC & Actor & 2 & 128 & 200 & 0.8 & 0.01 & -- & --\\
NC & Cora-CA & 2 & 128 & 100 & 0.8 & 0.001 & -- & --\\
NC & DBLP-CA & 3 & 128 & 100 & 0.8 & 0.001 & -- & --\\
NC & Cora & 3 & 128 & 100 & 0.8 & 0.001 & -- & --\\
NC & Pubmed & 3 & 128 & 100 & 0.8 & 0.001 & -- & --\\
NC & Gamers & 2 & 256 & 200 & 0.8 & 0.01 & -- & --\\
NC & Pokec & 2 & 256 & 100 & 0 & 0.01 & -- & --\\
NC & Ratings & 2 & 256 & 100 & 0.8 & 0.01 & -- & --\\
NC & Trivago & 2 & 128 & 500 & 0.8 & 0.05 & -- & --\\
NC & Walmart & 2 & 256 & 200 & 0.8 & 0.01 & -- & --\\
NC & Yelp & 3 & 128 & 500 & 0.8 & 0.01 & -- & --\\
\bottomrule\end{longtable}
\noindent 
Additional columns: $q_{1}$=\texttt{pooling}; $q_{2}$=\texttt{early\_\allowbreak{}stop}.\par\smallskip
\noindent Shared settings: \texttt{embedding\_\allowbreak{}hidden}=128; \texttt{wd}=0; \texttt{normalization}=ln; \texttt{InputNorm}=false.\par
\endgroup\medskip
\begingroup
\setlength{\tabcolsep}{3pt}
\setlength{\LTcapwidth}{\textwidth}
\renewcommand{\arraystretch}{1.08}
\begin{longtable}{@{}llrrrrrrrr@{}}
\caption{DHG-Bench YAML settings for CEGCN.}\label{tab:yaml-cegcn}\\
\toprule Task & Benchmark & $E$ & $H$ & $T$ & $p$ & $\eta$ & $q_{1}$ & $q_{2}$ & $q_{3}$\\\midrule
\endfirsthead
\toprule Task & Benchmark & $E$ & $H$ & $T$ & $p$ & $\eta$ & $q_{1}$ & $q_{2}$ & $q_{3}$\\\midrule
\endhead
EP & Actor & 128 & 64 & 100 & 0.5 & 0.05 & false & -- & --\\
EP & Cora-CA & 128 & 128 & 100 & 0.5 & 0.01 & false & -- & --\\
EP & DBLP-CA & 128 & 128 & 100 & 0.5 & 0.01 & false & -- & --\\
EP & Cora & 128 & 256 & 100 & 0 & 0.001 & false & -- & --\\
EP & Pubmed & 128 & 128 & 100 & 0.5 & 0.01 & false & -- & --\\
EP & Pokec & 128 & 64 & 100 & 0.5 & 0.05 & false & -- & --\\
HC & IMDB-Dir-Form & 64 & 128 & 200 & 0.2 & 0.001 & false & mean & true\\
HC & IMDB-Dir-Genre & 64 & 128 & 100 & 0.2 & 0.001 & false & max & true\\
HC & RHG-10 & 64 & 128 & 200 & 0.2 & 0.001 & false & mean & true\\
HC & RHG-3 & 64 & 128 & 100 & 0.2 & 0.001 & false & mean & true\\
HC & Steam-Player & 64 & 128 & 100 & 0.2 & 0.001 & false & max & true\\
HC & Twitter-Friend & 64 & 128 & 100 & 0.2 & 0.001 & false & max & true\\
NC & ModelNet40 & 128 & 256 & 300 & 0 & 0.001 & true & -- & --\\
NC & NTU2012 & 128 & 256 & 300 & 0 & 0.001 & true & -- & --\\
NC & Actor & 128 & 64 & 100 & 0.5 & 0.05 & false & -- & --\\
NC & Cora-CA & 128 & 128 & 100 & 0.5 & 0.01 & false & -- & --\\
NC & DBLP-CA & 128 & 128 & 100 & 0.5 & 0.01 & false & -- & --\\
NC & Cora & 128 & 256 & 100 & 0 & 0.001 & false & -- & --\\
NC & Pubmed & 128 & 128 & 100 & 0.5 & 0.01 & false & -- & --\\
NC & Gamers & 128 & 64 & 100 & 0.5 & 0.05 & false & -- & --\\
NC & Pokec & 128 & 64 & 100 & 0.5 & 0.05 & false & -- & --\\
NC & Ratings & 128 & 64 & 100 & 0.5 & 0.05 & false & -- & --\\
NC & Trivago & 128 & 128 & 1000 & 0.5 & 0.01 & false & -- & --\\
NC & Walmart & 128 & 128 & 100 & 0.5 & 0.01 & false & -- & --\\
NC & Yelp & 128 & 128 & 100 & 0.5 & 0.01 & false & -- & --\\
\bottomrule\end{longtable}
\noindent 
Additional columns: $q_{1}$=\texttt{use\_\allowbreak{}bn}; $q_{2}$=\texttt{pooling}; $q_{3}$=\texttt{early\_\allowbreak{}stop}.\par\smallskip
\noindent Shared settings: \texttt{All\_\allowbreak{}num\_\allowbreak{}layers}=2; \texttt{wd}=0.\par
\endgroup\medskip
\begingroup
\setlength{\tabcolsep}{3pt}
\setlength{\LTcapwidth}{\textwidth}
\renewcommand{\arraystretch}{1.08}
\begin{longtable}{@{}llrrrrrrrrrr@{}}
\caption{DHG-Bench YAML settings for CEGAT.}\label{tab:yaml-cegat}\\
\toprule Task & Benchmark & $E$ & $H$ & $T$ & $p$ & $\eta$ & $q_{1}$ & $q_{2}$ & $q_{3}$ & $q_{4}$ & $q_{5}$\\\midrule
\endfirsthead
\toprule Task & Benchmark & $E$ & $H$ & $T$ & $p$ & $\eta$ & $q_{1}$ & $q_{2}$ & $q_{3}$ & $q_{4}$ & $q_{5}$\\\midrule
\endhead
EP & Actor & 128 & 64 & 100 & 0.8 & 0.02 & 4 & false & false & -- & --\\
EP & Cora-CA & 128 & 64 & 100 & 0.8 & 0.02 & 4 & false & false & -- & --\\
EP & DBLP-CA & 128 & 64 & 100 & 0.8 & 0.02 & 4 & false & false & -- & --\\
EP & Cora & 128 & 64 & 100 & 0.8 & 0.001 & 4 & false & false & -- & --\\
EP & Pubmed & 128 & 64 & 100 & 0.8 & 0.02 & 4 & false & false & -- & --\\
EP & Pokec & 128 & 64 & 50 & 0.8 & 0.01 & 4 & false & false & -- & --\\
HC & IMDB-Dir-Form & 64 & 64 & 200 & 0.2 & 0.001 & 4 & false & false & max & true\\
HC & IMDB-Dir-Genre & 64 & 64 & 100 & 0.2 & 0.001 & 4 & false & false & max & true\\
HC & RHG-10 & 64 & 64 & 200 & 0.2 & 0.001 & 4 & false & false & mean & true\\
HC & RHG-3 & 64 & 64 & 100 & 0.2 & 0.001 & 4 & false & false & mean & true\\
HC & Steam-Player & 64 & 64 & 100 & 0.2 & 0.001 & 4 & false & false & max & true\\
HC & Twitter-Friend & 128 & 32 & 100 & 0.5 & 0.001 & 8 & true & true & -- & --\\
NC & ModelNet40 & 128 & 64 & 100 & 0.8 & 0.01 & 4 & false & true & -- & --\\
NC & NTU2012 & 128 & 64 & 100 & 0.5 & 0.01 & 4 & false & true & -- & --\\
NC & Actor & 128 & 64 & 100 & 0.8 & 0.02 & 4 & false & false & -- & --\\
NC & Cora-CA & 128 & 64 & 100 & 0.8 & 0.02 & 4 & false & false & -- & --\\
NC & DBLP-CA & 128 & 64 & 100 & 0.8 & 0.02 & 4 & false & false & -- & --\\
NC & Cora & 128 & 64 & 100 & 0.8 & 0.001 & 4 & false & false & -- & --\\
NC & Pubmed & 128 & 64 & 100 & 0.8 & 0.02 & 4 & false & false & -- & --\\
NC & Gamers & 128 & 64 & 100 & 0.8 & 0.02 & 4 & false & false & -- & --\\
NC & Pokec & 128 & 64 & 50 & 0.8 & 0.01 & 4 & false & false & -- & --\\
NC & Ratings & 128 & 64 & 100 & 0.8 & 0.05 & 4 & false & false & -- & --\\
NC & Trivago & 128 & 32 & 100 & 0.5 & 0.001 & 8 & true & true & -- & --\\
NC & Walmart & 128 & 64 & 100 & 0.8 & 0.02 & 4 & false & false & -- & --\\
NC & Yelp & 128 & 32 & 100 & 0.5 & 0.001 & 8 & true & true & -- & --\\
\bottomrule\end{longtable}
\noindent 
Additional columns: $q_{1}$=\texttt{heads}; $q_{2}$=\texttt{concat}; $q_{3}$=\texttt{use\_\allowbreak{}bn}; $q_{4}$=\texttt{pooling}; $q_{5}$=\texttt{early\_\allowbreak{}stop}.\par\smallskip
\noindent Shared settings: \texttt{All\_\allowbreak{}num\_\allowbreak{}layers}=2; \texttt{wd}=0.\par
\endgroup\medskip
\begingroup
\setlength{\tabcolsep}{3pt}
\setlength{\LTcapwidth}{\textwidth}
\renewcommand{\arraystretch}{1.08}
\begin{longtable}{@{}llrrrrrrrrr@{}}
\caption{DHG-Bench YAML settings for HGNN.}\label{tab:yaml-hgnn}\\
\toprule Task & Benchmark & $L$ & $E$ & $H$ & $T$ & $p$ & $\eta$ & $\lambda$ & $q_{1}$ & $q_{2}$\\\midrule
\endfirsthead
\toprule Task & Benchmark & $L$ & $E$ & $H$ & $T$ & $p$ & $\eta$ & $\lambda$ & $q_{1}$ & $q_{2}$\\\midrule
\endhead
EP & Actor & 1 & 128 & 64 & 400 & 0.5 & 0.01 & 0 & -- & --\\
EP & Cora-CA & 2 & 128 & 128 & 100 & 0.5 & 0.001 & 0 & -- & --\\
EP & DBLP-CA & 2 & 128 & 128 & 100 & 0.5 & 0.01 & 0 & -- & --\\
EP & Cora & 2 & 128 & 128 & 100 & 0.5 & 0.001 & 0 & -- & --\\
EP & Pubmed & 2 & 128 & 128 & 100 & 0.5 & 0.01 & 0 & -- & --\\
EP & Pokec & 2 & 128 & 128 & 200 & 0 & 0.01 & 0 & -- & --\\
HC & IMDB-Dir-Form & 1 & 64 & 64 & 200 & 0.2 & 0.01 & 0 & max & true\\
HC & IMDB-Dir-Genre & 1 & 128 & 64 & 200 & 0.2 & 0.01 & 0 & max & true\\
HC & RHG-10 & 1 & 64 & 64 & 300 & 0.2 & 0.001 & 0 & max & true\\
HC & RHG-3 & 1 & 64 & 64 & 200 & 0.2 & 0.001 & 0 & max & true\\
HC & Steam-Player & 1 & 64 & 64 & 200 & 0.2 & 0.001 & 0 & max & true\\
HC & Twitter-Friend & 1 & 64 & 64 & 200 & 0.2 & 0.001 & 0 & max & true\\
NC & ModelNet40 & 1 & 128 & 128 & 200 & 0.5 & 0.01 & 0 & -- & --\\
NC & NTU2012 & 1 & 128 & 128 & 200 & 0.5 & 0.01 & 0 & -- & --\\
NC & Actor & 1 & 128 & 64 & 400 & 0.5 & 0.01 & 0 & -- & --\\
NC & Cora-CA & 2 & 128 & 128 & 100 & 0.5 & 0.001 & 0 & -- & --\\
NC & DBLP-CA & 2 & 128 & 128 & 100 & 0.5 & 0.01 & 0 & -- & --\\
NC & Cora & 2 & 128 & 128 & 100 & 0.5 & 0.001 & 0 & -- & --\\
NC & Pubmed & 2 & 128 & 128 & 100 & 0.5 & 0.01 & 0 & -- & --\\
NC & Gamers & 1 & 128 & 64 & 200 & 0 & 0.01 & 0 & -- & --\\
NC & Pokec & 2 & 128 & 128 & 200 & 0 & 0.01 & 0 & -- & --\\
NC & Ratings & 2 & 128 & 128 & 100 & 0.5 & 0.001 & 0 & -- & --\\
NC & Trivago & 3 & 128 & 128 & 500 & 0.5 & 0.04 & 0 & -- & --\\
NC & Walmart & 2 & 128 & 128 & 200 & 0.5 & 0.01 & 0 & -- & --\\
NC & Yelp & 1 & 128 & 64 & 1000 & 0.5 & 0.01 & 0 & -- & --\\
\bottomrule\end{longtable}
\noindent 
Additional columns: $q_{1}$=\texttt{pooling}; $q_{2}$=\texttt{early\_\allowbreak{}stop}.\par\smallskip
\noindent Shared settings: \texttt{HCHA\_\allowbreak{}symdegnorm}=false.\par
\endgroup\medskip
\begingroup
\setlength{\tabcolsep}{3pt}
\setlength{\LTcapwidth}{\textwidth}
\renewcommand{\arraystretch}{1.08}
\begin{longtable}{@{}llrrrrrrrrrrrr@{}}
\caption{DHG-Bench YAML settings for HyperGCN.}\label{tab:yaml-hypergcn}\\
\toprule Task & Benchmark & $L$ & $E$ & $H$ & $T$ & $p$ & $\eta$ & $\lambda$ & $q_{1}$ & $q_{2}$ & $q_{3}$ & $q_{4}$ & $q_{5}$\\\midrule
\endfirsthead
\toprule Task & Benchmark & $L$ & $E$ & $H$ & $T$ & $p$ & $\eta$ & $\lambda$ & $q_{1}$ & $q_{2}$ & $q_{3}$ & $q_{4}$ & $q_{5}$\\\midrule
\endhead
EP & Actor & 1 & 128 & 64 & 400 & 0.5 & 0.1 & 0 & true & true & -- & -- & --\\
EP & Cora-CA & 2 & 128 & 128 & 200 & 0.8 & 0.001 & 0 & true & true & false & -- & --\\
EP & DBLP-CA & 2 & 128 & 128 & 200 & 0.8 & 0.001 & 0 & true & true & -- & -- & --\\
EP & Cora & 2 & 128 & 64 & 150 & 0.5 & 0.001 & 0 & true & true & -- & -- & --\\
EP & Pubmed & 2 & 128 & 256 & 200 & 0.5 & 0.01 & 0 & true & true & -- & -- & --\\
EP & Pokec & 2 & 128 & 64 & 200 & 0.5 & 0.001 & 0 & true & true & -- & -- & --\\
HC & IMDB-Dir-Form & 1 & 64 & 64 & 200 & 0.2 & 0.01 & 0 & true & false & -- & mean & true\\
HC & IMDB-Dir-Genre & 1 & 64 & 64 & 200 & 0.2 & 0.01 & 0 & true & false & -- & mean & true\\
HC & RHG-10 & 1 & 64 & 64 & 200 & 0.2 & 0.01 & 0 & true & false & -- & max & true\\
HC & RHG-3 & 1 & 64 & 64 & 200 & 0.2 & 0.01 & 0 & true & false & -- & max & true\\
HC & Steam-Player & 1 & 64 & 64 & 200 & 0.2 & 0.001 & 0 & true & false & -- & mean & true\\
HC & Twitter-Friend & 1 & 64 & 64 & 200 & 0.2 & 0.001 & 0 & true & false & -- & mean & true\\
NC & ModelNet40 & 2 & 128 & 128 & 500 & 0.5 & 0.001 & 0 & true & true & -- & -- & --\\
NC & NTU2012 & 2 & 128 & 128 & 500 & 0.5 & 0.001 & 0 & true & true & -- & -- & --\\
NC & Actor & 1 & 128 & 64 & 400 & 0.5 & 0.1 & 0 & true & true & -- & -- & --\\
NC & Cora-CA & 2 & 128 & 128 & 200 & 0.8 & 0.001 & 0 & true & true & false & -- & --\\
NC & DBLP-CA & 2 & 128 & 128 & 200 & 0.8 & 0.001 & 0 & true & true & -- & -- & --\\
NC & Cora & 2 & 128 & 64 & 150 & 0.5 & 0.001 & 0 & true & true & -- & -- & --\\
NC & Pubmed & 2 & 128 & 256 & 200 & 0.5 & 0.01 & 0 & true & true & -- & -- & --\\
NC & Gamers & 2 & 128 & 64 & 200 & 0.5 & 0.001 & 0 & true & true & -- & -- & --\\
NC & Pokec & 2 & 128 & 64 & 200 & 0.5 & 0.001 & 0 & true & true & -- & -- & --\\
NC & Ratings & 2 & 128 & 64 & 400 & 0.5 & 0.01 & 0 & true & true & -- & -- & --\\
NC & Trivago & 3 & 128 & 128 & 800 & 0.5 & 0.03 & 0 & true & true & -- & -- & --\\
NC & Walmart & 2 & 128 & 128 & 200 & 0.5 & 0.01 & 0 & true & true & -- & -- & --\\
NC & Yelp & 1 & 128 & 64 & 500 & 0.2 & 0.001 & 0 & false & true & -- & -- & --\\
\bottomrule\end{longtable}
\noindent 
Additional columns: $q_{1}$=\texttt{HyperGCN\_\allowbreak{}mediators}; $q_{2}$=\texttt{HyperGCN\_\allowbreak{}fast}; $q_{3}$=\texttt{HCHA\_\allowbreak{}symdegnorm}; $q_{4}$=\texttt{pooling}; $q_{5}$=\texttt{early\_\allowbreak{}stop}.\par\smallskip
\endgroup\medskip
\begingroup
\setlength{\tabcolsep}{3pt}
\setlength{\LTcapwidth}{\textwidth}
\renewcommand{\arraystretch}{1.08}
\begin{longtable}{@{}llrrrrrrrrr@{}}
\caption{DHG-Bench YAML settings for HCHA.}\label{tab:yaml-hcha}\\
\toprule Task & Benchmark & $L$ & $E$ & $H$ & $T$ & $p$ & $\eta$ & $\lambda$ & $q_{1}$ & $q_{2}$\\\midrule
\endfirsthead
\toprule Task & Benchmark & $L$ & $E$ & $H$ & $T$ & $p$ & $\eta$ & $\lambda$ & $q_{1}$ & $q_{2}$\\\midrule
\endhead
EP & Actor & 1 & 128 & 64 & 400 & 0.5 & 0.01 & 0 & -- & --\\
EP & Cora-CA & 2 & 128 & 128 & 100 & 0.5 & 0.001 & 0 & -- & --\\
EP & DBLP-CA & 2 & 128 & 128 & 50 & 0.5 & 0.01 & 0 & -- & --\\
EP & Cora & 2 & 128 & 128 & 100 & 0.5 & 0.001 & 0 & -- & --\\
EP & Pubmed & 2 & 128 & 128 & 100 & 0.5 & 0.01 & 0 & -- & --\\
EP & Pokec & 2 & 128 & 128 & 200 & 0 & 0.005 & 0 & -- & --\\
HC & IMDB-Dir-Form & 1 & 128 & 64 & 200 & 0.2 & 0.01 & 0 & max & true\\
HC & IMDB-Dir-Genre & 1 & 128 & 64 & 200 & 0.2 & 0.001 & 0 & max & true\\
HC & RHG-10 & 1 & 64 & 64 & 300 & 0.2 & 0.001 & 0 & max & true\\
HC & RHG-3 & 1 & 64 & 64 & 200 & 0.2 & 0.001 & 0 & max & true\\
HC & Steam-Player & 1 & 64 & 64 & 200 & 0.2 & 0.001 & 0 & max & true\\
HC & Twitter-Friend & 1 & 64 & 64 & 200 & 0.2 & 0.01 & 0 & max & true\\
NC & ModelNet40 & 1 & 128 & 128 & 200 & 0.5 & 0.01 & 0 & -- & --\\
NC & NTU2012 & 1 & 128 & 128 & 200 & 0.5 & 0.01 & 0 & -- & --\\
NC & Actor & 1 & 128 & 64 & 400 & 0.5 & 0.01 & 0 & -- & --\\
NC & Cora-CA & 2 & 128 & 128 & 100 & 0.5 & 0.001 & 0 & -- & --\\
NC & DBLP-CA & 2 & 128 & 128 & 50 & 0.5 & 0.01 & 0 & -- & --\\
NC & Cora & 2 & 128 & 128 & 100 & 0.5 & 0.001 & 0 & -- & --\\
NC & Pubmed & 2 & 128 & 128 & 100 & 0.5 & 0.01 & 0 & -- & --\\
NC & Gamers & 1 & 128 & 64 & 200 & 0 & 0.01 & 0 & -- & --\\
NC & Pokec & 2 & 128 & 128 & 200 & 0 & 0.005 & 0 & -- & --\\
NC & Ratings & 2 & 128 & 256 & 200 & 0.5 & 0.005 & 0 & -- & --\\
NC & Trivago & 3 & 128 & 128 & 500 & 0.5 & 0.04 & 0 & -- & --\\
NC & Walmart & 2 & 128 & 128 & 200 & 0.5 & 0.01 & 0 & -- & --\\
NC & Yelp & 1 & 128 & 64 & 500 & 0.5 & 0.01 & 0 & -- & --\\
\bottomrule\end{longtable}
\noindent 
Additional columns: $q_{1}$=\texttt{pooling}; $q_{2}$=\texttt{early\_\allowbreak{}stop}.\par\smallskip
\noindent Shared settings: \texttt{HCHA\_\allowbreak{}symdegnorm}=true.\par
\endgroup\medskip
\begingroup
\setlength{\tabcolsep}{3pt}
\setlength{\LTcapwidth}{\textwidth}
\renewcommand{\arraystretch}{1.08}
\begin{longtable}{@{}llrrrrrrrrr@{}}
\caption{DHG-Bench YAML settings for LEGCN.}\label{tab:yaml-legcn}\\
\toprule Task & Benchmark & $L$ & $E$ & $H$ & $T$ & $p$ & $\eta$ & $\lambda$ & $q_{1}$ & $q_{2}$\\\midrule
\endfirsthead
\toprule Task & Benchmark & $L$ & $E$ & $H$ & $T$ & $p$ & $\eta$ & $\lambda$ & $q_{1}$ & $q_{2}$\\\midrule
\endhead
EP & Actor & 2 & 128 & 128 & 100 & 0.1 & 0.01 & 0.0001 & -- & --\\
EP & Cora-CA & 2 & 128 & 64 & 100 & 0.1 & 0.01 & 0.01 & -- & --\\
EP & DBLP-CA & 2 & 128 & 64 & 100 & 0.1 & 0.01 & 0.01 & -- & --\\
EP & Cora & 2 & 128 & 256 & 100 & 0.1 & 0.01 & 0.01 & -- & --\\
EP & Pubmed & 2 & 128 & 256 & 100 & 0.5 & 0.01 & 0.0001 & -- & --\\
EP & Pokec & 4 & 128 & 256 & 200 & 0.1 & 0.001 & 0.0001 & -- & --\\
HC & IMDB-Dir-Form & 1 & 64 & 64 & 200 & 0.2 & 0.001 & 0 & mean & true\\
HC & IMDB-Dir-Genre & 2 & 64 & 64 & 200 & 0.2 & 0.01 & 0 & mean & true\\
HC & RHG-10 & 1 & 128 & 64 & 200 & 0.2 & 0.01 & 0 & mean & true\\
HC & RHG-3 & 1 & 64 & 64 & 200 & 0.2 & 0.01 & 0 & mean & true\\
HC & Steam-Player & 2 & 64 & 64 & 200 & 0.2 & 0.001 & 0 & mean & true\\
HC & Twitter-Friend & 1 & 64 & 64 & 200 & 0.2 & 0.001 & 0 & mean & true\\
NC & ModelNet40 & 1 & 128 & 128 & 300 & 0.5 & 0.01 & 0 & -- & --\\
NC & NTU2012 & 1 & 128 & 64 & 300 & 0.5 & 0.01 & 0 & -- & --\\
NC & Actor & 2 & 128 & 128 & 100 & 0.1 & 0.01 & 0.0001 & -- & --\\
NC & Cora-CA & 2 & 128 & 64 & 100 & 0.1 & 0.01 & 0.01 & -- & --\\
NC & DBLP-CA & 2 & 128 & 64 & 100 & 0.1 & 0.01 & 0.01 & -- & --\\
NC & Cora & 2 & 128 & 256 & 100 & 0.1 & 0.01 & 0.01 & -- & --\\
NC & Pubmed & 2 & 128 & 256 & 100 & 0.5 & 0.01 & 0.0001 & -- & --\\
NC & Gamers & 2 & 128 & 256 & 100 & 0.1 & 0.001 & 0.01 & -- & --\\
NC & Pokec & 4 & 128 & 256 & 200 & 0.1 & 0.001 & 0.0001 & -- & --\\
NC & Ratings & 3 & 128 & 256 & 50 & 0.1 & 0.01 & 0.01 & -- & --\\
NC & Trivago & 3 & 128 & 128 & 300 & 0.1 & 0.02 & 0 & -- & --\\
NC & Walmart & 3 & 128 & 128 & 100 & 0.1 & 0.01 & 0.01 & -- & --\\
NC & Yelp & 2 & 256 & 256 & 200 & 0.5 & 0.01 & 0 & -- & --\\
\bottomrule\end{longtable}
\noindent 
Additional columns: $q_{1}$=\texttt{pooling}; $q_{2}$=\texttt{early\_\allowbreak{}stop}.\par\smallskip
\endgroup\medskip
\begingroup
\setlength{\tabcolsep}{3pt}
\setlength{\LTcapwidth}{\textwidth}
\renewcommand{\arraystretch}{1.08}
\begin{longtable}{@{}llrrrrrrrrrrrrr@{}}
\caption{DHG-Bench YAML settings for HyperND.}\label{tab:yaml-hypernd}\\
\toprule Task & Benchmark & $E$ & $H$ & $T$ & $p$ & $\eta$ & $\lambda$ & $q_{1}$ & $q_{2}$ & $q_{3}$ & $q_{4}$ & $q_{5}$ & $q_{6}$ & $q_{7}$\\\midrule
\endfirsthead
\toprule Task & Benchmark & $E$ & $H$ & $T$ & $p$ & $\eta$ & $\lambda$ & $q_{1}$ & $q_{2}$ & $q_{3}$ & $q_{4}$ & $q_{5}$ & $q_{6}$ & $q_{7}$\\\midrule
\endhead
EP & Actor & 128 & 256 & 300 & 0.5 & 0.001 & 1e-05 & 1 & 0.001 & 0.8 & sum & -- & -- & --\\
EP & Cora-CA & 128 & 128 & 200 & 0.5 & 0.001 & 1e-05 & 1 & 0.001 & 0.3 & sum & -- & -- & --\\
EP & DBLP-CA & 128 & 128 & 500 & 0.5 & 0.001 & 1e-05 & 1 & 0.001 & 0.3 & sum & -- & -- & --\\
EP & Cora & 128 & 128 & 50 & 0.5 & 0.001 & 1e-05 & 2 & 1e-05 & 0.5 & sum & -- & -- & --\\
EP & Pubmed & 128 & 512 & 300 & 0.2 & 0.001 & 1e-05 & 1 & 0.001 & 0.2 & sum & -- & -- & --\\
EP & Pokec & 128 & 256 & 300 & 0.5 & 0.01 & 1e-05 & 1 & 0.001 & 0.6 & sum & -- & -- & --\\
HC & IMDB-Dir-Form & 64 & 64 & 200 & 0.2 & 0.01 & 0 & 1 & 0.001 & 0.9 & mean & mean & true & 128\\
HC & IMDB-Dir-Genre & 64 & 64 & 200 & 0.2 & 0.01 & 0 & 1 & 0.001 & 0.9 & mean & mean & true & 128\\
HC & RHG-10 & 64 & 64 & 200 & 0.2 & 0.01 & 0 & 1 & 0.001 & 0.9 & mean & mean & true & 128\\
HC & RHG-3 & 64 & 64 & 300 & 0.2 & 0.01 & 0 & 1 & 0.0001 & 0.9 & mean & mean & true & 128\\
HC & Steam-Player & 64 & 64 & 200 & 0.2 & 0.01 & 0 & 1 & 0.001 & 0.9 & mean & mean & true & 128\\
HC & Twitter-Friend & 64 & 64 & 200 & 0.2 & 0.01 & 0 & 1 & 0.001 & 0.9 & mean & mean & true & 128\\
NC & ModelNet40 & 128 & 64 & 200 & 0.5 & 0.01 & 0.001 & 2 & 0.001 & 0.5 & sum & -- & -- & --\\
NC & NTU2012 & 128 & 128 & 200 & 0.5 & 0.01 & 0.001 & 2 & 0.001 & 0.8 & sum & -- & -- & --\\
NC & Actor & 128 & 256 & 300 & 0.5 & 0.01 & 1e-05 & 1 & 0.001 & 0.8 & sum & -- & -- & --\\
NC & Cora-CA & 128 & 128 & 200 & 0.5 & 0.001 & 1e-05 & 1 & 0.001 & 0.3 & sum & -- & -- & --\\
NC & DBLP-CA & 128 & 128 & 500 & 0.5 & 0.001 & 1e-05 & 1 & 0.001 & 0.3 & sum & -- & -- & --\\
NC & Cora & 128 & 128 & 50 & 0.5 & 0.01 & 1e-05 & 2 & 1e-05 & 0.5 & sum & -- & -- & --\\
NC & Pubmed & 128 & 512 & 300 & 0.2 & 0.01 & 1e-05 & 1 & 0.001 & 0.2 & sum & -- & -- & --\\
NC & Gamers & 128 & 256 & 300 & 0.5 & 0.01 & 1e-05 & 1 & 0.001 & 0.4 & sum & -- & -- & --\\
NC & Pokec & 128 & 256 & 300 & 0.5 & 0.01 & 1e-05 & 1 & 0.001 & 0.6 & sum & -- & -- & --\\
NC & Ratings & 128 & 256 & 300 & 0.5 & 0.01 & 1e-05 & 1 & 0.001 & 0.2 & sum & -- & -- & --\\
NC & Trivago & 128 & 64 & 1000 & 0.2 & 0.004 & 0 & 1 & 0.0001 & 0 & sum & -- & -- & --\\
NC & Walmart & 128 & 256 & 600 & 0.5 & 0.01 & 1e-05 & 1 & 0.001 & 0.3 & sum & -- & -- & --\\
NC & Yelp & 128 & 128 & 500 & 0.5 & 0.001 & 1e-05 & 1 & 0.001 & 0.3 & sum & -- & -- & --\\
\bottomrule\end{longtable}
\noindent 
Additional columns: $q_{1}$=\texttt{HyperND\_\allowbreak{}ord}; $q_{2}$=\texttt{HyperND\_\allowbreak{}tol}; $q_{3}$=\texttt{restart\_\allowbreak{}alpha}; $q_{4}$=\texttt{aggregate}; $q_{5}$=\texttt{pooling}; $q_{6}$=\texttt{early\_\allowbreak{}stop}; $q_{7}$=\texttt{hg\_\allowbreak{}batch\_\allowbreak{}size}.\par\smallskip
\noindent Shared settings: \texttt{MLP\_\allowbreak{}num\_\allowbreak{}layers}=2; \texttt{HyperND\_\allowbreak{}steps}=100; \texttt{normalization}=ln.\par
\endgroup\medskip
\begingroup
\setlength{\tabcolsep}{3pt}
\setlength{\LTcapwidth}{\textwidth}
\renewcommand{\arraystretch}{1.08}
\begin{longtable}{@{}llrrrrrrrrrrrr@{}}
\caption{DHG-Bench YAML settings for PhenomNN.}\label{tab:yaml-phenomnn}\\
\toprule Task & Benchmark & $E$ & $H$ & $T$ & $p$ & $\eta$ & $\lambda$ & $q_{1}$ & $q_{2}$ & $q_{3}$ & $q_{4}$ & $q_{5}$ & $q_{6}$\\\midrule
\endfirsthead
\toprule Task & Benchmark & $E$ & $H$ & $T$ & $p$ & $\eta$ & $\lambda$ & $q_{1}$ & $q_{2}$ & $q_{3}$ & $q_{4}$ & $q_{5}$ & $q_{6}$\\\midrule
\endhead
EP & Actor & 128 & 128 & 200 & 0.5 & 0.001 & 1e-05 & 0 & 4 & 2 & 128 & -- & --\\
EP & Cora-CA & 128 & 64 & 50 & 0.5 & 0.001 & 1e-05 & 0.1 & 16 & 1 & 128 & -- & --\\
EP & DBLP-CA & 128 & 64 & 200 & 0.5 & 0.001 & 1e-05 & 0.1 & 16 & 1 & 128 & -- & --\\
EP & Cora & 128 & 64 & 50 & 0.5 & 0.001 & 1e-05 & 0.5 & 16 & 1 & 128 & -- & --\\
EP & Pubmed & 128 & 64 & 300 & 0.5 & 0.001 & 1e-05 & 0.2 & 16 & 1 & 128 & -- & --\\
EP & Pokec & 128 & 64 & 200 & 0.5 & 0.01 & 1e-05 & 0 & 8 & 2 & 128 & -- & --\\
HC & IMDB-Dir-Form & 64 & 64 & 200 & 0.2 & 0.01 & 0 & 0.3 & 2 & 1 & 64 & mean & true\\
HC & IMDB-Dir-Genre & 64 & 64 & 200 & 0.2 & 0.01 & 0 & 0.1 & 2 & 1 & 64 & mean & true\\
HC & RHG-10 & 64 & 64 & 200 & 0.2 & 0.01 & 0 & 0.1 & 1 & 1 & 64 & mean & true\\
HC & RHG-3 & 64 & 64 & 200 & 0.2 & 0.01 & 0 & 0.1 & 2 & 1 & 64 & mean & true\\
HC & Steam-Player & 64 & 64 & 200 & 0.2 & 0.01 & 0 & 0.3 & 2 & 1 & 64 & max & true\\
HC & Twitter-Friend & 64 & 64 & 200 & 0.2 & 0.001 & 0 & 0.1 & 2 & 1 & 64 & max & true\\
NC & ModelNet40 & 128 & 256 & 100 & 0.5 & 0.01 & 0 & 0.1 & 8 & 1 & 128 & -- & --\\
NC & NTU2012 & 128 & 256 & 200 & 0.5 & 0.01 & 0 & 0.1 & 8 & 1 & 128 & -- & --\\
NC & Actor & 128 & 128 & 200 & 0.5 & 0.01 & 1e-05 & 0 & 4 & 2 & 128 & -- & --\\
NC & Cora-CA & 128 & 64 & 50 & 0.5 & 0.01 & 1e-05 & 0.1 & 16 & 1 & 128 & -- & --\\
NC & DBLP-CA & 128 & 64 & 200 & 0.5 & 0.01 & 1e-05 & 0.1 & 16 & 1 & 128 & -- & --\\
NC & Cora & 128 & 64 & 50 & 0.5 & 0.01 & 1e-05 & 0.5 & 16 & 1 & 128 & -- & --\\
NC & Pubmed & 128 & 64 & 300 & 0.5 & 0.01 & 1e-05 & 0.2 & 16 & 1 & 128 & -- & --\\
NC & Gamers & 128 & 128 & 100 & 0.5 & 0.01 & 1e-05 & 0 & 8 & 2 & 128 & -- & --\\
NC & Pokec & 128 & 64 & 200 & 0.5 & 0.01 & 1e-05 & 0 & 8 & 2 & 128 & -- & --\\
NC & Ratings & 128 & 128 & 200 & 0.5 & 0.01 & 1e-05 & 0 & 4 & 2 & 128 & -- & --\\
NC & Trivago & 128 & 128 & 100 & 0.5 & 0.001 & 0 & 0.1 & 16 & 1 & 128 & -- & --\\
NC & Walmart & 128 & 64 & 200 & 0.5 & 0.01 & 1e-05 & 0.1 & 16 & 1 & 128 & -- & --\\
NC & Yelp & 128 & 64 & 200 & 0.5 & 0.01 & 1e-05 & 0.1 & 16 & 1 & 128 & -- & --\\
\bottomrule\end{longtable}
\noindent 
Additional columns: $q_{1}$=\texttt{alpha}; $q_{2}$=\texttt{prop\_\allowbreak{}step}; $q_{3}$=\texttt{encoder\_\allowbreak{}num\_\allowbreak{}layers}; $q_{4}$=\texttt{decoder\_\allowbreak{}hidden}; $q_{5}$=\texttt{pooling}; $q_{6}$=\texttt{early\_\allowbreak{}stop}.\par\smallskip
\noindent Shared settings: \texttt{lam0}=10; \texttt{lam1}=10; \texttt{normalization}=ln; \texttt{decoder\_\allowbreak{}num\_\allowbreak{}layers}=1.\par
\endgroup\medskip
\begingroup
\setlength{\tabcolsep}{3pt}
\setlength{\LTcapwidth}{\textwidth}
\renewcommand{\arraystretch}{1.08}
\begin{longtable}{@{}llrrrrrrrrrrrr@{}}
\caption{DHG-Bench YAML settings for SheafHyperGNN.}\label{tab:yaml-sheafhypergnn}\\
\toprule Task & Benchmark & $L$ & $E$ & $H$ & $T$ & $p$ & $\eta$ & $\lambda$ & $q_{1}$ & $q_{2}$ & $q_{3}$ & $q_{4}$ & $q_{5}$\\\midrule
\endfirsthead
\toprule Task & Benchmark & $L$ & $E$ & $H$ & $T$ & $p$ & $\eta$ & $\lambda$ & $q_{1}$ & $q_{2}$ & $q_{3}$ & $q_{4}$ & $q_{5}$\\\midrule
\endhead
EP & Actor & 1 & 128 & 256 & 200 & 0.5 & 0.001 & 0 & avg & sym\_\allowbreak{}degree\_\allowbreak{}norm & 4 & -- & --\\
EP & Cora-CA & 4 & 128 & 256 & 15 & 0.5 & 0.001 & 0 & avg & sym\_\allowbreak{}degree\_\allowbreak{}norm & 4 & -- & --\\
EP & DBLP-CA & 4 & 128 & 256 & 20 & 0.5 & 0.001 & 0 & avg & sym\_\allowbreak{}degree\_\allowbreak{}norm & 4 & -- & --\\
EP & Cora & 3 & 128 & 512 & 10 & 0.5 & 0.001 & 0 & avg & sym\_\allowbreak{}degree\_\allowbreak{}norm & 4 & -- & --\\
EP & Pubmed & 3 & 128 & 256 & 80 & 0.5 & 0.001 & 0 & avg & sym\_\allowbreak{}degree\_\allowbreak{}norm & 4 & -- & --\\
EP & Pokec & 1 & 128 & 256 & 200 & 0.5 & 0.001 & 0 & avg & sym\_\allowbreak{}degree\_\allowbreak{}norm & 4 & -- & --\\
HC & IMDB-Dir-Form & 1 & 64 & 64 & 200 & 0.2 & 0.01 & 0 & avg & sym\_\allowbreak{}degree\_\allowbreak{}norm & 4 & mean & true\\
HC & IMDB-Dir-Genre & 1 & 64 & 64 & 200 & 0.2 & 0.01 & 0 & avg & sym\_\allowbreak{}degree\_\allowbreak{}norm & 4 & mean & true\\
HC & RHG-10 & 1 & 64 & 64 & 200 & 0.2 & 0.01 & 0 & avg & sym\_\allowbreak{}degree\_\allowbreak{}norm & 4 & max & true\\
HC & RHG-3 & 1 & 64 & 64 & 200 & 0.2 & 0.01 & 0 & avg & sym\_\allowbreak{}degree\_\allowbreak{}norm & 4 & mean & true\\
HC & Steam-Player & 1 & 64 & 64 & 200 & 0.2 & 0.01 & 0 & avg & sym\_\allowbreak{}degree\_\allowbreak{}norm & 4 & max & true\\
HC & Twitter-Friend & 1 & 64 & 64 & 200 & 0.2 & 0.01 & 0 & avg & sym\_\allowbreak{}degree\_\allowbreak{}norm & 4 & max & true\\
NC & ModelNet40 & 1 & 128 & 512 & 100 & 0.5 & 0.001 & 0 & avg & sym\_\allowbreak{}degree\_\allowbreak{}norm & 4 & -- & --\\
NC & NTU2012 & 1 & 128 & 512 & 100 & 0.5 & 0.001 & 0 & avg & sym\_\allowbreak{}degree\_\allowbreak{}norm & 4 & -- & --\\
NC & Actor & 1 & 128 & 256 & 200 & 0.5 & 0.001 & 0 & avg & sym\_\allowbreak{}degree\_\allowbreak{}norm & 4 & -- & --\\
NC & Cora-CA & 4 & 128 & 256 & 15 & 0.5 & 0.001 & 0 & avg & sym\_\allowbreak{}degree\_\allowbreak{}norm & 4 & -- & --\\
NC & DBLP-CA & 4 & 128 & 256 & 20 & 0.5 & 0.001 & 0 & avg & sym\_\allowbreak{}degree\_\allowbreak{}norm & 4 & -- & --\\
NC & Cora & 3 & 128 & 512 & 10 & 0.5 & 0.001 & 0 & avg & sym\_\allowbreak{}degree\_\allowbreak{}norm & 4 & -- & --\\
NC & Pubmed & 3 & 128 & 256 & 80 & 0.5 & 0.001 & 0 & avg & sym\_\allowbreak{}degree\_\allowbreak{}norm & 4 & -- & --\\
NC & Gamers & 2 & 128 & 256 & 100 & 0.5 & 0.001 & 0 & avg & sym\_\allowbreak{}degree\_\allowbreak{}norm & 4 & -- & --\\
NC & Pokec & 1 & 128 & 256 & 200 & 0.5 & 0.001 & 0 & avg & sym\_\allowbreak{}degree\_\allowbreak{}norm & 4 & -- & --\\
NC & Ratings & 1 & 128 & 256 & 80 & 0.5 & 0.001 & 0 & avg & sym\_\allowbreak{}degree\_\allowbreak{}norm & 4 & -- & --\\
NC & Trivago & 2 & 128 & 128 & 100 & 0.5 & 0.001 & 0 & rand & degree\_\allowbreak{}norm & 2 & -- & --\\
NC & Walmart & 2 & 128 & 128 & 100 & 0.5 & 0.001 & 0 & rand & degree\_\allowbreak{}norm & 2 & -- & --\\
NC & Yelp & 2 & 128 & 128 & 100 & 0.5 & 0.001 & 0 & rand & degree\_\allowbreak{}norm & 2 & -- & --\\
\bottomrule\end{longtable}
\noindent 
Additional columns: $q_{1}$=\texttt{init\_\allowbreak{}hedge}; $q_{2}$=\texttt{sheaf\_\allowbreak{}normtype}; $q_{3}$=\texttt{stalk\_\allowbreak{}dim}; $q_{4}$=\texttt{pooling}; $q_{5}$=\texttt{early\_\allowbreak{}stop}.\par\smallskip
\noindent Shared settings: \texttt{sheaf\_\allowbreak{}act}=sigmoid; \texttt{sheaf\_\allowbreak{}dropout}=false; \texttt{sheaf\_\allowbreak{}left\_\allowbreak{}proj}=false; \texttt{dynamic\_\allowbreak{}sheaf}=true; \texttt{sheaf\_\allowbreak{}special\_\allowbreak{}head}=false; \texttt{sheaf\_\allowbreak{}pred\_\allowbreak{}block}=MLP\_\allowbreak{}var1; \texttt{sheaf\_\allowbreak{}transformer\_\allowbreak{}head}=1; \texttt{AllSet\_\allowbreak{}input\_\allowbreak{}norm}=true; \texttt{residual\_\allowbreak{}sheaf}=false.\par
\endgroup\medskip
\begingroup
\setlength{\tabcolsep}{3pt}
\setlength{\LTcapwidth}{\textwidth}
\renewcommand{\arraystretch}{1.08}
\begin{longtable}{@{}llrrrrrrrrrr@{}}
\caption{DHG-Bench YAML settings for HJRL.}\label{tab:yaml-hjrl}\\
\toprule Task & Benchmark & $L$ & $E$ & $H$ & $T$ & $p$ & $\eta$ & $q_{1}$ & $q_{2}$ & $q_{3}$ & $q_{4}$\\\midrule
\endfirsthead
\toprule Task & Benchmark & $L$ & $E$ & $H$ & $T$ & $p$ & $\eta$ & $q_{1}$ & $q_{2}$ & $q_{3}$ & $q_{4}$\\\midrule
\endhead
EP & Actor & 2 & 128 & 128 & 500 & 0.5 & 0.01 & 0.1 & false & -- & --\\
EP & Cora-CA & 2 & 128 & 128 & 200 & 0.5 & 0.001 & 1 & false & -- & --\\
EP & DBLP-CA & 2 & 128 & 128 & 100 & 0.5 & 0.001 & 0.1 & 0.1 & -- & --\\
EP & Cora & 2 & 128 & 256 & 100 & 0.5 & 0.001 & 0.01 & false & -- & --\\
EP & Pubmed & 2 & 128 & 128 & 100 & 0.5 & 0.01 & 0.1 & false & -- & --\\
EP & Pokec & 2 & 128 & 128 & 500 & 0.1 & 0.001 & 0 & false & -- & --\\
HC & IMDB-Dir-Form & 1 & 64 & 64 & 200 & 0.2 & 0.01 & 0 & false & max & true\\
HC & IMDB-Dir-Genre & 1 & 64 & 64 & 200 & 0.2 & 0.01 & 0 & false & max & true\\
HC & RHG-10 & 1 & 64 & 64 & 200 & 0.2 & 0.01 & 0 & false & max & true\\
HC & RHG-3 & 1 & 64 & 64 & 200 & 0.2 & 0.01 & 0 & false & max & true\\
HC & Steam-Player & 1 & 64 & 64 & 200 & 0.2 & 0.001 & 0 & false & max & true\\
HC & Twitter-Friend & 1 & 64 & 64 & 200 & 0.2 & 0.001 & 0 & false & max & true\\
NC & ModelNet40 & 2 & 128 & 128 & 300 & 0.5 & 0.1 & 0.01 & false & -- & --\\
NC & NTU2012 & 2 & 128 & 128 & 300 & 0.5 & 0.1 & 0.01 & false & -- & --\\
NC & Actor & 2 & 128 & 128 & 500 & 0.5 & 0.01 & 0.1 & false & -- & --\\
NC & Cora-CA & 2 & 128 & 128 & 200 & 0.5 & 0.001 & 1 & false & -- & --\\
NC & DBLP-CA & 2 & 128 & 128 & 100 & 0.5 & 0.001 & 0.1 & 0.1 & -- & --\\
NC & Cora & 2 & 128 & 256 & 100 & 0.5 & 0.001 & 0.01 & false & -- & --\\
NC & Pubmed & 2 & 128 & 128 & 100 & 0.5 & 0.01 & 0.1 & false & -- & --\\
NC & Gamers & 2 & 128 & 256 & 200 & 0.1 & 0.001 & 0.1 & false & -- & --\\
NC & Pokec & 2 & 128 & 128 & 500 & 0.1 & 0.001 & 0 & false & -- & --\\
NC & Ratings & 2 & 128 & 256 & 200 & 0.1 & 0.001 & 0 & false & -- & --\\
NC & Trivago & 2 & 128 & 128 & 100 & 0.5 & 0.001 & 0.1 & 0.1 & -- & --\\
NC & Walmart & 2 & 128 & 128 & 100 & 0.5 & 0.001 & 0.1 & 0.1 & -- & --\\
NC & Yelp & 2 & 128 & 128 & 100 & 0.5 & 0.001 & 0.1 & 0.1 & -- & --\\
\bottomrule\end{longtable}
\noindent 
Additional columns: $q_{1}$=\texttt{gamma}; $q_{2}$=\texttt{sample\_\allowbreak{}ratio}; $q_{3}$=\texttt{pooling}; $q_{4}$=\texttt{early\_\allowbreak{}stop}.\par\smallskip
\noindent Shared settings: \texttt{wd}=0; \texttt{activation}=relu; \texttt{neg\_\allowbreak{}slope}=0.01; \texttt{weight\_\allowbreak{}init}=1; \texttt{pos\_\allowbreak{}weight\_\allowbreak{}thresh}=1.0e4.\par
\endgroup\medskip
\begingroup
\setlength{\tabcolsep}{3pt}
\setlength{\LTcapwidth}{\textwidth}
\renewcommand{\arraystretch}{1.08}
\begin{longtable}{@{}llrrrrrrrrrrr@{}}
\caption{DHG-Bench YAML settings for DPHGNN.}\label{tab:yaml-dphgnn}\\
\toprule Task & Benchmark & $L$ & $E$ & $H$ & $T$ & $p$ & $\eta$ & $q_{1}$ & $q_{2}$ & $q_{3}$ & $q_{4}$ & $q_{5}$\\\midrule
\endfirsthead
\toprule Task & Benchmark & $L$ & $E$ & $H$ & $T$ & $p$ & $\eta$ & $q_{1}$ & $q_{2}$ & $q_{3}$ & $q_{4}$ & $q_{5}$\\\midrule
\endhead
EP & Actor & 1 & 128 & 256 & 100 & 0.5 & 0.01 & 256 & 1 & 1 & -- & --\\
EP & Cora-CA & 1 & 128 & 256 & 100 & 0.5 & 0.01 & 256 & 1 & 1 & -- & --\\
EP & DBLP-CA & 1 & 128 & 256 & 100 & 0.5 & 0.01 & 256 & 1 & 1 & -- & --\\
EP & Cora & 1 & 128 & 256 & 100 & 0.5 & 0.01 & 256 & 2 & 1 & -- & --\\
EP & Pubmed & 1 & 128 & 256 & 100 & 0.5 & 0.01 & 256 & 1 & 1 & -- & --\\
EP & Pokec & 1 & 128 & 256 & 100 & 0.5 & 0.01 & 256 & 2 & 2 & -- & --\\
HC & IMDB-Dir-Form & 1 & 64 & 64 & 200 & 0.2 & 0.001 & 128 & 2 & 1 & max & true\\
HC & IMDB-Dir-Genre & 1 & 64 & 64 & 200 & 0.2 & 0.001 & 128 & 2 & 1 & mean & true\\
HC & RHG-10 & 1 & 64 & 64 & 200 & 0.2 & 0.001 & 128 & 2 & 1 & max & true\\
HC & RHG-3 & 1 & 64 & 64 & 200 & 0.2 & 0.001 & 128 & 2 & 1 & max & true\\
HC & Steam-Player & 1 & 64 & 64 & 200 & 0.2 & 0.001 & 128 & 2 & 1 & max & true\\
HC & Twitter-Friend & 1 & 64 & 64 & 200 & 0.2 & 0.001 & 128 & 2 & 1 & max & true\\
NC & ModelNet40 & 1 & 128 & 512 & 300 & 0.5 & 0.01 & 256 & 2 & 1 & -- & --\\
NC & NTU2012 & 1 & 128 & 512 & 200 & 0.5 & 0.01 & 256 & 2 & 1 & -- & --\\
NC & Actor & 1 & 128 & 256 & 100 & 0.5 & 0.01 & 256 & 1 & 1 & -- & --\\
NC & Cora-CA & 1 & 128 & 256 & 100 & 0.5 & 0.01 & 256 & 1 & 1 & -- & --\\
NC & DBLP-CA & 1 & 128 & 256 & 100 & 0.5 & 0.01 & 256 & 1 & 1 & -- & --\\
NC & Cora & 1 & 128 & 256 & 100 & 0.5 & 0.01 & 256 & 2 & 1 & -- & --\\
NC & Pubmed & 1 & 128 & 256 & 100 & 0.5 & 0.01 & 256 & 1 & 1 & -- & --\\
NC & Gamers & 1 & 128 & 256 & 80 & 0.5 & 0.01 & 256 & 1 & 1 & -- & --\\
NC & Pokec & 1 & 128 & 256 & 100 & 0.5 & 0.01 & 256 & 2 & 2 & -- & --\\
NC & Ratings & 1 & 128 & 256 & 100 & 0.5 & 0.01 & 256 & 1 & 1 & -- & --\\
NC & Trivago & 2 & 128 & 128 & 100 & 0.5 & 0.001 & 128 & 2 & 1 & -- & --\\
NC & Walmart & 2 & 128 & 128 & 100 & 0.5 & 0.001 & 128 & 2 & 1 & -- & --\\
NC & Yelp & 2 & 128 & 128 & 100 & 0.5 & 0.001 & 128 & 2 & 1 & -- & --\\
\bottomrule\end{longtable}
\noindent 
Additional columns: $q_{1}$=\texttt{expan\_\allowbreak{}dim}; $q_{2}$=\texttt{cg\_\allowbreak{}num\_\allowbreak{}layer}; $q_{3}$=\texttt{dff\_\allowbreak{}num\_\allowbreak{}layers}; $q_{4}$=\texttt{pooling}; $q_{5}$=\texttt{early\_\allowbreak{}stop}.\par\smallskip
\noindent Shared settings: \texttt{wd}=0; \texttt{cg\_\allowbreak{}MLP\_\allowbreak{}hidden}=64; \texttt{cg\_\allowbreak{}dropout}=0.5; \texttt{hg\_\allowbreak{}num\_\allowbreak{}layer}=2; \texttt{hg\_\allowbreak{}MLP\_\allowbreak{}hidden}=64; \texttt{hg\_\allowbreak{}dropout}=0.5; \texttt{mediator}=true; \texttt{sg\_\allowbreak{}num\_\allowbreak{}layer}=2; \texttt{sg\_\allowbreak{}MLP\_\allowbreak{}hidden}=64; \texttt{sg\_\allowbreak{}dropout}=0.5; \texttt{taa\_\allowbreak{}spatial\_\allowbreak{}dim}=64; \texttt{taa\_\allowbreak{}spectral\_\allowbreak{}dim}=64; \texttt{num\_\allowbreak{}heads}=1; \texttt{chunk\_\allowbreak{}size}=-1; \texttt{spectral\_\allowbreak{}embed\_\allowbreak{}dim}=64; \texttt{fc\_\allowbreak{}dim}=64; \texttt{dff\_\allowbreak{}MLP\_\allowbreak{}hidden}=64; \texttt{atten\_\allowbreak{}neg\_\allowbreak{}slope}=0.2.\par
\endgroup\medskip
\begingroup
\setlength{\tabcolsep}{3pt}
\setlength{\LTcapwidth}{\textwidth}
\renewcommand{\arraystretch}{1.08}
\begin{longtable}{@{}llrrrrrrrrrrrr@{}}
\caption{DHG-Bench YAML settings for TF-HNN.}\label{tab:yaml-tfhnn}\\
\toprule Task & Benchmark & $L$ & $E$ & $H$ & $M$ & $T$ & $p$ & $\eta$ & $\lambda$ & $q_{1}$ & $q_{2}$ & $q_{3}$ & $q_{4}$\\\midrule
\endfirsthead
\toprule Task & Benchmark & $L$ & $E$ & $H$ & $M$ & $T$ & $p$ & $\eta$ & $\lambda$ & $q_{1}$ & $q_{2}$ & $q_{3}$ & $q_{4}$\\\midrule
\endhead
EP & Actor & 2 & 128 & 512 & 3 & 100 & 0.1 & 0.001 & 0 & 1 & -- & -- & --\\
EP & Cora-CA & 2 & 128 & 1024 & 3 & 60 & 0.7 & 0.001 & 0 & 0.3 & -- & -- & --\\
EP & DBLP-CA & 2 & 128 & 1024 & 3 & 50 & 0.8 & 0.001 & 0 & 0.3 & -- & -- & --\\
EP & Cora & 2 & 128 & 256 & 3 & 30 & 0.7 & 0.001 & 0 & 0.1 & -- & -- & --\\
EP & Pubmed & 2 & 128 & 512 & 3 & 200 & 0.7 & 0.001 & 0 & 0.05 & -- & -- & --\\
EP & Pokec & 2 & 128 & 1024 & 3 & 150 & 0.5 & 0.01 & 0.0001 & 0.9 & -- & -- & --\\
HC & IMDB-Dir-Form & 1 & 64 & 64 & 2 & 200 & 0.2 & 0.01 & 0 & 0.1 & mean & true & 128\\
HC & IMDB-Dir-Genre & 1 & 64 & 64 & 2 & 200 & 0.2 & 0.01 & 0 & 0.1 & mean & true & 128\\
HC & RHG-10 & 1 & 64 & 64 & 2 & 200 & 0.2 & 0.01 & 0 & 0.1 & mean & true & 128\\
HC & RHG-3 & 1 & 64 & 64 & 2 & 200 & 0.2 & 0.01 & 0 & 0.1 & mean & true & 128\\
HC & Steam-Player & 1 & 64 & 64 & 2 & 200 & 0.2 & 0.01 & 0 & 0.9 & max & true & 128\\
HC & Twitter-Friend & 1 & 64 & 64 & 2 & 200 & 0.2 & 0.01 & 0 & 0.9 & max & true & 128\\
NC & ModelNet40 & 1 & 128 & 512 & 2 & 300 & 0.7 & 0.001 & 0 & 0.1 & -- & -- & --\\
NC & NTU2012 & 1 & 128 & 512 & 2 & 300 & 0.7 & 0.001 & 0 & 0.1 & -- & -- & --\\
NC & Actor & 2 & 128 & 512 & 3 & 100 & 0.1 & 0.01 & 0 & 1 & -- & -- & --\\
NC & Cora-CA & 2 & 128 & 1024 & 3 & 60 & 0.7 & 0.001 & 0 & 0.3 & -- & -- & --\\
NC & DBLP-CA & 2 & 128 & 1024 & 3 & 50 & 0.8 & 0.0006 & 0 & 0.3 & -- & -- & --\\
NC & Cora & 2 & 128 & 256 & 3 & 30 & 0.7 & 0.001 & 0 & 0.1 & -- & -- & --\\
NC & Pubmed & 2 & 128 & 512 & 3 & 200 & 0.7 & 0.001 & 0 & 0.05 & -- & -- & --\\
NC & Gamers & 2 & 128 & 1024 & 3 & 150 & 0.5 & 0.01 & 0.0001 & 0.1 & -- & -- & --\\
NC & Pokec & 2 & 128 & 1024 & 3 & 150 & 0.5 & 0.01 & 0.0001 & 0.9 & -- & -- & --\\
NC & Ratings & 2 & 128 & 128 & 3 & 190 & 0.5 & 0.01 & 0 & 0.1 & -- & -- & --\\
NC & Trivago & 128 & 128 & 64 & 3 & 1000 & 0.2 & 0.004 & 0 & 0 & -- & -- & --\\
NC & Walmart & 2 & 128 & 512 & 3 & 100 & 0.7 & 0.01 & 0 & 0 & -- & -- & --\\
NC & Yelp & 2 & 128 & 64 & 3 & 500 & 0.2 & 0.001 & 0 & 0 & -- & -- & --\\
\bottomrule\end{longtable}
\noindent 
Additional columns: $q_{1}$=\texttt{alpha}; $q_{2}$=\texttt{pooling}; $q_{3}$=\texttt{early\_\allowbreak{}stop}; $q_{4}$=\texttt{hg\_\allowbreak{}batch\_\allowbreak{}size}.\par\smallskip
\noindent Shared settings: \texttt{normalization}=ln.\par
\endgroup\medskip
\begingroup
\setlength{\tabcolsep}{3pt}
\setlength{\LTcapwidth}{\textwidth}
\renewcommand{\arraystretch}{1.08}
\begin{longtable}{@{}llrrrrrrrrrrrr@{}}
\caption{DHG-Bench YAML settings for HNHN.}\label{tab:yaml-hnhn}\\
\toprule Task & Benchmark & $L$ & $E$ & $H$ & $T$ & $p$ & $\eta$ & $\lambda$ & $q_{1}$ & $q_{2}$ & $q_{3}$ & $q_{4}$ & $q_{5}$\\\midrule
\endfirsthead
\toprule Task & Benchmark & $L$ & $E$ & $H$ & $T$ & $p$ & $\eta$ & $\lambda$ & $q_{1}$ & $q_{2}$ & $q_{3}$ & $q_{4}$ & $q_{5}$\\\midrule
\endhead
EP & Actor & 1 & 256 & 128 & 300 & 0.5 & 0.001 & 0.0001 & -1.5 & -0.5 & true & -- & --\\
EP & Cora-CA & 1 & 256 & 256 & 100 & 0.5 & 0.001 & 1e-05 & -1.5 & -0.5 & true & -- & --\\
EP & DBLP-CA & 1 & 256 & 256 & 100 & 0.5 & 0.001 & 0.0001 & -1.5 & -0.5 & true & -- & --\\
EP & Cora & 1 & 256 & 128 & 200 & 0.5 & 0.001 & 0.0001 & -1.5 & -0.5 & true & -- & --\\
EP & Pubmed & 1 & 256 & 512 & 50 & 0.5 & 0.001 & 0.0001 & -1.5 & -0.5 & false & -- & --\\
EP & Pokec & 1 & 256 & 128 & 200 & 0.5 & 0.01 & 0.0001 & -1.5 & -0.5 & true & -- & --\\
HC & IMDB-Dir-Form & 1 & 128 & 64 & 200 & 0.2 & 0.01 & 0 & -1.5 & -0.5 & true & max & true\\
HC & IMDB-Dir-Genre & 1 & 128 & 64 & 200 & 0.2 & 0.01 & 0 & -1.5 & -0.5 & true & mean & true\\
HC & RHG-10 & 1 & 128 & 64 & 200 & 0.2 & 0.01 & 0 & -1.5 & -0.5 & true & max & true\\
HC & RHG-3 & 1 & 64 & 64 & 200 & 0.2 & 0.01 & 0 & -1.5 & -0.5 & true & max & true\\
HC & Steam-Player & 1 & 128 & 64 & 200 & 0.2 & 0.01 & 0 & -1.5 & -0.5 & true & max & true\\
HC & Twitter-Friend & 1 & 64 & 64 & 200 & 0.2 & 0.01 & 0 & -1.5 & -0.5 & true & max & true\\
NC & ModelNet40 & 1 & 256 & 128 & 400 & 0.5 & 0.1 & 0 & -1.5 & -0.5 & true & -- & --\\
NC & NTU2012 & 1 & 256 & 256 & 400 & 0.8 & 0.1 & 0 & -1.5 & -0.5 & true & -- & --\\
NC & Actor & 1 & 256 & 128 & 300 & 0.5 & 0.01 & 0.0001 & -1.5 & -0.5 & true & -- & --\\
NC & Cora-CA & 1 & 256 & 256 & 100 & 0.5 & 0.01 & 1e-05 & -1.5 & -0.5 & true & -- & --\\
NC & DBLP-CA & 1 & 256 & 256 & 100 & 0.5 & 0.01 & 0.0001 & -1.5 & -0.5 & true & -- & --\\
NC & Cora & 1 & 256 & 128 & 200 & 0.5 & 0.01 & 0.0001 & -1.5 & -0.5 & true & -- & --\\
NC & Pubmed & 1 & 256 & 512 & 50 & 0.5 & 0.01 & 0.0001 & -1.5 & -0.5 & false & -- & --\\
NC & Gamers & 1 & 256 & 128 & 200 & 0.5 & 0.001 & 0.0001 & -1.5 & -0.5 & true & -- & --\\
NC & Pokec & 1 & 256 & 128 & 200 & 0.5 & 0.01 & 0.0001 & -1.5 & -0.5 & true & -- & --\\
NC & Ratings & 1 & 256 & 128 & 50 & 0.5 & 0.001 & 0.0001 & -1.5 & -0.5 & true & -- & --\\
NC & Trivago & 2 & 256 & 128 & 1000 & 0.5 & 0.03 & 0 & 1 & 1.5 & true & -- & --\\
NC & Walmart & 1 & 256 & 128 & 300 & 0.5 & 0.01 & 0.0001 & -1.5 & -0.5 & true & -- & --\\
NC & Yelp & 3 & 256 & 128 & 100 & 0.2 & 0.01 & 0.0001 & -1.5 & -0.5 & true & -- & --\\
\bottomrule\end{longtable}
\noindent 
Additional columns: $q_{1}$=\texttt{HNHN\_\allowbreak{}alpha}; $q_{2}$=\texttt{HNHN\_\allowbreak{}beta}; $q_{3}$=\texttt{HNHN\_\allowbreak{}nonlinear\_\allowbreak{}inbetween}; $q_{4}$=\texttt{pooling}; $q_{5}$=\texttt{early\_\allowbreak{}stop}.\par\smallskip
\endgroup\medskip
\begingroup
\setlength{\tabcolsep}{3pt}
\setlength{\LTcapwidth}{\textwidth}
\renewcommand{\arraystretch}{1.08}
\begin{longtable}{@{}llrrrrrrrrrrrrrr@{}}
\caption{DHG-Bench YAML settings for UniGCNII.}\label{tab:yaml-unigcnii}\\
\toprule Task & Benchmark & $L$ & $E$ & $H$ & $T$ & $p$ & $\eta$ & $\lambda$ & $q_{1}$ & $q_{2}$ & $q_{3}$ & $q_{4}$ & $q_{5}$ & $q_{6}$ & $q_{7}$\\\midrule
\endfirsthead
\toprule Task & Benchmark & $L$ & $E$ & $H$ & $T$ & $p$ & $\eta$ & $\lambda$ & $q_{1}$ & $q_{2}$ & $q_{3}$ & $q_{4}$ & $q_{5}$ & $q_{6}$ & $q_{7}$\\\midrule
\endhead
EP & Actor & 1 & -- & 256 & 200 & 0.5 & 0.01 & 0 & 0.5 & prelu & 1 & 0.5 & mean & -- & --\\
EP & Cora-CA & 2 & -- & 256 & 50 & 0.8 & 0.001 & 0 & 0.5 & prelu & 0.3 & 0.8 & mean & -- & --\\
EP & DBLP-CA & 2 & -- & 128 & 100 & 0.5 & 0.001 & 0 & 0.3 & prelu & 0.2 & 0.3 & mean & -- & --\\
EP & Cora & 2 & -- & 256 & 50 & 0.8 & 0.001 & 0 & 0.5 & prelu & 0.3 & 0.8 & mean & -- & --\\
EP & Pubmed & 2 & -- & 128 & 100 & 0 & 0.001 & 0 & 0.3 & prelu & 0.5 & 0.5 & mean & -- & --\\
EP & Pokec & 1 & -- & 128 & 100 & 0 & 0.001 & 0 & 0.5 & prelu & 0.5 & 0.5 & mean & -- & --\\
HC & IMDB-Dir-Form & 1 & 64 & 64 & 200 & 0.2 & 0.01 & 0 & 0.5 & prelu & 0.8 & 0.8 & mean & max & true\\
HC & IMDB-Dir-Genre & 2 & 64 & 64 & 200 & 0.2 & 0.01 & 0 & 0.5 & prelu & 0.1 & 0.8 & mean & mean & true\\
HC & RHG-10 & 2 & 64 & 64 & 200 & 0.2 & 0.01 & 0 & 0.5 & prelu & 0.8 & 0.8 & mean & mean & true\\
HC & RHG-3 & 2 & 64 & 64 & 200 & 0.2 & 0.01 & 0 & 0.5 & prelu & 0.8 & 0.8 & mean & mean & true\\
HC & Steam-Player & 2 & 64 & 64 & 200 & 0.2 & 0.01 & 0 & 0.5 & prelu & 0.1 & 0.8 & mean & mean & true\\
HC & Twitter-Friend & 1 & 64 & 64 & 200 & 0.2 & 0.01 & 0 & 0.5 & prelu & 0.8 & 0.8 & mean & max & true\\
NC & ModelNet40 & 1 & -- & 512 & 200 & 0.5 & 0.01 & 0 & 0.5 & prelu & 0.3 & 0.1 & mean & -- & --\\
NC & NTU2012 & 1 & -- & 512 & 200 & 0.5 & 0.01 & 0 & 0.5 & prelu & 0.3 & 0.1 & mean & -- & --\\
NC & Actor & 1 & -- & 256 & 200 & 0.5 & 0.01 & 0 & 0.5 & prelu & 1 & 0.5 & mean & -- & --\\
NC & Cora-CA & 2 & -- & 256 & 50 & 0.8 & 0.001 & 0 & 0.5 & prelu & 0.3 & 0.8 & mean & -- & --\\
NC & DBLP-CA & 2 & -- & 128 & 100 & 0.5 & 0.001 & 0 & 0.3 & prelu & 0.2 & 0.3 & mean & -- & --\\
NC & Cora & 2 & -- & 256 & 50 & 0.8 & 0.001 & 0 & 0.5 & prelu & 0.3 & 0.8 & mean & -- & --\\
NC & Pubmed & 2 & -- & 128 & 100 & 0 & 0.001 & 0 & 0.3 & prelu & 0.5 & 0.5 & mean & -- & --\\
NC & Gamers & 1 & -- & 128 & 100 & 0.1 & 0.001 & 0 & 0.5 & prelu & 0 & 0.5 & mean & -- & --\\
NC & Pokec & 1 & -- & 128 & 100 & 0 & 0.001 & 0 & 0.5 & prelu & 0.5 & 0.5 & mean & -- & --\\
NC & Ratings & 1 & -- & 128 & 50 & 0.1 & 0.001 & 0 & 0.5 & prelu & 0 & 0.8 & mean & -- & --\\
NC & Trivago & 2 & -- & 64 & 1000 & 0.5 & 0.01 & 0 & 0.2 & relu & 0 & 0.3 & sum & -- & --\\
NC & Walmart & 2 & -- & 128 & 200 & 0.5 & 0.001 & 0 & 0.3 & prelu & 0 & 0.3 & mean & -- & --\\
NC & Yelp & 2 & -- & 128 & 180 & 0.2 & 0.01 & 0 & 0 & prelu & 0.8 & 0.8 & mean & -- & --\\
\bottomrule\end{longtable}
\noindent 
Additional columns: $q_{1}$=\texttt{input\_\allowbreak{}drop}; $q_{2}$=\texttt{activation}; $q_{3}$=\texttt{restart\_\allowbreak{}alpha}; $q_{4}$=\texttt{lamda}; $q_{5}$=\texttt{reduce}; $q_{6}$=\texttt{pooling}; $q_{7}$=\texttt{early\_\allowbreak{}stop}.\par\smallskip
\noindent Shared settings: \texttt{use\_\allowbreak{}norm}=false.\par
\endgroup\medskip
\begingroup
\setlength{\tabcolsep}{3pt}
\setlength{\LTcapwidth}{\textwidth}
\renewcommand{\arraystretch}{1.08}
\begin{longtable}{@{}llrrrrrrrrrrrr@{}}
\caption{DHG-Bench YAML settings for AllSetTransformer.}\label{tab:yaml-allsetformer}\\
\toprule Task & Benchmark & $L$ & $E$ & $H$ & $M$ & $T$ & $p$ & $\eta$ & $q_{1}$ & $q_{2}$ & $q_{3}$ & $q_{4}$ & $q_{5}$\\\midrule
\endfirsthead
\toprule Task & Benchmark & $L$ & $E$ & $H$ & $M$ & $T$ & $p$ & $\eta$ & $q_{1}$ & $q_{2}$ & $q_{3}$ & $q_{4}$ & $q_{5}$\\\midrule
\endhead
EP & Actor & 2 & 128 & 128 & 2 & 200 & 0.5 & 0.001 & true & 128 & 2 & -- & --\\
EP & Cora-CA & 1 & 128 & 256 & 2 & 300 & 0.5 & 0.001 & true & 128 & 2 & -- & --\\
EP & DBLP-CA & 2 & 128 & 128 & 2 & 50 & 0.5 & 0.001 & true & 128 & 2 & -- & --\\
EP & Cora & 2 & 128 & 256 & 2 & 100 & 0.5 & 0.001 & true & 64 & 2 & -- & --\\
EP & Pubmed & 1 & 128 & 256 & 2 & 300 & 0.5 & 0.001 & true & 64 & 2 & -- & --\\
EP & Pokec & 2 & 128 & 64 & 2 & 200 & 0.5 & 0.001 & true & 128 & 2 & -- & --\\
HC & IMDB-Dir-Form & 1 & 64 & 64 & 2 & 200 & 0.2 & 0.01 & false & 64 & 2 & mean & true\\
HC & IMDB-Dir-Genre & 1 & 64 & 64 & 2 & 200 & 0.2 & 0.01 & false & 64 & 2 & mean & true\\
HC & RHG-10 & 1 & 64 & 64 & 2 & 200 & 0.2 & 0.01 & false & 64 & 2 & mean & true\\
HC & RHG-3 & 1 & 64 & 64 & 2 & 200 & 0.2 & 0.01 & false & 64 & 2 & mean & true\\
HC & Steam-Player & 1 & 64 & 64 & 2 & 200 & 0.2 & 0.01 & false & 64 & 2 & mean & true\\
HC & Twitter-Friend & 1 & 64 & 64 & 2 & 200 & 0.2 & 0.01 & false & 64 & 2 & mean & true\\
NC & ModelNet40 & 1 & 128 & 512 & 2 & 100 & 0.5 & 0.001 & true & 64 & 2 & -- & --\\
NC & NTU2012 & 1 & 128 & 512 & 2 & 100 & 0.5 & 0.001 & true & 64 & 2 & -- & --\\
NC & Actor & 2 & 128 & 128 & 2 & 200 & 0.5 & 0.001 & true & 128 & 2 & -- & --\\
NC & Cora-CA & 1 & 128 & 256 & 2 & 300 & 0.5 & 0.001 & true & 128 & 2 & -- & --\\
NC & DBLP-CA & 2 & 128 & 128 & 2 & 50 & 0.5 & 0.001 & true & 128 & 2 & -- & --\\
NC & Cora & 2 & 128 & 256 & 2 & 100 & 0.5 & 0.001 & true & 64 & 2 & -- & --\\
NC & Pubmed & 1 & 128 & 256 & 2 & 300 & 0.5 & 0.001 & true & 64 & 2 & -- & --\\
NC & Gamers & 2 & 128 & 128 & 2 & 200 & 0.5 & 0.001 & true & 128 & 2 & -- & --\\
NC & Pokec & 2 & 128 & 64 & 2 & 200 & 0.5 & 0.001 & true & 128 & 2 & -- & --\\
NC & Ratings & 2 & 128 & 128 & 2 & 200 & 0.5 & 0.01 & true & 128 & 2 & -- & --\\
NC & Trivago & 4 & 128 & 64 & 1 & 1000 & 0.5 & 0.01 & true & 64 & 1 & -- & --\\
NC & Walmart & 2 & 128 & 128 & 2 & 200 & 0.5 & 0.001 & true & 128 & 2 & -- & --\\
NC & Yelp & 1 & 128 & 64 & 1 & 500 & 0.2 & 0.001 & true & 128 & 2 & -- & --\\
\bottomrule\end{longtable}
\noindent 
Additional columns: $q_{1}$=\texttt{LearnMask}; $q_{2}$=\texttt{decoder\_\allowbreak{}hidden}; $q_{3}$=\texttt{decoder\_\allowbreak{}num\_\allowbreak{}layers}; $q_{4}$=\texttt{pooling}; $q_{5}$=\texttt{early\_\allowbreak{}stop}.\par\smallskip
\noindent Shared settings: \texttt{wd}=0; \texttt{aggregate}=mean; \texttt{normalization}=ln; \texttt{deepset\_\allowbreak{}input\_\allowbreak{}norm}=true; \texttt{GPR}=true; \texttt{PMA}=true; \texttt{heads}=1; \texttt{output\_\allowbreak{}heads}=1.\par
\endgroup\medskip
\begingroup
\setlength{\tabcolsep}{3pt}
\setlength{\LTcapwidth}{\textwidth}
\renewcommand{\arraystretch}{1.08}
\begin{longtable}{@{}llrrrrrrrrrrrrrr@{}}
\caption{DHG-Bench YAML settings for ED-HNN.}\label{tab:yaml-edhnn}\\
\toprule Task & Benchmark & $L$ & $E$ & $H$ & $M$ & $T$ & $p$ & $\eta$ & $q_{1}$ & $q_{2}$ & $q_{3}$ & $q_{4}$ & $q_{5}$ & $q_{6}$ & $q_{7}$\\\midrule
\endfirsthead
\toprule Task & Benchmark & $L$ & $E$ & $H$ & $M$ & $T$ & $p$ & $\eta$ & $q_{1}$ & $q_{2}$ & $q_{3}$ & $q_{4}$ & $q_{5}$ & $q_{6}$ & $q_{7}$\\\midrule
\endhead
EP & Actor & 1 & 128 & 512 & 0 & 180 & 0.5 & 0.001 & 1 & 1 & 256 & 2 & 0.2 & -- & --\\
EP & Cora-CA & 1 & 128 & 512 & 0 & 500 & 0.5 & 0.001 & 1 & 1 & 256 & 1 & 0.1 & -- & --\\
EP & DBLP-CA & 1 & 128 & 512 & 1 & 40 & 0.5 & 0.001 & 1 & 1 & 256 & 2 & 0.7 & -- & --\\
EP & Cora & 1 & 128 & 512 & 0 & 500 & 0.5 & 0.001 & 1 & 1 & 256 & 1 & 0.5 & -- & --\\
EP & Pubmed & 1 & 128 & 512 & 0 & 100 & 0.5 & 0.001 & 1 & 1 & 256 & 1 & 0.8 & -- & --\\
EP & Pokec & 1 & 128 & 512 & 0 & 140 & 0.5 & 0.001 & 1 & 1 & 256 & 1 & 0.2 & -- & --\\
HC & IMDB-Dir-Form & 1 & 128 & 64 & 1 & 200 & 0.2 & 0.001 & -1 & -1 & 64 & 1 & 0.1 & max & true\\
HC & IMDB-Dir-Genre & 1 & 64 & 64 & 1 & 200 & 0.2 & 0.01 & -1 & -1 & 64 & 1 & 0.1 & mean & true\\
HC & RHG-10 & 1 & 64 & 64 & 1 & 200 & 0.2 & 0.001 & -1 & -1 & 128 & 1 & 0.1 & mean & true\\
HC & RHG-3 & 1 & 64 & 64 & 1 & 200 & 0.2 & 0.001 & -1 & -1 & 64 & 1 & 0.1 & mean & true\\
HC & Steam-Player & 1 & 128 & 64 & 2 & 200 & 0.2 & 0.001 & -1 & -1 & 64 & 1 & 0.1 & max & true\\
HC & Twitter-Friend & 1 & 64 & 64 & 1 & 200 & 0.2 & 0.001 & -1 & -1 & 64 & 1 & 0.1 & max & true\\
NC & ModelNet40 & 1 & 128 & 512 & 0 & 100 & 0.5 & 0.001 & 1 & 1 & 256 & 1 & 0.2 & -- & --\\
NC & NTU2012 & 1 & 128 & 512 & 0 & 100 & 0.5 & 0.001 & 1 & 1 & 256 & 1 & 0.2 & -- & --\\
NC & Actor & 1 & 128 & 512 & 0 & 180 & 0.5 & 0.001 & 1 & 1 & 256 & 2 & 0.2 & -- & --\\
NC & Cora-CA & 1 & 128 & 512 & 0 & 500 & 0.5 & 0.001 & 1 & 1 & 256 & 1 & 0.1 & -- & --\\
NC & DBLP-CA & 1 & 128 & 512 & 1 & 40 & 0.5 & 0.001 & 1 & 1 & 256 & 2 & 0.7 & -- & --\\
NC & Cora & 1 & 128 & 512 & 0 & 200 & 0.5 & 0.001 & 1 & 1 & 256 & 1 & 0.2 & -- & --\\
NC & Pubmed & 1 & 128 & 512 & 0 & 100 & 0.5 & 0.001 & 1 & 1 & 256 & 1 & 0.8 & -- & --\\
NC & Gamers & 1 & 128 & 512 & 1 & 160 & 0.5 & 0.001 & 1 & 1 & 256 & 2 & 0.2 & -- & --\\
NC & Pokec & 1 & 128 & 512 & 0 & 140 & 0.5 & 0.001 & 1 & 1 & 256 & 1 & 0.2 & -- & --\\
NC & Ratings & 1 & 128 & 512 & 0 & 160 & 0.5 & 0.001 & 1 & 1 & 256 & 2 & 0.1 & -- & --\\
NC & Trivago & 8 & 128 & 64 & 0 & 1000 & 0.5 & 0.01 & 0 & 1 & 64 & 1 & 0 & -- & --\\
NC & Walmart & 1 & 128 & 512 & 1 & 200 & 0.5 & 0.001 & 1 & 1 & 256 & 2 & 0.7 & -- & --\\
NC & Yelp & 1 & 128 & 64 & 1 & 500 & 0.2 & 0.001 & 1 & 1 & 256 & 2 & 0.2 & -- & --\\
\bottomrule\end{longtable}
\noindent 
Additional columns: $q_{1}$=\texttt{MLP2\_\allowbreak{}num\_\allowbreak{}layers}; $q_{2}$=\texttt{MLP3\_\allowbreak{}num\_\allowbreak{}layers}; $q_{3}$=\texttt{decoder\_\allowbreak{}hidden}; $q_{4}$=\texttt{decoder\_\allowbreak{}num\_\allowbreak{}layer}; $q_{5}$=\texttt{alpha}; $q_{6}$=\texttt{pooling}; $q_{7}$=\texttt{early\_\allowbreak{}stop}.\par\smallskip
\noindent Shared settings: \texttt{wd}=0; \texttt{edconv\_\allowbreak{}type}=EquivSet; \texttt{aggregate}=mean; \texttt{normalization}=ln; \texttt{activation}=prelu; \texttt{AllSet\_\allowbreak{}input\_\allowbreak{}norm}=true.\par
\endgroup\medskip
\begingroup
\setlength{\tabcolsep}{3pt}
\setlength{\LTcapwidth}{\textwidth}
\renewcommand{\arraystretch}{1.08}
\begin{longtable}{@{}llrrrrrrrrrrr@{}}
\caption{DHG-Bench YAML settings for HyperGT.}\label{tab:yaml-hypergt}\\
\toprule Task & Benchmark & $L$ & $E$ & $T$ & $p$ & $\eta$ & $\lambda$ & $q_{1}$ & $q_{2}$ & $q_{3}$ & $q_{4}$ & $q_{5}$\\\midrule
\endfirsthead
\toprule Task & Benchmark & $L$ & $E$ & $T$ & $p$ & $\eta$ & $\lambda$ & $q_{1}$ & $q_{2}$ & $q_{3}$ & $q_{4}$ & $q_{5}$\\\midrule
\endhead
EP & Actor & 2 & 128 & 100 & 0.5 & 0.005 & 0.001 & 128 & false & max & -- & --\\
EP & Cora-CA & 2 & 128 & 100 & 0.5 & 0.005 & 0.001 & 256 & false & maxmin & -- & --\\
EP & DBLP-CA & 2 & 128 & 100 & 0.5 & 0.0005 & 0.01 & 128 & false & maxmin & -- & --\\
EP & Cora & 2 & 128 & 100 & 0.5 & 0.005 & 0.001 & 128 & true & maxmin & -- & --\\
EP & Pubmed & 2 & 128 & 100 & 0.5 & 0.01 & 0.001 & 128 & true & maxmin & -- & --\\
EP & Pokec & 2 & 128 & 100 & 0.5 & 0.01 & 0.01 & 128 & false & maxmin & -- & --\\
HC & IMDB-Dir-Form & 1 & 64 & 100 & 0 & 0.005 & 0 & 64 & true & -- & mean & true\\
HC & IMDB-Dir-Genre & 1 & 64 & 100 & 0 & 0.005 & 0 & 64 & true & -- & mean & true\\
HC & RHG-10 & 1 & 64 & 200 & 0 & 0.005 & 0 & 64 & true & -- & max & true\\
HC & RHG-3 & 1 & 64 & 200 & 0 & 0.01 & 0 & 64 & true & -- & mean & true\\
HC & Steam-Player & 1 & 64 & 100 & 0 & 0.005 & 0 & 64 & true & -- & max & true\\
HC & Twitter-Friend & 1 & 64 & 100 & 0 & 0.002 & 0 & 64 & true & -- & max & true\\
NC & ModelNet40 & 1 & 128 & 200 & 0.5 & 0.001 & 0.01 & 512 & false & -- & -- & --\\
NC & NTU2012 & 1 & 128 & 200 & 0.5 & 0.001 & 0.01 & 512 & false & -- & -- & --\\
NC & Actor & 2 & 128 & 100 & 0.5 & 0.005 & 0.001 & 128 & false & -- & -- & --\\
NC & Cora-CA & 2 & 128 & 100 & 0.5 & 0.005 & 0.001 & 256 & false & -- & -- & --\\
NC & DBLP-CA & 2 & 128 & 100 & 0.5 & 0.0005 & 0.01 & 128 & false & -- & -- & --\\
NC & Cora & 2 & 128 & 100 & 0.5 & 0.005 & 0.001 & 128 & true & -- & -- & --\\
NC & Pubmed & 2 & 128 & 100 & 0.5 & 0.01 & 0.001 & 128 & true & -- & -- & --\\
NC & Gamers & 2 & 128 & 100 & 0.5 & 0.01 & 0.01 & 128 & false & -- & -- & --\\
NC & Pokec & 2 & 128 & 100 & 0.5 & 0.01 & 0.01 & 128 & false & -- & -- & --\\
NC & Ratings & 2 & 128 & 100 & 0.5 & 0.005 & 0.001 & 256 & false & -- & -- & --\\
NC & Trivago & 2 & 128 & 100 & 0.5 & 0.0005 & 0.01 & 128 & false & -- & -- & --\\
NC & Walmart & 2 & 128 & 100 & 0.5 & 0.0005 & 0.01 & 128 & false & -- & -- & --\\
NC & Yelp & 2 & 128 & 100 & 0.5 & 0.001 & 0 & 32 & true & -- & -- & --\\
\bottomrule\end{longtable}
\noindent 
Additional columns: $q_{1}$=\texttt{hidden\_\allowbreak{}channels}; $q_{2}$=\texttt{use\_\allowbreak{}edge\_\allowbreak{}loss}; $q_{3}$=\texttt{aggr\_\allowbreak{}mode}; $q_{4}$=\texttt{pooling}; $q_{5}$=\texttt{early\_\allowbreak{}stop}.\par\smallskip
\noindent Shared settings: \texttt{num\_\allowbreak{}heads}=4; \texttt{nb\_\allowbreak{}random\_\allowbreak{}features}=30; \texttt{use\_\allowbreak{}bn}=true; \texttt{use\_\allowbreak{}gumbel}=true; \texttt{use\_\allowbreak{}residual}=true; \texttt{use\_\allowbreak{}act}=true; \texttt{use\_\allowbreak{}jk}=true; \texttt{nb\_\allowbreak{}gumbel\_\allowbreak{}sample}=10; \texttt{rb\_\allowbreak{}order}=0; \texttt{rb\_\allowbreak{}trans}=sigmoid; \texttt{pe}=HEPEHtEPE; \texttt{hefeat}=mean.\par
\endgroup\medskip
\begingroup
\setlength{\tabcolsep}{3pt}
\setlength{\LTcapwidth}{\textwidth}
\renewcommand{\arraystretch}{1.08}
\begin{longtable}{@{}llrrrrr@{}}
\caption{DHG-Bench YAML settings for EHNN.}\label{tab:yaml-ehnn}\\
\toprule Task & Benchmark & $H$ & $T$ & $p$ & $\lambda$ & $q_{1}$\\\midrule
\endfirsthead
\toprule Task & Benchmark & $H$ & $T$ & $p$ & $\lambda$ & $q_{1}$\\\midrule
\endhead
EP & Actor & 128 & 500 & 0.2 & 1e-05 & 4\\
EP & Cora-CA & 128 & 100 & 0 & 0.001 & 8\\
EP & DBLP-CA & 128 & 200 & 0 & 1e-05 & 4\\
EP & Cora & 128 & 80 & 0.2 & 1e-05 & 8\\
EP & Pubmed & 128 & 200 & 0 & 1e-05 & 8\\
EP & Pokec & 128 & 180 & 0 & 0 & 4\\
NC & ModelNet40 & 256 & 500 & 0.2 & 1e-05 & 8\\
NC & NTU2012 & 256 & 500 & 0.2 & 1e-05 & 8\\
NC & Actor & 128 & 500 & 0.2 & 1e-05 & 4\\
NC & Cora-CA & 128 & 100 & 0 & 0.001 & 8\\
NC & DBLP-CA & 128 & 200 & 0 & 1e-05 & 4\\
NC & Cora & 128 & 80 & 0.2 & 1e-05 & 8\\
NC & Pubmed & 128 & 200 & 0 & 1e-05 & 8\\
NC & Gamers & 128 & 200 & 0.1 & 1e-05 & 4\\
NC & Pokec & 128 & 180 & 0 & 0 & 4\\
NC & Ratings & 128 & 140 & 0 & 1e-05 & 4\\
NC & Trivago & 256 & 500 & 0.2 & 1e-05 & 4\\
NC & Walmart & 256 & 500 & 0.2 & 1e-05 & 4\\
NC & Yelp & 256 & 1000 & 0.2 & 1e-05 & 4\\
\bottomrule\end{longtable}
\noindent 
Additional columns: $q_{1}$=\texttt{ehnn\_\allowbreak{}n\_\allowbreak{}heads}.\par\smallskip
\noindent Shared settings: \texttt{All\_\allowbreak{}num\_\allowbreak{}layers}=1; \texttt{embedding\_\allowbreak{}hidden}=128; \texttt{normalization}=ln; \texttt{clip\_\allowbreak{}grad}=false; \texttt{clip\_\allowbreak{}thresh}=5; \texttt{lr}=0.001; \texttt{chunk\_\allowbreak{}size}=1000; \texttt{ehnn\_\allowbreak{}ff\_\allowbreak{}layers}=2; \texttt{ehnn\_\allowbreak{}qk\_\allowbreak{}channel}=64; \texttt{ehnn\_\allowbreak{}inner\_\allowbreak{}channel}=64; \texttt{ehnn\_\allowbreak{}hidden\_\allowbreak{}channel}=64; \texttt{ehnn\_\allowbreak{}type}=linear; \texttt{ehnn\_\allowbreak{}pe\_\allowbreak{}dim}=64; \texttt{ehnn\_\allowbreak{}hyper\_\allowbreak{}dim}=64; \texttt{ehnn\_\allowbreak{}hyper\_\allowbreak{}layers}=2; \texttt{ehnn\_\allowbreak{}hyper\_\allowbreak{}dropout}=0.2; \texttt{ehnn\_\allowbreak{}force\_\allowbreak{}broadcast}=False; \texttt{ehnn\_\allowbreak{}input\_\allowbreak{}dropout}=0; \texttt{ehnn\_\allowbreak{}mlp\_\allowbreak{}classifier}=True; \texttt{decoder\_\allowbreak{}hidden}=64; \texttt{decoder\_\allowbreak{}num\_\allowbreak{}layers}=2; \texttt{ehnn\_\allowbreak{}att0\_\allowbreak{}dropout}=0; \texttt{ehnn\_\allowbreak{}att1\_\allowbreak{}dropout}=0.\par
\endgroup\medskip
\begingroup
\setlength{\tabcolsep}{3pt}
\setlength{\LTcapwidth}{\textwidth}
\renewcommand{\arraystretch}{1.08}
\begin{longtable}{@{}llrrrrrrr@{}}
\caption{DHG-Bench YAML settings for T-HyperGNN.}\label{tab:yaml-tmphn}\\
\toprule Task & Benchmark & $H$ & $T$ & $\eta$ & $q_{1}$ & $q_{2}$ & $q_{3}$ & $q_{4}$\\\midrule
\endfirsthead
\toprule Task & Benchmark & $H$ & $T$ & $\eta$ & $q_{1}$ & $q_{2}$ & $q_{3}$ & $q_{4}$\\\midrule
\endhead
EP & Actor & 256 & 50 & 0.01 & max & 10 & 256 & [5]\\
EP & Cora-CA & 256 & 60 & 0.01 & max & 10 & 256 & [5]\\
EP & DBLP-CA & 256 & 40 & 0.01 & max & 10 & 128 & [5]\\
EP & Cora & 128 & 80 & 0.01 & max & 10 & 128 & [5]\\
EP & Pubmed & 256 & 60 & 0.01 & max & 10 & 256 & [5]\\
EP & Pokec & 256 & 80 & 0.01 & max & 1 & 256 & [5]\\
NC & ModelNet40 & 128 & 50 & 0.01 & -- & -- & 128 & [5]\\
NC & NTU2012 & 128 & 50 & 0.01 & -- & -- & 128 & [5]\\
NC & Actor & 256 & 100 & 0.01 & -- & -- & 256 & [5]\\
NC & Cora-CA & 256 & 100 & 0.01 & -- & -- & 256 & [5]\\
NC & DBLP-CA & 256 & 20 & 0.01 & -- & -- & 128 & [5]\\
NC & Cora & 128 & 20 & 0.01 & -- & -- & 128 & [5]\\
NC & Pubmed & 256 & 100 & 0.01 & -- & -- & 256 & [5]\\
NC & Gamers & 256 & 50 & 0.001 & -- & -- & 256 & [5]\\
NC & Pokec & 256 & 50 & 0.01 & -- & -- & 256 & [5]\\
NC & Ratings & 256 & 50 & 0.01 & -- & -- & 256 & [5]\\
NC & Trivago & 128 & 100 & 0.001 & -- & -- & 128 & [3]\\
NC & Walmart & 128 & 100 & 0.001 & -- & -- & 128 & [3]\\
NC & Yelp & 128 & 100 & 0.001 & -- & -- & 128 & [3]\\
\bottomrule\end{longtable}
\noindent 
Additional columns: $q_{1}$=\texttt{aggr\_\allowbreak{}mode}; $q_{2}$=\texttt{display\_\allowbreak{}step}; $q_{3}$=\texttt{hid\_\allowbreak{}dim}; $q_{4}$=\texttt{Mlst}.\par\smallskip
\noindent Shared settings: \texttt{All\_\allowbreak{}num\_\allowbreak{}layers}=1; \texttt{embedding\_\allowbreak{}hidden}=128; \texttt{dropout}=0.5; \texttt{wd}=0; \texttt{hyperG\_\allowbreak{}norm}=false; \texttt{M}=5; \texttt{combine}=concat.\par
\endgroup\medskip

\section{Details of Synthetic Construction and Evaluation (Supplementing \cref{sec:synthetic})}
\label{app:synthetic}

\subsection{Construction}

Each hypergraph contains $m=21$ disjoint six-node motif instances, each of
which follows either template $\mathcal T_0$ or template $\mathcal T_1$.
Let $z_j\in\{0,1\}$ indicate the template of motif instance $j$.  The
hypergraph label is determined by the majority template:
\begin{equation}
  Y=\mathbf{1}\!\left\{\sum_{j=1}^{m}z_j>m/2\right\}.
  \label{eq:synthetic-majority-label}
\end{equation}

\paragraph{Sampling motif counts.}
Each train/validation/test split contains equal numbers of the two
classes. For an assigned label $Y=y$, we independently sample
$z_j\sim\operatorname{Bernoulli}(p_y)$ for $j=1,\ldots,21$,
where $p_0=0.1$ and $p_1=0.9$.
We reject and resample the template assignments if their majority
does not agree with $y$.
Thus, every hypergraph contains exactly 21 motif instances, while
the counts of the two templates vary.

\paragraph{Motif templates and features.}
Node features are one-hot indicators of the six within-motif roles and encode
neither the motif template nor the hypergraph label.  Writing $(abc)$ for the
hyperedge $\{a,b,c\}$, the two motif templates are
\begin{equation}
  \mathcal T_0=\{(012),(034),(513),(524)\},\qquad
  \mathcal T_1=\{(013),(024),(512),(534)\}.
  \label{eq:synthetic-motif-templates}
\end{equation}
Every unordered pair of roles occurs with the same multiplicity in the two
templates.  Their weighted clique expansions are therefore identical, so
$\mathcal T_0$ and $\mathcal T_1$ are lower-order equivalent.
Each motif instance additionally contains one unit-weight two-node
hyperedge for every pair in the common clique-expansion support.
These identical pairwise edges provide no template or label signal
and preserve the lower-order equivalence of the two templates.

In the case with \emph{essential} higher-order information, no additional
signal is included.  In the case with \emph{redundant} higher-order
information, we add a two-node hyperedge specific to each template, called a
\emph{pairwise tag}, to each motif instance.
Each instance of $\mathcal T_0$ receives the pairwise tag
$\{1,4\}$ with weight four, and each instance of $\mathcal T_1$ receives
$\{2,3\}$ with weight four.  These pairs do not occur in the original motif
templates.  The tags therefore make the two templates distinguishable from
their weighted clique expansions and reveal the majority template using
lower-order information alone.

\subsection{Evaluation}

We evaluate the 15 higher-order and three lower-order models from
\cref{sec:empirical_analysis:setup} that support hypergraph classification.
For each case, we independently select the best model in each family by mean
original-view validation accuracy, using validation Macro-F1 to break ties,
and fix its identity and hyperparameters throughout the intervention.

Final test performance is averaged over three independently generated
datasets and three training seeds, giving nine runs per model and view.  Each
dataset contains balanced 300/100/100 train/validation/test hypergraphs.  We
evaluate $r\in\{0,.25,.50,.75,1\}$, with partial perturbations constructed as
described in \cref{app:partial-perturbation-construction}.

\section{Details of Benchmark Results (Supplementing \cref{sec:real_world})} \label{app:real-world-results}

\subsection{Detailed benchmark-wise results}
\label{app:gap-retention}

\cref{tab:registered-pair-trajectories} gives original, intermediate, and fully perturbed gaps for all 25 benchmarks.

\begin{table}[t!]
  \centering
  \small
  \caption{\textbf{Complete selected-pair gaps.} $\Delta$ is the higher-order
  model's test performance minus that of the fixed lower-order comparator, in
  percentage points.  The metric is accuracy for node classification and hypergraph classification and AP
  for hyperedge prediction.}
  \setlength{\tabcolsep}{3.2pt}
  \begin{tabular}{lllrrr}
\toprule
Task & Dataset & Fixed pair & $\Delta^{(0)}$ & $\Delta^{(0.5)}$ & $\Delta^{(1)}$ \\
\midrule
Node classification & ModelNet40 & EDHNN--MLP & +2.00 & +1.85 & +1.87 \\
Node classification & NTU2012 & LEGCN--MLP & +2.45 & +1.68 & +1.39 \\
Node classification & Actor & EHNN--MLP & +0.42 & +0.07 & -0.27 \\
Node classification & Cora-CA & TFHNN--CEGCN & +6.50 & +6.15 & +6.40 \\
Node classification & DBLP-CA & PhenomNN--CEGCN & +2.12 & +2.13 & +2.05 \\
Node classification & Cora & TFHNN--CEGCN & +4.38 & +4.38 & +4.52 \\
Node classification & Pubmed & LEGCN--MLP & +1.49 & +1.40 & +1.47 \\
Node classification & Gamers & HCHA--MLP & -0.01 & -0.01 & -0.01 \\
Node classification & Pokec & TFHNN--MLP & +0.09 & +0.24 & -0.55 \\
Node classification & Ratings & LEGCN--MLP & +1.76 & +1.69 & +1.75 \\
Node classification & Trivago & TFHNN--CEGCN & +44.39 & +43.11 & +42.79 \\
Node classification & Walmart & EDHNN--CEGCN & +10.61 & +7.39 & +8.70 \\
Node classification & Yelp$^{\ast}$ & TFHNN--MLP & +3.28 & +0.80 & +1.07 \\
\addlinespace[2pt]
Hyperedge prediction & Actor & TFHNN--MLP & +10.15 & +10.15 & +10.15 \\
Hyperedge prediction & Cora-CA & UniGCNII--CEGAT & +9.39 & +8.97 & +7.85 \\
Hyperedge prediction & DBLP-CA & PhenomNN--CEGAT & +2.76 & +2.87 & +2.61 \\
Hyperedge prediction & Cora & PhenomNN--CEGAT & -1.61 & -1.97 & -1.57 \\
Hyperedge prediction & Pubmed & EHNN--CEGCN & -3.06 & -5.23 & -5.70 \\
Hyperedge prediction & Pokec & HGNN--CEGCN & +23.54 & +23.86 & +23.55 \\
\addlinespace[2pt]
Hypergraph classification & IMDB-Dir-Form & HCHA--CEGCN & +0.00 & -1.19 & -1.90 \\
Hypergraph classification & IMDB-Dir-Genre & HNHN--CEGCN & -2.06 & -1.70 & -0.66 \\
Hypergraph classification & RHG-10 & HNHN--CEGAT & +1.00 & +2.61 & +1.72 \\
Hypergraph classification & RHG-3 & HNHN--CEGCN & +1.33 & +1.33 & +1.33 \\
Hypergraph classification & Steam-Player & HNHN--CEGAT & +3.72 & +0.48 & -0.22 \\
Hypergraph classification & Twitter-Friend & HJRL--MLP & +1.02 & -1.27 & -2.71 \\
\bottomrule
\end{tabular}

  \label{tab:registered-pair-trajectories}
\end{table}

The 80\% boundary in \cref{fig:real-gap-map} is a descriptive
convention for a continuous quantity, not a threshold for statistical significance. For a
positive original advantage, define retention as
$R=\Delta^{(1)}/\Delta^{(0)}$. \cref{tab:gap-retention} reports both
gaps and the retention ratio for all 25 benchmarks. 
Thus the qualitative conclusion that a majority of material advantages
largely survive the intervention does not depend on choosing 80\% rather than
90\% or 95\%. Ratios above 100\% indicate a larger gap on the fully perturbed view, while negative ratios indicate that the advantage reverses. 
Ratios for nonpositive original gaps are N/A.
Of the 18 comparisons with an original gap of at least one point, 14 retain at least 80\%, 12 retain at least 90\%, and 10 retain at least 95\%. The unweighted median retention across these 18 benchmarks is 96.7\%.

\clearpage
\begin{table}[H]
\centering
\small
\caption{\textbf{Gap retention for every benchmark.} Gaps are accuracy points (node and hypergraph classification) or AP points (hyperedge prediction). Retention is $100\Delta^{(1)}/\Delta^{(0)}$, displayed to one decimal place. $\dagger$ marks a positive original gap below one point. N/A denotes a nonpositive original gap.}
\label{tab:gap-retention}
\begin{tabular}{@{}llrrr@{}}
\toprule
Task & Dataset & Original gap & Endpoint gap & Retained (\%) \\
\midrule
Node classification & Actor$^{\dagger}$ & 0.418 & -0.271 & -64.7 \\
Node classification & Cora & 4.376 & 4.523 & 103.4 \\
Node classification & Cora-CA & 6.499 & 6.401 & 98.5 \\
Node classification & DBLP-CA & 2.118 & 2.053 & 97.0 \\
Node classification & Pokec$^{\dagger}$ & 0.089 & -0.551 & -620.0 \\
Node classification & Pubmed & 1.487 & 1.467 & 98.6 \\
Node classification & Gamers & -0.008 & -0.008 & N/A \\
Node classification & ModelNet40 & 2.003 & 1.873 & 93.5 \\
Node classification & NTU2012 & 2.452 & 1.392 & 56.8 \\
Node classification & Ratings & 1.758 & 1.752 & 99.7 \\
Node classification & Trivago & 44.385 & 42.790 & 96.4 \\
Node classification & Walmart & 10.607 & 8.701 & 82.0 \\
Node classification & Yelp & 3.278 & 1.069 & 32.6 \\
\midrule
Hyperedge prediction & Actor & 10.146 & 10.146 & 100.0 \\
Hyperedge prediction & Cora & -1.614 & -1.566 & N/A \\
Hyperedge prediction & Cora-CA & 9.394 & 7.851 & 83.6 \\
Hyperedge prediction & DBLP-CA & 2.760 & 2.606 & 94.4 \\
Hyperedge prediction & Pokec & 23.540 & 23.549 & 100.0 \\
Hyperedge prediction & Pubmed & -3.060 & -5.703 & N/A \\
\midrule
Hypergraph classification & IMDB-Dir-Form & 0.000 & -1.900 & N/A \\
Hypergraph classification & IMDB-Dir-Genre & -2.060 & -0.657 & N/A \\
Hypergraph classification & RHG-10 & 1.000 & 1.723 & 172.3 \\
Hypergraph classification & RHG-3 & 1.330 & 1.330 & 100.0 \\
Hypergraph classification & Steam-Player & 3.720 & -0.217 & -5.8 \\
Hypergraph classification & Twitter-Friend & 1.020 & -2.711 & -265.8 \\
\bottomrule
\end{tabular}
\end{table}

\subsection{Secondary-metric results}
\label{app:secondary-gaps}

\begin{figure}[t!]
  \centering
  \includegraphics[width=\textwidth]{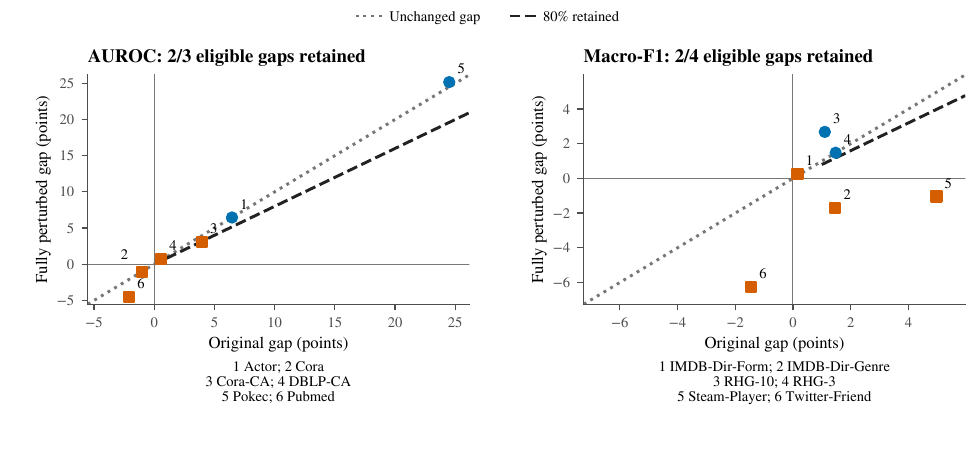}
  \caption{\textbf{Original and fully perturbed performance gaps using AUROC for
  hyperedge prediction and Macro-F1 for hypergraph classification.}
  Checkpoints and model pairs are selected using the corresponding secondary
  validation metric.
  Blue circles indicate original gaps of at least one percentage point that retain at least
  80\% after perturbation; orange squares
  denote the other comparisons. The dotted diagonal denotes unchanged gaps and the dashed line 80\% retention.}
  \label{fig:secondary-gaps}
\end{figure}

We report the results using AUROC for hyperedge prediction
and Macro-F1 for hypergraph classification.
\cref{fig:secondary-gaps} shows the original and fully perturbed gaps under
these metric-specific selections. Blue circles mark positive original gaps of
at least one point for which at least 80\% of the gap remains after
perturbation; orange squares show the remaining comparisons. The diagonal
indicates an unchanged gap, while points below it indicate gap reduction.
The corresponding model pairs, absolute gaps, and retention ratios are
reported in \cref{tab:secondary-gaps}.

\begin{table}[t!]
\centering
\small
\caption{Secondary-metric gaps after selecting checkpoints and model pairs
using the corresponding original-view validation metric: AUROC for
hyperedge prediction and Macro-F1 for hypergraph classification. Model
identities and hyperparameters remain fixed across views. Gaps are measured
in percentage points. Retention is reported only when
$\Delta^{(0)}\geq1$ point; ``--'' denotes an ineligible comparison.}
\label{tab:secondary-gaps}
\setlength{\tabcolsep}{3pt}
\begin{tabular}{llllrrr}
\toprule
Task & Dataset & Higher-order & Lower-order &
$\Delta^{(0)}$ & $\Delta^{(1)}$ & Retention \\
\midrule
Hyperedge prediction & Actor
& TF-HNN & MLP & 6.46 & 6.46 & 100.0\% \\
Hyperedge prediction & Cora
& PhenomNN & CEGAT & $-1.04$ & $-1.11$ & -- \\
Hyperedge prediction & Cora-CA
& UniGCNII & CEGAT & 3.94 & 3.04 & 77.2\% \\
Hyperedge prediction & DBLP-CA
& PhenomNN & CEGAT & 0.56 & 0.69 & -- \\
Hyperedge prediction & Pokec
& HGNN & CEGCN & 24.50 & 25.17 & 102.7\% \\
Hyperedge prediction & Pubmed
& EHNN & CEGCN & $-2.09$ & $-4.52$ & -- \\
\addlinespace[2pt]
Hypergraph classification & IMDB-Dir-Form
& TF-HNN & CEGCN & 0.16 & 0.26 & -- \\
Hypergraph classification & IMDB-Dir-Genre
& Sheaf & CEGCN & 1.47 & $-1.71$ & $-116.2\%$ \\
Hypergraph classification & RHG-10
& HNHN & CEGCN & 1.10 & 2.68 & 243.3\% \\
Hypergraph classification & RHG-3
& HNHN & CEGCN & 1.48 & 1.48 & 100.0\% \\
Hypergraph classification & Steam-Player
& HNHN & CEGAT & 4.97 & $-1.03$ & $-20.8\%$ \\
Hypergraph classification & Twitter-Friend
& HJRL & MLP & $-1.45$ & $-6.25$ & -- \\
\bottomrule
\end{tabular}
\end{table}

\subsection{Empirical check of lower-order baseline invariance}
\label{app:baseline-invariance}

We distinguish two empirical checks:
(1) evaluating the same fitted lower-order baseline on both views tests whether the representation and prediction pipelines preserve its outputs, and (2) training independently on each view with aligned seeds additionally tests the reproducibility of the training pipeline.
Identical lower-order inputs imply the first invariance in exact
arithmetic; they do not eliminate numerical or optimization variability
between independent fits.

The lower-order representations actually supplied to the models are
identical across views in all 75 comparisons using the same data splits. 
When the same trained model is evaluated on both views, 150/150 output comparisons meet the specified absolute/relative tolerance; the largest absolute logit or query-score difference is $6.20\times 10^{-6}$, and the largest absolute primary-score change is $0.00727$ points. 
Independent retraining need not produce identical scores, but empirically the gap is often small: the largest absolute three-seed mean primary-score change is $0.783$ points.
\cref{tab:baseline-invariance,tab:baseline-invariance-secondary} report all comparisons. The distinction matters for attribution: identical lower-order information guarantees identical outputs from the
same trained model in exact arithmetic, but separately trained models
can still differ because of numerical and optimization variability. The reported gap decomposition retains its common lower-order reference and does not add this variability between separate training runs to the estimated intervention effect.

\begin{table}[t]
\centering
\begingroup\small
\setlength{\tabcolsep}{3pt}
\caption{{Independent retraining of the selected lower-order baseline on original and fully perturbed views. Scores are three-seed means (percent). Max is the largest absolute seed-level score difference. Frozen compares output logits or query scores across views using the same checkpoint, reporting the largest absolute difference across both source views and three seeds.}}
\label{tab:baseline-invariance}
\begin{tabular}{@{}llrrrr@{}}
\toprule
Benchmark & Model & $P^{(0)}$ & $P^{(1)}$ & Max & Frozen \\
\midrule
\multicolumn{6}{@{}l}{\textit{Hyperedge prediction}} \\
Actor & MLP & 72.46 & 72.46 & 0.000 & $0$ \\
Cora-CA & CEGAT & 63.31 & 64.09 & 1.785 & $2.38\times 10^{-7}$ \\
DBLP-CA & CEGAT & 73.94 & 73.92 & 1.983 & $6.20\times 10^{-6}$ \\
Cora & CEGAT & 65.75 & 65.35 & 0.636 & $3.28\times 10^{-7}$ \\
Pubmed & CEGCN & 71.11 & 71.43 & 0.961 & $3.58\times 10^{-7}$ \\
Pokec & CEGCN & 59.71 & 59.61 & 0.301 & $1.19\times 10^{-7}$ \\
\multicolumn{6}{@{}l}{\textit{Hypergraph classification}} \\
IMDB-Dir-Form & CEGCN & 63.30 & 63.30 & 0.000 & $9.54\times 10^{-7}$ \\
IMDB-Dir-Genre & CEGCN & 78.63 & 78.33 & 1.471 & $1.43\times 10^{-6}$ \\
RHG-10 & CEGAT & 96.00 & 96.00 & 1.000 & $2.86\times 10^{-6}$ \\
RHG-3 & CEGCN & 98.44 & 98.44 & 0.000 & $1.43\times 10^{-6}$ \\
Steam-Player & CEGAT & 52.59 & 52.10 & 1.456 & $4.77\times 10^{-7}$ \\
Twitter-Friend & MLP & 60.31 & 60.05 & 0.763 & $4.77\times 10^{-7}$ \\
\multicolumn{6}{@{}l}{\textit{Node classification}} \\
ModelNet40 & MLP & 96.37 & 96.37 & 0.000 & $0$ \\
NTU2012 & MLP & 87.87 & 87.87 & 0.000 & $0$ \\
Actor & MLP & 85.99 & 85.99 & 0.000 & $0$ \\
Cora-CA & CEGCN & 79.57 & 79.57 & 0.000 & $3.81\times 10^{-6}$ \\
DBLP-CA & CEGCN & 89.83 & 89.83 & 0.000 & $0$ \\
Cora & CEGCN & 77.73 & 77.73 & 0.000 & $3.81\times 10^{-6}$ \\
Pubmed & MLP & 86.46 & 86.46 & 0.000 & $0$ \\
Gamers & MLP & 51.80 & 51.80 & 0.000 & $0$ \\
Pokec & MLP & 59.37 & 59.37 & 0.000 & $0$ \\
Ratings & MLP & 26.95 & 26.95 & 0.000 & $0$ \\
Trivago & CEGCN & 47.75 & 47.75 & 0.000 & $0$ \\
Walmart & CEGCN & 67.98 & 67.98 & 0.000 & $0$ \\
Yelp & MLP & 31.59 & 31.59 & 0.000 & $0$ \\
\bottomrule
\end{tabular}
\endgroup
\end{table}
\begin{table}[t]
\centering
\begingroup\small
\setlength{\tabcolsep}{3pt}
\caption{{Secondary scores from the independent baseline audit: AUROC for hyperedge prediction and Macro-F1 for hypergraph classification.}}
\label{tab:baseline-invariance-secondary}
\begin{tabular}{@{}llrrr@{}}
\toprule
Benchmark & Model & $P^{(0)}$ & $P^{(1)}$ & Max \\
\midrule
\multicolumn{5}{@{}l}{\textit{Hyperedge prediction}} \\
Actor & MLP & 76.58 & 76.58 & 0.000 \\
Cora-CA & CEGAT & 66.13 & 66.62 & 1.720 \\
DBLP-CA & CEGAT & 75.38 & 76.12 & 1.690 \\
Cora & CEGAT & 69.24 & 68.47 & 1.090 \\
Pubmed & CEGCN & 74.59 & 74.36 & 0.361 \\
Pokec & CEGCN & 56.14 & 56.02 & 0.363 \\
\multicolumn{5}{@{}l}{\textit{Hypergraph classification}} \\
IMDB-Dir-Form & CEGCN & 57.19 & 57.19 & 0.023 \\
IMDB-Dir-Genre & CEGCN & 74.35 & 73.68 & 2.025 \\
RHG-10 & CEGAT & 95.95 & 95.96 & 0.928 \\
RHG-3 & CEGCN & 98.52 & 98.52 & 0.000 \\
Steam-Player & CEGAT & 52.04 & 51.43 & 1.766 \\
Twitter-Friend & MLP & 55.11 & 54.77 & 1.017 \\
\bottomrule
\end{tabular}
\endgroup
\end{table}

\clearpage
\subsection{Node classification}\label{app:node_cls}

\begin{figure}[t!]
  \centering
  \includegraphics[width=\textwidth]{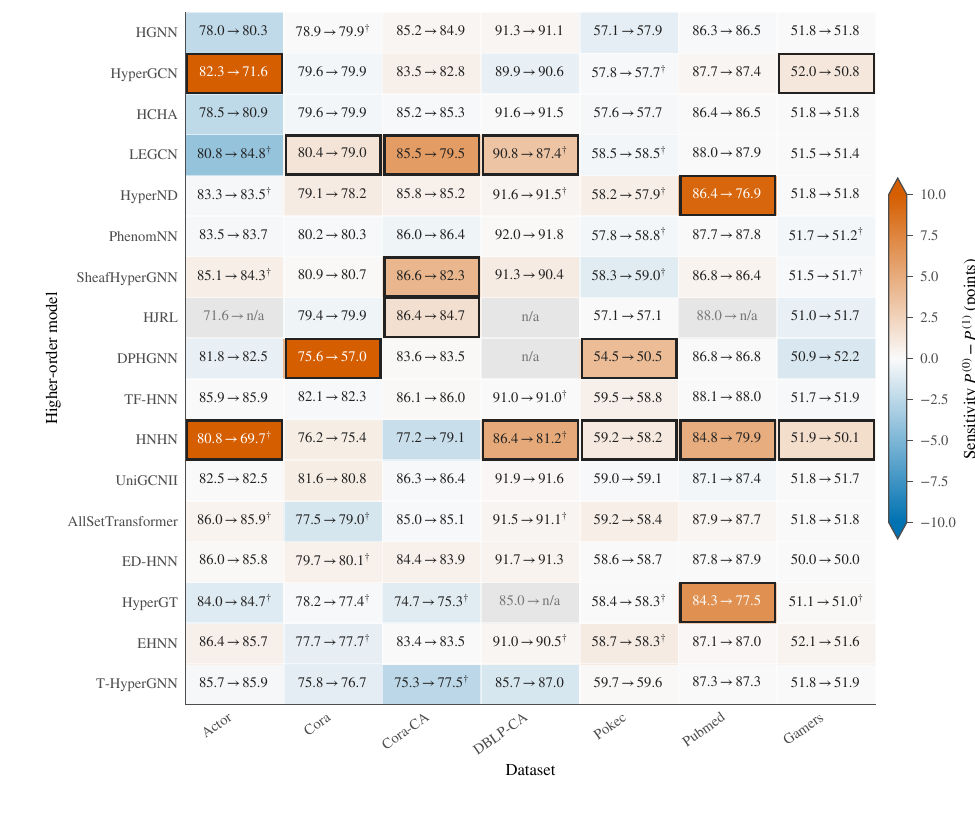}
  \caption{\textbf{Original and fully perturbed performance for every higher-order model on the first seven node-classification benchmarks.} Entries are test-accuracy percentages. Color is clipped at $\pm10$ points while the annotations retain the exact values.}
  \label{fig:all-hnn-node-a}
\end{figure}

\begin{figure}[t!]
  \centering
  \includegraphics[width=\textwidth]{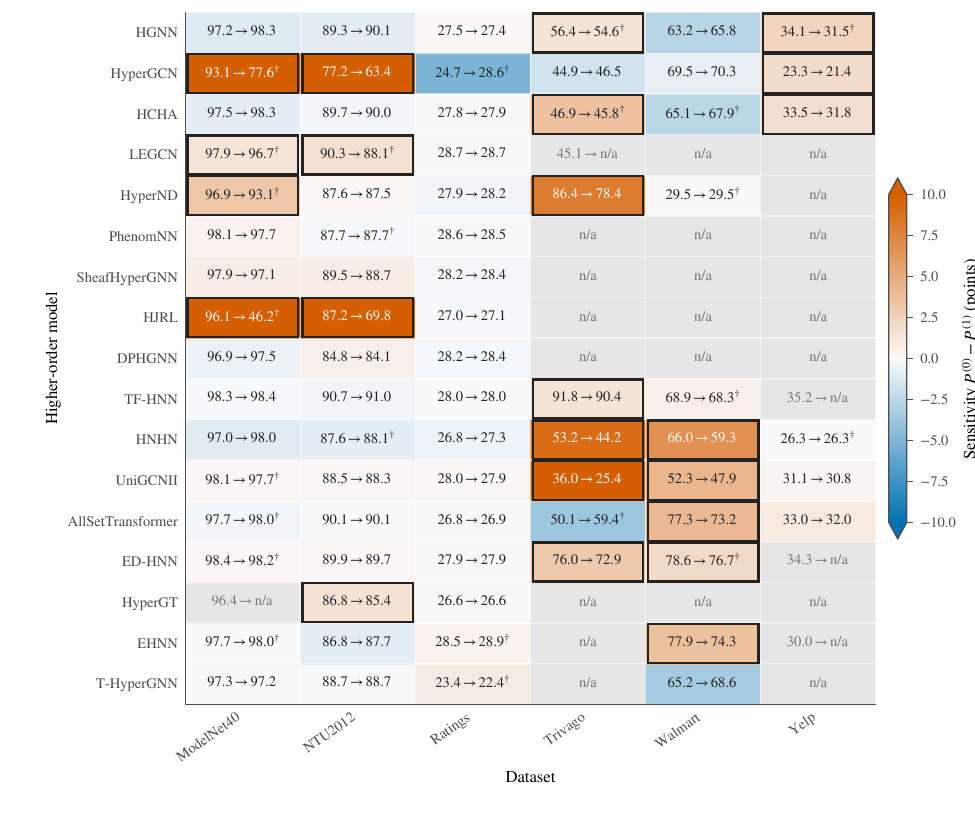}
  \caption{\textbf{Original and fully perturbed performance for every higher-order model on the remaining six node-classification benchmarks.} Styling follows
  \cref{fig:all-hnn-node-a}.}  
  \label{fig:all-hnn-node-b}
\end{figure}

\paragraph{All-model endpoint results.}
\cref{fig:all-hnn-node-a,fig:all-hnn-node-b} extend the
endpoint comparison from the higher-order model with the best original-view validation performance to all higher-order models.
Each cell gives the original performance $P^{(0)}$ and fully-perturbed performance $P^{(1)}$.
Its color encodes the value of $P^{(0)}-P^{(1)}$, and a black border marks a decrease of at least one point.
Gray cells mark an unavailable endpoint while retaining the
original performance when it is known.
A dagger indicates that fewer than all three runs using matching data splits and training seeds completed.

\begin{figure}[t!]
  \centering
  \includegraphics[width=\textwidth]{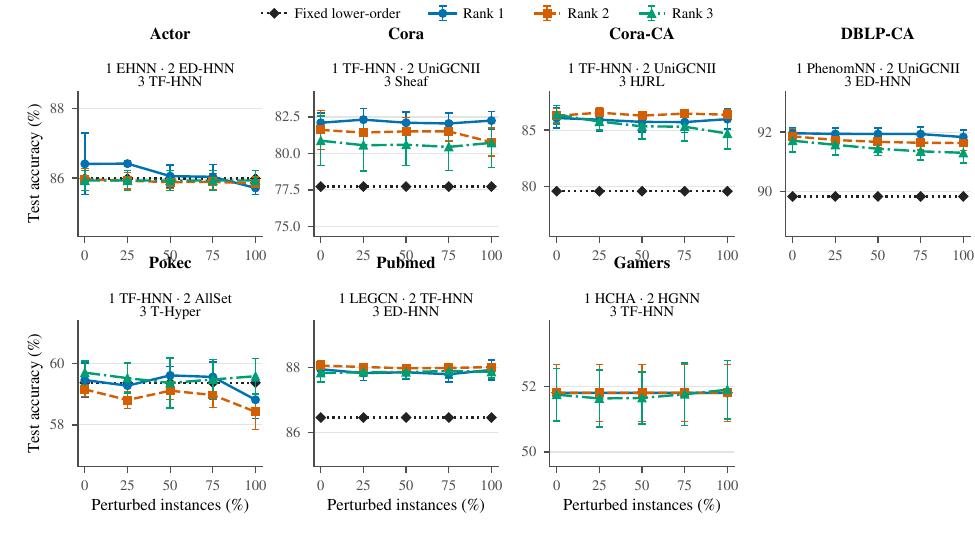}
  \caption{\textbf{Performance at five perturbation levels for the three higher-order
  models with the best original-view validation performance on the
  first seven node-classification benchmarks.} Rank colors,
  markers, and line styles are shared across panels; each panel identifies the
  corresponding model names.}
  \label{fig:top3-node-a}
\end{figure}

\begin{figure}[t!]
  \centering
  \includegraphics[width=\textwidth]{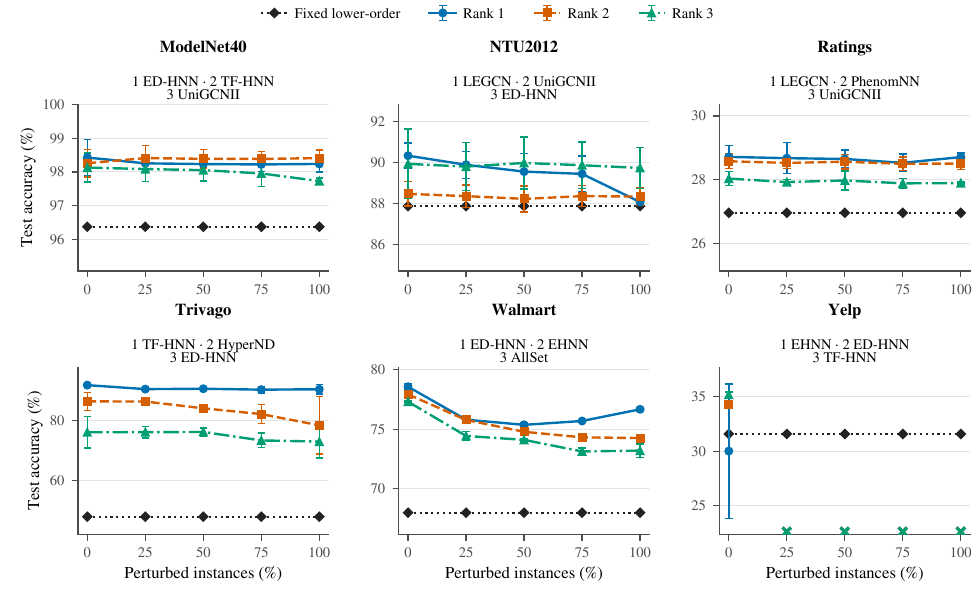}
  \caption{\textbf{Performance at five perturbation levels for the three higher-order
  models with the best original-view validation performance on the
  remaining six node-classification benchmarks.} Styling
  and aggregation follow \cref{fig:top3-node-a}.}
  \label{fig:top3-node-b}
\end{figure}

\paragraph{Performance across perturbation levels for the top three models.}
\cref{fig:top3-node-a,fig:top3-node-b} retain the top
three higher-order models ranked by validation performance on the original view. 
At 25\%, 50\%, and 75\%, we average the three sampled perturbations and then report the mean and standard deviation.  
The black dotted curve is the same selected lower-order model used in \cref{fig:real-gap-map}. 
Crosses mark unavailable cells.
Among the 36 top-three trajectories with complete endpoints, 28 change by
less than one point, and eight decrease by at least one accuracy point. 

\clearpage
\subsection{Hyperedge prediction} \label{app:edge_pred}

\begin{figure}[b!]
  \centering
  \includegraphics[width=\textwidth]{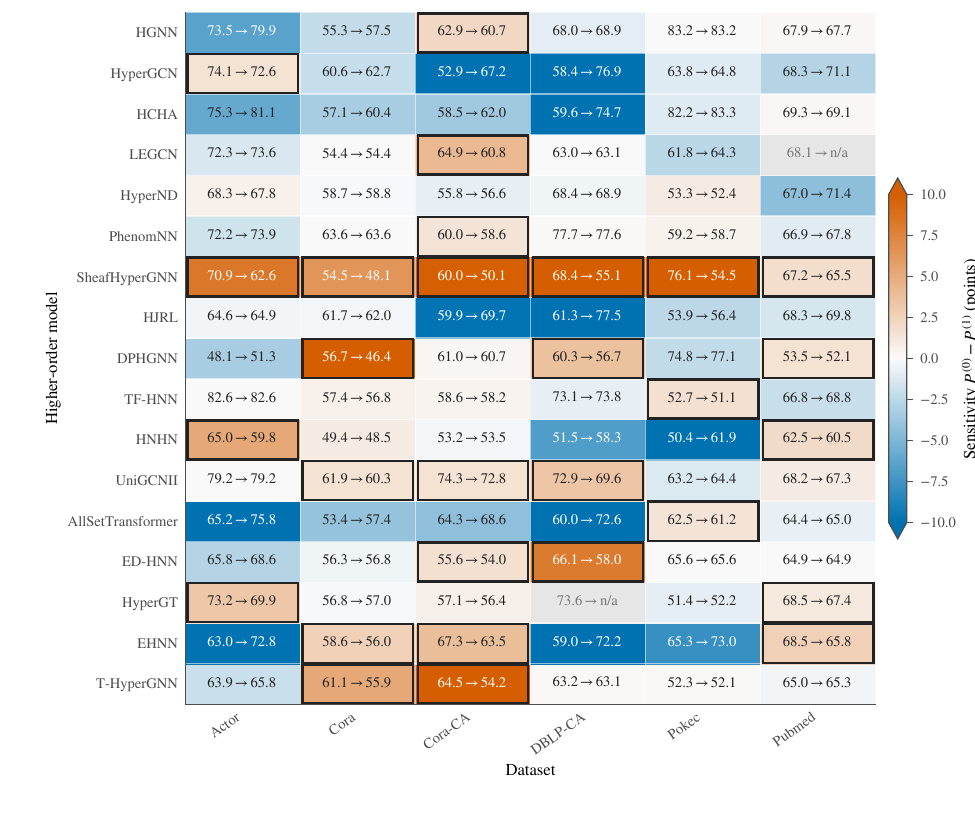}
  \caption{\textbf{Original and fully perturbed AP for every
  higher-order model supporting hyperedge prediction.} Each cell gives original $\rightarrow$ fully perturbed test performance
  ($P^{(0)}\!\rightarrow P^{(1)}$), in percent.  
  The background encodes their difference, black borders mark decreases of at least one point, and gray cells are unavailable.}
  \label{fig:all-hnn-edge}
\end{figure}

\paragraph{All-model endpoint results.}
As shown in \cref{fig:all-hnn-edge}, the endpoint response is not unidirectional.
Of 100 available cells, 29 decrease by at least one AP point, 35 improve by at least one point, and 36 change by less than one point.

\begin{figure}[t!]
  \centering
  \includegraphics[width=\textwidth]{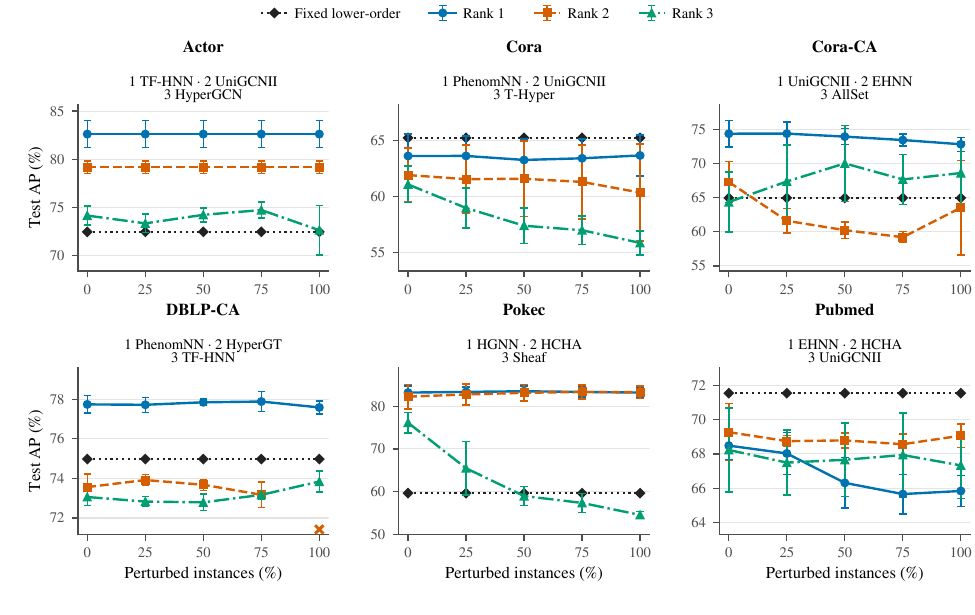}
  \caption{\textbf{Performance at five perturbation levels for the three higher-order models with the best original-view validation performance on all six hyperedge-prediction benchmarks.} The black dotted line is the same selected lower-order model used in \cref{fig:real-gap-map}.}
  \label{fig:top3-edge}
\end{figure}

\paragraph{Performance across perturbation levels for the top three models.}
For each benchmark, we select the three higher-order models with the
best validation performance on the original view.
As shown in \cref{fig:top3-edge}, of the 17 top-three trajectories with complete endpoints, seven decrease by at least one AP point, eight change by less than one point, and two improve by at least one point.  
The responses can disagree even within one benchmark:
on Cora-CA, two selected models decrease while one improves, and on Pokec the three models respectively remain stable, improve, and decrease.

\begin{figure}[t!]
  \centering
  \includegraphics[width=\textwidth]{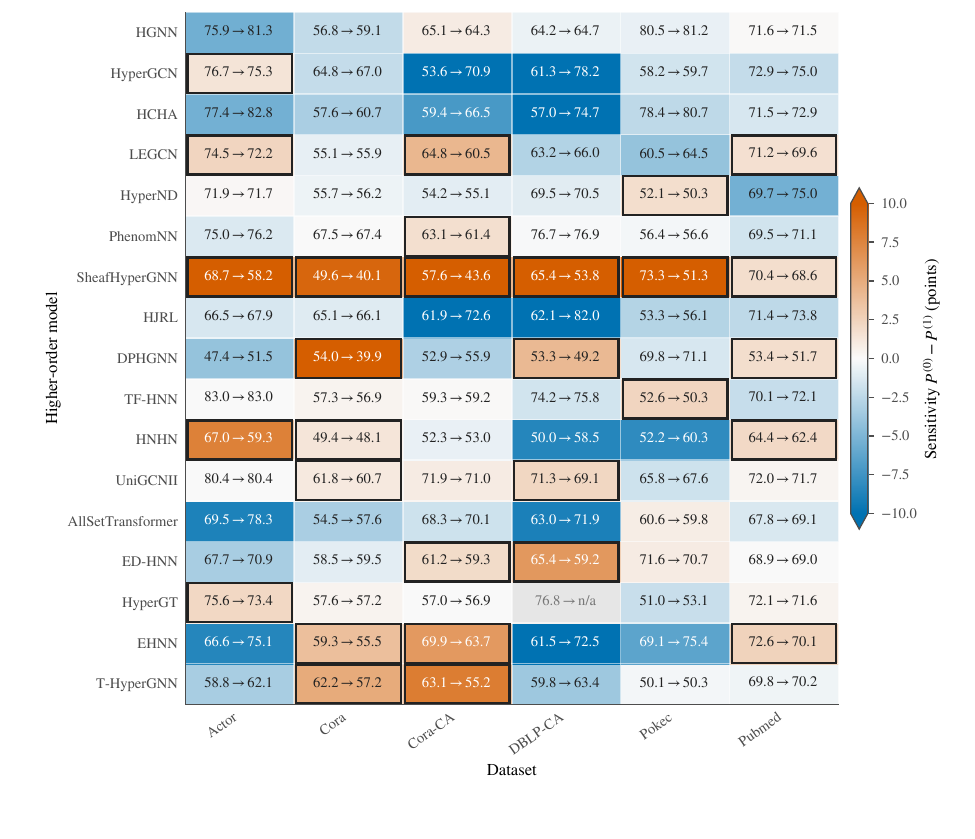}
  \caption{\textbf{Original and fully perturbed performance for all higher-order models using AUROC.} Each cell shows original
  $\rightarrow$ fully perturbed performance (percent), with checkpoints selected by validation AUROC. Color denotes the paired decrease in points. Outlines mark decreases of at least one point. Gray cells indicate unavailable comparisons; a dagger marks fewer than three paired seeds.}
  \label{fig:secondary-edge-endpoints}
\end{figure}

\begin{figure}[t!]
  \centering
  \includegraphics[width=\textwidth]{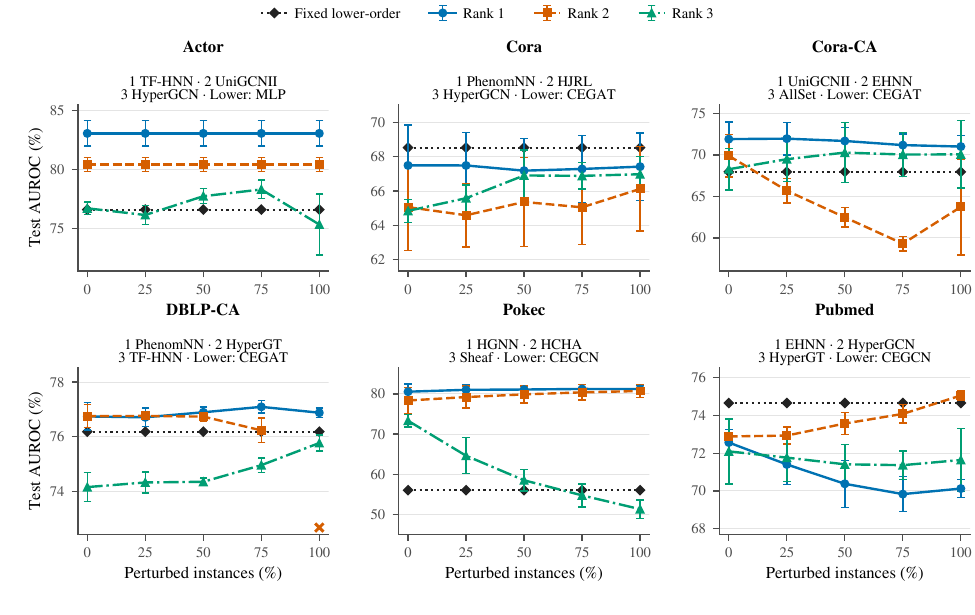}
  \caption{\textbf{Five-level AUROC trajectories.} The top three HNNs and the
  lower-order comparator are selected by original-view validation AUROC;
  checkpoints at every level are also selected by validation AUROC.
  HNN error bars show sample SD across available seed means (three seeds planned),
  after averaging the three perturbation trajectories within each seed at
  intermediate levels. Crosses mark unavailable cells.}
  \label{fig:secondary-edge-trajectories}
\end{figure}

\paragraph{Secondary metric: AUROC.}
\cref{fig:secondary-edge-endpoints,fig:secondary-edge-trajectories} report all-model endpoints and top-three trajectories using AUROC.

\clearpage
\subsection{Hypergraph classification}\label{app:hg_cls}

\begin{figure}[b!]
  \centering
  \includegraphics[width=\textwidth]{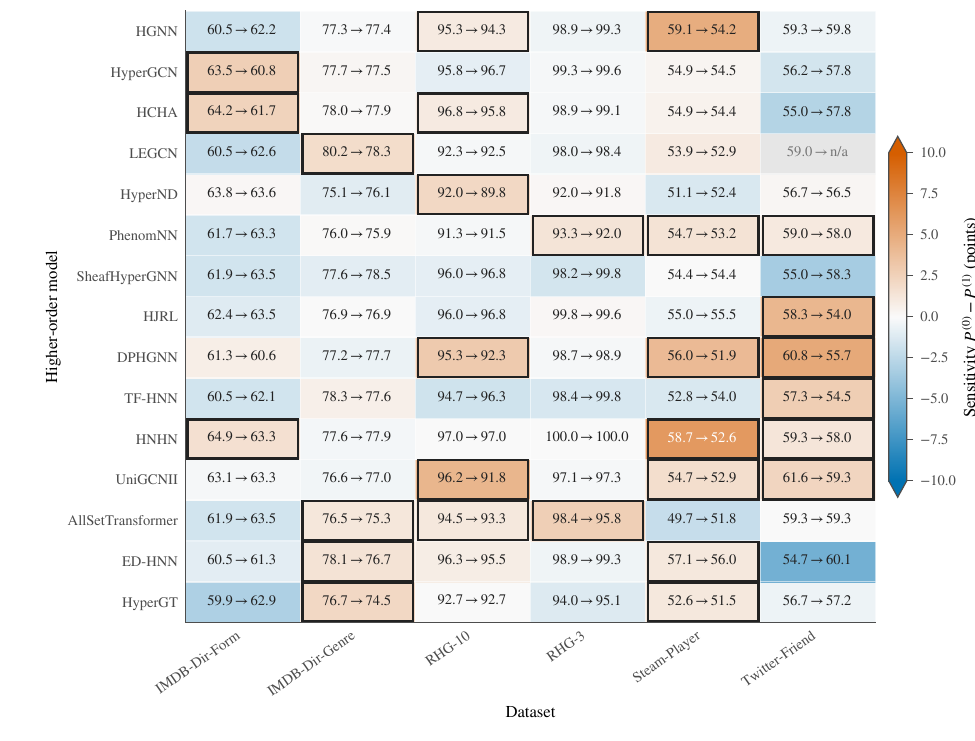}
  \caption{\textbf{Original and fully perturbed accuracy for every
  higher-order model supporting hypergraph classification.} Each cell gives original $\rightarrow$ fully perturbed test performance
  ($P^{(0)}\!\rightarrow P^{(1)}$), in percent.  The background encodes their difference, black borders mark decreases of at least one point, and gray cells are unavailable.}
  \label{fig:all-hnn-hg}
\end{figure}

\paragraph{All-model endpoint results.}
As shown in \cref{fig:all-hnn-hg}, among the 89 available cells, 28 decrease by at least one accuracy point, 19 improve by at least one point, and 42 change by less than one point.
This model-level heterogeneity is especially visible on
Steam-Player and Twitter-Friend, where decreases of at least one accuracy point coexist with stable or improving models.

\begin{figure}[t!]
  \centering
  \includegraphics[width=\textwidth]{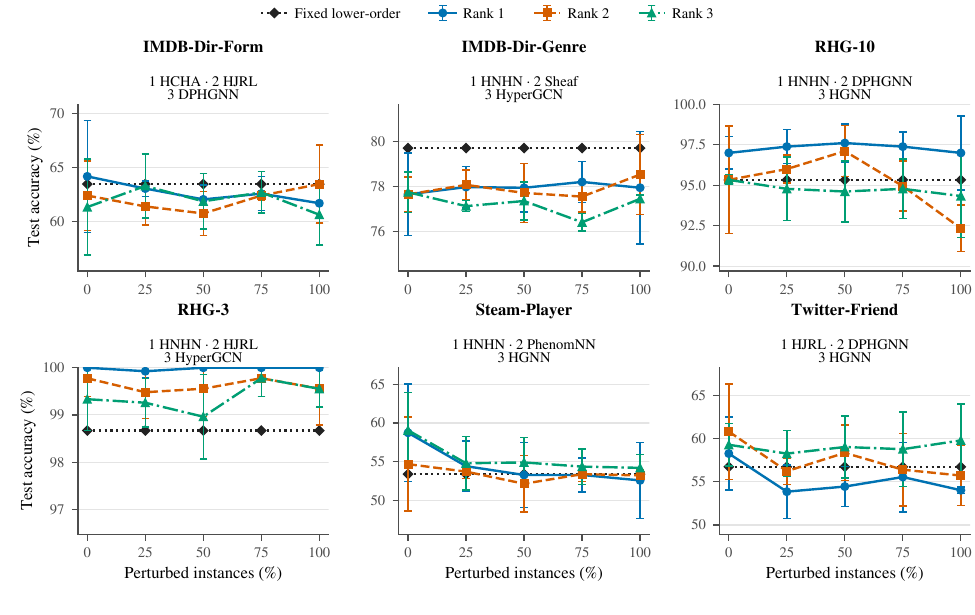}
  \caption{\textbf{Performance at five perturbation levels for the three higher-order
  models with the best original-view validation performance on
  all six hypergraph-classification benchmarks.} The black dotted line is the same selected lower-order model used in \cref{fig:real-gap-map}.}
  \label{fig:top3-hg}
\end{figure}

\paragraph{Performance across perturbation levels for the top three models.}
For each benchmark, we select the three higher-order models with the
best validation performance on the original view.
As shown in \cref{fig:top3-hg}, across the 18 complete top-three trajectories, eight models decrease by at least one accuracy point, nine change by less than one point, and one improves.  
All three selected models decrease on Steam-Player, whereas all
three remain stable on RHG-3.
Moreover, six trajectories have an intermediate mean more than one point outside the range spanned by their two
endpoint means.
Partial perturbations therefore reveal model-specific and
occasionally non-monotone responses.

\begin{figure}[t!]
  \centering
  \includegraphics[width=\textwidth]{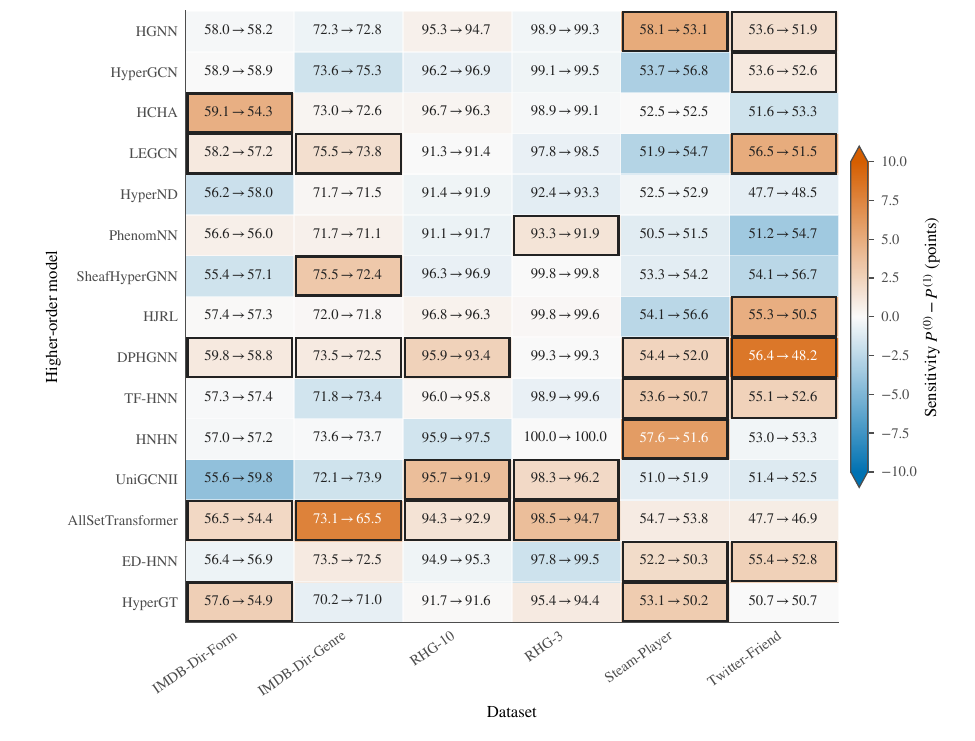}
  \caption{\textbf{Original and fully perturbed performance for all higher-order models using Macro-F1.} Original $\rightarrow$ fully perturbed performance (percent), with checkpoints selected by validation Macro-F1. Color denotes the paired decrease. Outlines mark decreases of at least one point. Gray cells indicate unavailable comparisons; a dagger marks fewer than three paired seeds.}
  \label{fig:secondary-hg-endpoints}
\end{figure}

\begin{figure}[t!]
  \centering
  \includegraphics[width=\textwidth]{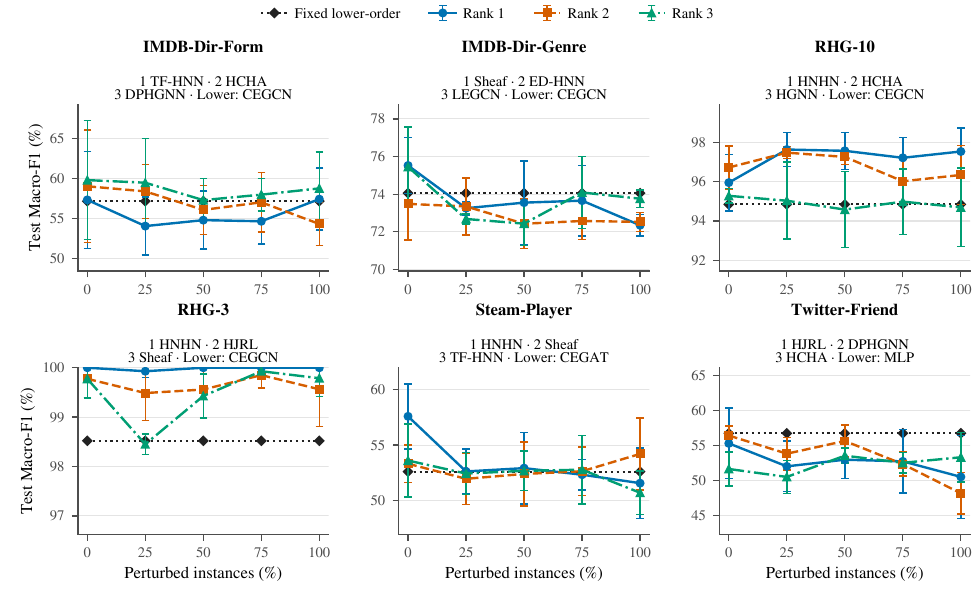}
  \caption{\textbf{Five-level Macro-F1 trajectories.} The top three HNNs and
  the lower-order comparator are selected by original-view validation Macro-F1;
  checkpoints at every level are also selected by validation Macro-F1.
  HNN error bars show sample SD across available seed means (three seeds planned),
  after averaging the three perturbation trajectories within each seed at
  intermediate levels. Crosses mark unavailable cells.}
  \label{fig:secondary-hg-trajectories}
\end{figure}

\paragraph{Secondary metric: Macro-F1.}
\cref{fig:secondary-hg-endpoints,fig:secondary-hg-trajectories} report
all-model endpoints and top-three trajectories using Macro-F1.

\clearpage

\clearpage
\section{Additional Analyses of Retained Gaps (Supplementing \cref{sec:remaining_gaps})}
\label{app:remaining-gap-cases}

\subsection{Detailed benchmark-wise analysis}
\label{app:remaining-gap-analysis}

We attempted a lower-order diagnosis for all 14 largely retained gaps in \cref{fig:real-gap-map}: eight node-classification, four hyperedge-prediction, and two hypergraph-classification comparisons.

\paragraph{Walmart: current operator and diffusion controls.}
The main ladder uses the same CEGCN
architecture, inputs, and training settings at every stage. The successive operators use binary connections, pair counts, and
pair counts together with structural self weights:
\[
 \mathcal N(B+I)\ \longrightarrow\ \mathcal N(C+I)\
 \longrightarrow\ \mathcal N\!\left(C+\operatorname{diag}(d+2\mathbf1)\right),
\]
where $B$ is binary off-diagonal adjacency, $C$ contains pair counts, $d=C\mathbf1$, and
$\mathcal N(A)=D_A^{-1/2}AD_A^{-1/2}$. 
For the incidence matrix $H$ recording which nodes belong to each pair,
$H\operatorname{diag}(1/|e|)H^\top+I
=\tfrac12C+\tfrac12\operatorname{diag}(d)+I$;
multiplying by two leaves symmetric normalization unchanged.
Accuracy rises from $67.98$ to $68.15$ to $74.56$.
Input diffusion $W^kX$ then uses the same operator $W$ incorporating pair counts and structural self weights.
Mean validation accuracy selects $k=1$, reaching $76.32$.
The depth table below retains the full sweep, including deterioration
at greater depths; ties in validation favor the smallest depth.

\paragraph{Trivago: fixed-architecture diffusion sweeps.}
The current main panel applies $W^kX$ before CEGCN's two learned layers,
which retain binary propagation $P$. Both $P$ and $W$ include unit self
loops and symmetric normalization; $W$ additionally uses pair counts and
structural self weights. We report
$k\in\{0,1,2,4,8,16,32,64,128,256\}$.
The $k=0$ path returns the original features without arithmetic.

The supporting binary-input sweep $P^kX$ reaches $77.64\pm0.25$ at
$k=64$, versus $47.75\pm0.52$ at zero extra steps, but falls to $68.90\pm0.45$ at $k=256$. 
The weighted operator performs worse at (k=64) but better at (k=256);
the $0.68$-point difference between the selected sweeps changes both
operator and depth. \cref{tab:counts-self-depth} preserves every depth.

\begin{table}[t]
\centering
\small
\caption{Input-diffusion depth sensitivity for the Walmart and Trivago controls. Each entry reports test accuracy as mean $\pm$ sample SD. Diffusion is applied to the input features before the fixed CEGCN classifier; $k=0$ denotes no extra diffusion. $W$ is the
operator incorporating pair multiplicities and structural self weights, while $P$ is the binary CEGCN operator.}
\label{tab:counts-self-depth}
\begin{tabular}{rccc}
\toprule
Extra steps $k$ & Walmart: $W^k$ & Trivago: $P^k$ & Trivago: $W^k$ \\
\midrule
0 & $74.56 \pm 0.19$ & $47.75 \pm 0.52$ & $47.75 \pm 0.52$ \\
1 & $76.32 \pm 0.23$ & $52.90 \pm 0.23$ & $50.05 \pm 0.50$ \\
2 & $75.68 \pm 0.37$ & $55.72 \pm 0.28$ & $51.98 \pm 0.07$ \\
4 & $72.20 \pm 0.58$ & $59.84 \pm 0.35$ & $54.72 \pm 0.28$ \\
8 & $64.43 \pm 0.47$ & $64.95 \pm 0.22$ & $58.49 \pm 0.37$ \\
16 & $49.78 \pm 0.34$ & $70.45 \pm 0.15$ & $63.49 \pm 0.30$ \\
32 & $43.97 \pm 0.37$ & $74.95 \pm 0.45$ & $69.35 \pm 0.22$ \\
64 & $43.16 \pm 0.29$ & $77.64 \pm 0.25$ & $74.41 \pm 0.15$ \\
128 & $43.17 \pm 0.27$ & $76.62 \pm 0.89$ & $77.68 \pm 0.41$ \\
256 & $43.17 \pm 0.24$ & $68.90 \pm 0.45$ & $78.32 \pm 0.65$ \\
\bottomrule
\end{tabular}
\end{table}

\paragraph{Trivago: classifier controls.}
At $k=64$ and $256$, controls change one component at a time: they add LayerNorm after the hidden ReLU,
replace binary propagation with $W$ in both learned layers, or remove
learned-layer propagation while retaining their two affine maps.
A single frozen input tensor per depth is shared across variants; other settings remain fixed. LayerNorm adds learned scale and bias.
It improves accuracy by $13.55$ and $12.60$ points, reaching $87.96$ and $90.92$.
Weighted propagation throughout reduces accuracy by $0.67$ and $0.95$ points; removing classifier propagation reduces it by $1.47$ and $2.49$.

\begin{table}[t!]
\centering
\small
\caption{Classifier controls on Trivago with the same propagated input features $W^kX$, where $W$ incorporates pair multiplicities and structural self weights. Entries report test accuracy as mean $\pm$ sample SD. The matched baseline retains binary
propagation in the two learned CEGCN layers; each subsequent row changes one classifier component while keeping the remaining settings fixed.}
\label{tab:trivago-classifier-controls}
\begin{tabular}{lcc}
\toprule
Classifier change & $k=64$ & $k=256$ \\
\midrule
Matched baseline & $74.41 \pm 0.15$ & $78.32 \pm 0.65$ \\
+ LayerNorm & $87.96 \pm 0.96$ & $90.92 \pm 0.75$ \\
Weighted propagation throughout & $73.74 \pm 0.19$ & $77.37 \pm 0.91$ \\
Remove classifier propagation & $72.94 \pm 0.43$ & $75.83 \pm 0.91$ \\
\bottomrule
\end{tabular}
\end{table}

\paragraph{Pokec: matched policy and activation controls.}
The identity-policy CEGCN baseline is $59.71$;
the HGNN original/perturbed references are $83.18$/$83.19$.
CEGCN runs using the same zero-message policy as HGNN yield $78.22$; changing only the hidden ReLU to HGNN's ELU reaches $80.14$, a paired gain of $1.92$ points.
Query nodes, offsets, labels, encoder, and scorer settings are held fixed.
ELU adds no parameters. 
\cref{tab:pokec-complement} also retains
LayerNorm's $+1.48$-point effect and count-mean
propagation's $+0.26$.

\begin{table}[t!]
\centering
\small
\caption{Complementary encoder controls on Pokec after matching the
higher-order reference's cold-node policy. Entries report test AP as mean $\pm$ sample SD. Each subsequent row adds one encoder
change to the CEGCN baseline using the same zero-message policy as HGNN.}
\label{tab:pokec-complement}
\begin{tabular}{lc}
\toprule
Encoder change & Test AP (\%) \\
\midrule
Matched cold-node policy & $78.22 \pm 1.73$ \\
+ ELU & $80.14 \pm 2.11$ \\
+ LayerNorm & $79.70 \pm 0.48$ \\
+ count-mean propagation & $78.48 \pm 1.73$ \\
\bottomrule
\end{tabular}
\end{table}

\paragraph{Other hyperedge-prediction cases.}
The remaining hyperedge-prediction cases provide complementary evidence.
On Actor, the selected TFHNN configuration uses node features without
incidence-based message passing, so its unchanged $10.15$-point advantage
after the intervention does not require information from the hypergraph
structure.
For Cora-CA and Pokec, we also evaluate separately constructed pairwise
lower-order controls that retain pair multiplicities and average pairwise
messages. The Cora-CA control
matches the fully perturbed score of UniGCNII, and the Pokec control matches
the fully perturbed score of HGNN.
Together, these cases show that several retained gaps can be matched by
pairwise or feature-only mechanisms.

\paragraph{Additional node-classification analyses.}
Four further cases suggest that pairwise processing and model design can
help explain retained advantages, although the controls provide different
levels of evidence.
On Cora-CA, matched comparisons attribute an $8.49$-point accuracy gain
to pairwise weighting and a further $2.96$ points to architecture and
diffusion choices. Adding TFHNN's native non-pairwise component instead
changes accuracy by $-0.45$ points. Thus, in these comparisons, the gains
come from lower-order processing rather than the native grouping component.
On DBLP-CA, disabling PhenomNN's propagation reduces accuracy by about
$5.1$ points. Pairwise operators using count-mean or pair-diagonal
normalization also outperform binary clique expansion. These results
indicate that both propagation and the choice of pairwise normalization
matter.
On Cora, removing diffusion from the fully pairwise TFHNN control reduces
accuracy by $4.51$ points, close to the selected model's $4.52$-point
remaining advantage. On ModelNet40, restricting EDHNN to feature-only
processing reduces accuracy by about $1.9$ points, also close to its
remaining advantage; separate comparisons identify pairwise weighting
and normalization as the largest source of improvement among the tested
changes. These ablations show that propagation helps the tested models.
However, the similarity between the performance losses and the remaining
gaps does not establish that propagation alone explains those gaps.

\paragraph{Pairwise controls for hypergraph classification.}
On RHG-10, the Pair-Multiplicity control outperforms CEGAT by $2.25$
points, exceeding HNHN's $1.72$-point advantage at the fully perturbed
endpoint. On RHG-3, Pair-SAGE-Mean outperforms CEGCN by $1.03$ points,
recovering about $78\%$ of HNHN's $1.33$-point remaining advantage.
Both controls use only weighted clique-expansion information. They
therefore demonstrate that lower-order models can recover all or much
of these advantages. These additional controls were outside the
original model-selection pool and do not replace the selected model
pairs reported in the main results.

\paragraph{Pubmed and Ratings: feature processing versus propagation.}
To examine what sustains LEGCN's advantage, we disable messages between
distinct node--hyperedge incidences (a node's memberships in hyperedges)
while retaining its layers, hidden
dimensions, nonlinearities, training settings, and feature paths.
This control tests how much of the advantage remains without information
being passed between incidences. The resulting self-only model outperforms the selected MLP baseline by $1.29$ accuracy points on Pubmed, while restoring propagation adds only $0.09$ points.
On Ratings, the corresponding gains are $1.22$ and $0.21$ points.
Thus, most of LEGCN's advantage in these comparisons remains without passing information between incidences, suggesting that differences in feature processing and model configuration account for much of the gap.

\paragraph{What the 17 HNNs compute at the fully perturbed endpoint.}
\label{app:pair-endpoint-models}
The complete input at the fully perturbed endpoint contains two-node hyperedge instances and
retained singleton instances. All endpoint incidences are determined by
weighted pairwise connections and singleton multiplicities.
This removes the original grouping into larger hyperedges, but does not make every model an ordinary GCN. Models can still differ in normalization, nonlinear messages,
representations of nodes and edges, paths that reintroduce the initial
features (restart), and how representations are converted into predictions
(readout).
\cref{tab:pair-endpoint-models} summarizes what the 17 HNNs do at this endpoint.

For the following operator derivations, let $s_v$ denote the number of
retained singleton hyperedge instances at node $v$, and let $s$ collect
these counts. This notation describes the fixed background incidences
retained by the implementation. Let $C$ denote off-diagonal pair
multiplicities and $d=C\mathbf1$. Before any additional model-specific
self loops, write
\[
 K=\tfrac12\bigl(C+\operatorname{diag}(d)\bigr)
   +\operatorname{diag}(s),
 \qquad q=d+s,\qquad D_q=\operatorname{diag}(q).
\]
On rows with $q_v>0$, mean node--edge--node aggregation becomes
\begin{equation}
 Q_{\mathrm{mean}}=D_q^{-1}K,
 \qquad Q_{\mathrm{sym}}=D_q^{-1/2}KD_q^{-1/2}.
 \label{eq:pair-endpoint-mean}
\end{equation}
Inverse degrees are set to zero where $q_v=0$ before applying any explicit
cold-node policy. When $s=0$, these expressions reduce to
$Q_{\mathrm{mean}}=\tfrac12I+\tfrac12D_d^{-1}C$ on nodes with $d_v>0$.
Thus even unweighted pair instances induce a structural self contribution
when messages travel to a pair and back. Singleton instances contribute
only self messages and also enter the incidence-degree normalization.

HGNN uses the row-normalized form and HCHA the symmetric form; both instantiate the same convolution class without enabling its attention branch. The layer still retains its learned feature map, bias, and inter-layer activation.

UniGCNII's mean aggregation has the same $Q_{\mathrm{mean}}$ step, followed by its configured normalization, restart, and learned residual map.
TFHNN instead diffuses inputs using $\mathcal N(K+I)$, with its
configured restart and depth, before applying its decoder. PhenomNN uses both $\mathcal N(C+\operatorname{diag}(q)+I)$ and
$\mathcal N(K+I)$ inside its iterative update. 
These distinctions explain why preserving pair counts and singleton multiplicities does not make the architectures or their performance interchangeable.

\setlength{\LTcapwidth}{\textwidth}
{
\begin{longtable}{@{}p{0.17\textwidth}p{0.43\textwidth}p{0.34\textwidth}@{}}
\caption{Computations of the 17 HNNs on fully perturbed endpoints.
Pair and singleton multiplicities are retained in the endpoint inputs.
The table identifies propagation, normalization, and architectural choices;
singleton handling and additional self loops follow each implementation.}
\label{tab:pair-endpoint-models}\\
\toprule
Model & Computation on the fully perturbed endpoint & Weighting and architectural details \\
\midrule
\endfirsthead
\toprule
Model & Computation on the fully perturbed endpoint & Weighting and architectural details \\
\midrule
\endhead
\bottomrule
\endfoot

HGNN
& Learned feature maps with $Q_{\mathrm{mean}}$ propagation and ELU between layers.
& Averaging through each pair induces both neighbor and self messages; repeated pairs and singleton incidences enter aggregation and degree normalization. \\
\addlinespace

HCHA
& The released implementation uses the same convolution class as HGNN, with $Q_{\mathrm{sym}}$ propagation.
& Symmetric degree normalization replaces row normalization; the attention branch is not enabled by the model constructor. \\
\addlinespace

HyperGCN
& Graph convolution on the graph constructed from endpoint hyperedges by the extremal-node/mediator routine.
& Edge accumulation, normalization, explicit self loops, singleton handling, and feature-dependent tie handling follow the released implementation. \\
\addlinespace

TFHNN
& Fixed input diffusion using $\mathcal N(K+I)$ and the configured restart, followed by an MLP decoder.
& Pair counts, singleton multiplicities, and degree-dependent self weights enter diffusion; restart weight $\alpha=1$ gives a feature-only configuration, as on Actor hyperedge prediction. \\
\addlinespace

UniGCNII
& Mean aggregation through endpoint hyperedges, optional feature normalization, restart to the initial representation, and learned residual maps.
& Uses $Q_{\mathrm{mean}}$; the restart path can preserve features even when the structural aggregate is zero. \\
\addlinespace

PhenomNN
& Iterative propagation with two normalized matrices determined by pair and singleton incidences, feature injection, and ReLU, between an encoder and decoder.
& Both $C+\operatorname{diag}(q)+I$ and $K+I$ enter; setting the propagation depth to zero retains feature processing. \\
\addlinespace

HyperND
& Nonlinear diffusion using power means of the features in each endpoint hyperedge, global normalization, and restart, followed by an MLP.
& Repeated pairs and singleton incidences affect messages and incidence degrees; the power-mean operation is not generally a fixed linear operator. \\
\addlinespace

HNHN
& Alternating learned node-to-hyperedge and hyperedge-to-node transformations with configured degree weights.
& An optional intermediate nonlinearity acts on hyperedge states; pair and singleton multiplicities enter aggregation and degree weighting. \\
\addlinespace

AllSetformer
& Attention-based aggregation from each hyperedge's one or two nodes into a hyperedge state, then from incident hyperedge states to a node.
& Learned attention, inter-stage nonlinearities, and configured self and residual paths remain. \\
\addlinespace

EDHNN
& Learned messages combine each receiving node's state with an aggregation of its hyperedge's node features.
& Receiver-dependent nonlinear messages and restart remain; repeated pair instances contribute through the configured aggregation. \\
\addlinespace

LEGCN
& GCN on incidence states, connected when they share a node or a hyperedge; outputs are averaged back to nodes.
& Line-graph normalization and self loops remain; repeated pairs and singletons give distinct incidence states. \\
\addlinespace

HJRL
& Coupled node and hyperedge-state updates through normalized incidence and node--node and hyperedge--hyperedge products.
& Both state spaces, their normalization, and feature initialization remain; repeated pairs retain separate states. \\
\addlinespace

SheafHyperGNN
& Diffusion with learned channel-wise incidence restriction maps, using node and hyperedge attributes.
& Learned transport and block normalization remain; pair multiplicity and hyperedge-attribute initialization affect the computation. \\
\addlinespace

DPHGNN
& Fusion of hypergraph, clique, and star-derived spatial/spectral branches, all constructed from pair and singleton incidence.
& Each branch retains its normalization and self paths; the star branch represents each endpoint hyperedge as an additional vertex. \\
\addlinespace

EHNN
& Order- and overlap-conditioned equivariant computations over node and hyperedge states, with configured broadcast or attention terms.
& Size-one and size-two hyperedges and their overlaps remain represented; the computation need not reduce to a fixed node propagation matrix. \\
\addlinespace

HyperGT
& Transformer computation over node and hyperedge tokens with incidence-derived positional encodings and structural biases.
& Token self loops, attention, and positional parameters remain; computation is not restricted to local graph convolution. \\
\addlinespace

T-HyperGNN
& Configured tensor-product messages over endpoint features; repeated-index products are used when an endpoint hyperedge is smaller than the tensor order.
& Messages can remain nonlinear on pairs; self paths and repeated pair contributions follow the neighborhood construction. \\

\end{longtable}
}

\subsection{O/D/L additions shared across datasets and primary-metric results}
\label{app:cegcn-additions}

The full study compares official CEGCN, +O, +D,
+L, and +O+D+L on all 13 node-classification and six hypergraph-classification
datasets, and under both identity and zero-message policies on all six
hyperedge-prediction datasets.
The same diffusion depths, mixing weights, and normalization settings
are used across datasets.
They were motivated by the observed case-study results and evaluated
without further hyperparameter tuning.

\paragraph{O: multiplicities and structural self weights.}
Let $B$ denote binary off-diagonal adjacency, $C$ the off-diagonal pair
multiplicities, and $d=C\mathbf1$. Here $s$ records retained singleton
multiplicities as fixed background incidences. We separate the operator
change into
\textbf{O$_1$}, replacing $B$ by $C$, and \textbf{O$_2$}, replacing
unit self weights by the structural diagonal, including retained
singleton contributions, after pair multiplicities have been added by
O$_1$:
\[
 P_{\mathrm{binary}}=\mathcal N(B+I),\qquad
 P_{\mathrm{O}_1}=\mathcal N(C+I),\qquad
 P_{\mathrm{O}_1+\mathrm{O}_2}
 =\mathcal N\!\left(C+\operatorname{diag}(d+2s+2\mathbf1)\right).
\]
Here $\mathcal N(A)=D_A^{-1/2}AD_A^{-1/2}$, with
$D_A=\operatorname{diag}(A\mathbf1)$.
The existing label \textbf{O} denotes the combined O$_1$+O$_2$ operator.
The term $2s$ accounts for singleton incidences in the implemented
operator and vanishes when no singleton hyperedges are present.
Changes along this ladder depend on their order: O$_2$ changes the
normalization of neighbor messages as well as self messages.

\paragraph{Matched decomposition of O across benchmarks.}
\label{app:operator-controls}
We compare binary CEGCN, +O$_1$, and +O$_1$+O$_2$ on all 25 benchmarks,
using both identity and zero-message policies for hyperedge prediction.
D and L are disabled. Learned-layer architectures, activations, prediction
heads, splits, and training settings are matched within each comparison;
each operator variant is trained separately.
For each seed, the O$_1$ increment is the counts-only score minus binary
CEGCN, and the O$_2$ increment is the full-O score minus counts-only CEGCN.
We report original-view scores and the means and sample standard
deviations of these paired changes.

\paragraph{D: residual diffusion.}
Let $B$ be the exact normalized binary clique-expansion operator used by
official CEGCN.  Before any trainable transformation, combine the original features with
features propagated for different numbers of steps:

\[
 \widetilde X=0.8X+\frac{0.2}{6}
 \sum_{k\in\{1,2,4,8,16,32\}}Q^kX,
 \qquad
 Q=\begin{cases}B,&\text{CEGCN+D},\\P,&\text{CEGCN+O+D}.\end{cases}
\]

\paragraph{L: classifier normalization.}
L inserts LayerNorm after the hidden activation and before dropout, with
$\epsilon=10^{-5}$ and learned affine scale and bias. O and D add no
trainable parameters.

\paragraph{Hyperedge-prediction policies.}
Cold nodes do not occur in any positive training hyperedge.
The identity policy retains their transformed features; the zero-message
policy removes their feature-dependent messages at each learned layer.

\paragraph{Full primary-metric results.}
\cref{tab:stronger-node_cls-scores,tab:stronger-node_cls-deltas,tab:node-cls-o1o2,tab:stronger-hg_cls-scores,tab:stronger-hg_cls-deltas,tab:hg-cls-o1o2-primary,tab:stronger-edge_pred_identity-scores,tab:stronger-edge_pred_identity-deltas,tab:operator-edge-pred-identity-primary,tab:stronger-edge_pred_zero-scores,tab:stronger-edge_pred_zero-deltas,tab:operator-edge-pred-zero-primary}
report absolute scores and paired changes for every benchmark under the five
main recipes. For hyperedge prediction, changes are measured against
CEGCN using the same policy for cold nodes; \cref{tab:stronger-edge-policy} isolates identity versus
zero-message policies for cold nodes within each recipe.

\begin{table}[p]
\centering
\small
\caption{Node classification. Mean $\pm$ sample SD. The metric is accuracy.}
\label{tab:stronger-node_cls-scores}
\setlength{\tabcolsep}{4pt}
\begin{tabular}{@{}lrrrrr@{}}
\toprule
Dataset & CEGCN & +O & +D & +L & +O+D+L \\
\midrule
ModelNet40 & $92.75 \pm 0.56$ & $97.85 \pm 0.42$ & $92.63 \pm 0.66$ & $92.61 \pm 0.64$ & $97.45 \pm 0.47$ \\
NTU2012 & $84.76 \pm 0.64$ & $88.87 \pm 0.72$ & $84.63 \pm 0.75$ & $84.10 \pm 1.05$ & $88.87 \pm 1.30$ \\
Actor & $67.86 \pm 0.42$ & $82.26 \pm 0.53$ & $67.77 \pm 0.47$ & $68.23 \pm 0.31$ & $82.44 \pm 0.36$ \\
Cora-CA & $79.57 \pm 2.79$ & $84.59 \pm 0.45$ & $79.67 \pm 3.13$ & $79.76 \pm 2.74$ & $84.64 \pm 0.15$ \\
DBLP-CA & $89.83 \pm 0.15$ & $91.19 \pm 0.35$ & $89.81 \pm 0.07$ & $90.24 \pm 0.17$ & $91.73 \pm 0.19$ \\
Cora & $77.73 \pm 2.23$ & $78.52 \pm 1.73$ & $78.07 \pm 2.00$ & $76.55 \pm 3.32$ & $78.96 \pm 3.63$ \\
Pubmed & $86.39 \pm 0.21$ & $86.98 \pm 0.28$ & $86.37 \pm 0.35$ & $86.31 \pm 0.34$ & $87.13 \pm 0.11$ \\
Gamers & $51.34 \pm 1.29$ & $51.75 \pm 1.38$ & $51.36 \pm 1.32$ & $51.30 \pm 0.84$ & $51.94 \pm 1.37$ \\
Pokec & $57.19 \pm 0.66$ & $58.34 \pm 0.67$ & $57.27 \pm 0.90$ & $57.89 \pm 0.97$ & $58.37 \pm 1.08$ \\
Ratings & $26.53 \pm 1.35$ & $25.70 \pm 0.05$ & $26.49 \pm 1.22$ & $27.18 \pm 1.14$ & $26.84 \pm 0.78$ \\
Trivago & $47.75 \pm 0.52$ & $44.81 \pm 0.77$ & $55.01 \pm 0.30$ & $44.90 \pm 0.54$ & $51.10 \pm 6.84$ \\
Walmart & $67.97 \pm 0.47$ & $74.56 \pm 0.21$ & $67.63 \pm 0.42$ & $70.41 \pm 0.24$ & $76.07 \pm 0.24$ \\
Yelp & $26.84 \pm 0.35$ & $29.57 \pm 1.22$ & $26.81 \pm 0.30$ & $27.00 \pm 0.14$ & $28.88 \pm 0.86$ \\
\bottomrule
\end{tabular}
\end{table}

\begin{table}[p]
\centering
\small
\caption{Node classification. Changes relative to the official CEGCN, mean $\pm$ sample SD.}
\label{tab:stronger-node_cls-deltas}
\setlength{\tabcolsep}{4pt}
\begin{tabular}{@{}lrrrr@{}}
\toprule
Dataset & +O & +D & +L & +O+D+L \\
\midrule
ModelNet40 & $+5.10 \pm 0.34$ & $-0.12 \pm 0.11$ & $-0.14 \pm 0.21$ & $+4.70 \pm 0.27$ \\
NTU2012 & $+4.11 \pm 0.61$ & $-0.13 \pm 0.11$ & $-0.66 \pm 0.50$ & $+4.11 \pm 0.98$ \\
Actor & $+14.40 \pm 0.40$ & $-0.10 \pm 0.05$ & $+0.37 \pm 0.23$ & $+14.58 \pm 0.37$ \\
Cora-CA & $+5.02 \pm 2.60$ & $+0.10 \pm 0.70$ & $+0.20 \pm 0.37$ & $+5.07 \pm 2.74$ \\
DBLP-CA & $+1.37 \pm 0.22$ & $-0.01 \pm 0.15$ & $+0.41 \pm 0.03$ & $+1.90 \pm 0.21$ \\
Cora & $+0.79 \pm 1.34$ & $+0.34 \pm 0.61$ & $-1.18 \pm 1.29$ & $+1.23 \pm 1.51$ \\
Pubmed & $+0.59 \pm 0.13$ & $-0.02 \pm 0.14$ & $-0.08 \pm 0.13$ & $+0.74 \pm 0.26$ \\
Gamers & $+0.41 \pm 0.10$ & $+0.02 \pm 0.04$ & $-0.03 \pm 0.65$ & $+0.60 \pm 0.10$ \\
Pokec & $+1.15 \pm 0.12$ & $+0.08 \pm 0.51$ & $+0.70 \pm 0.33$ & $+1.18 \pm 0.71$ \\
Ratings & $-0.83 \pm 1.33$ & $-0.04 \pm 0.13$ & $+0.65 \pm 2.30$ & $+0.31 \pm 2.04$ \\
Trivago & $-2.93 \pm 0.57$ & $+7.26 \pm 0.72$ & $-2.84 \pm 0.98$ & $+3.35 \pm 6.76$ \\
Walmart & $+6.59 \pm 0.34$ & $-0.35 \pm 0.06$ & $+2.43 \pm 0.36$ & $+8.09 \pm 0.28$ \\
Yelp & $+2.73 \pm 1.57$ & $-0.03 \pm 0.14$ & $+0.16 \pm 0.33$ & $+2.04 \pm 1.11$ \\
\bottomrule
\end{tabular}
\end{table}

\begin{table}[t]
\centering
\begingroup\small
\setlength{\tabcolsep}{3pt}
\caption{{Node classification: successive changes to the propagation operator (accuracy). Original-view scores (percent) and paired stepwise changes (percentage points) are mean $\pm$ sample SD. O$_1$ adds pair multiplicities; O$_2$ changes structural self weights and normalization after pair multiplicities have been added by O$_1$. D and L are disabled.}}
\label{tab:node-cls-o1o2}
\begin{tabular}{@{}lrrrrr@{}}
\toprule
Dataset & CEGCN & +O$_1$ & +O$_1$+O$_2$ & $\delta_1$ & $\delta_2$ \\
\midrule
ModelNet40 & $92.75 \pm 0.56$ & $92.93 \pm 0.86$ & $97.85 \pm 0.42$ & $0.18 \pm 0.41$ & $4.92 \pm 0.47$ \\
NTU2012 & $84.76 \pm 0.64$ & $85.69 \pm 0.72$ & $88.87 \pm 0.72$ & $0.93 \pm 0.61$ & $3.18 \pm 0.00$ \\
Actor & $67.86 \pm 0.42$ & $67.59 \pm 0.51$ & $82.26 \pm 0.53$ & $-0.27 \pm 0.22$ & $14.67 \pm 0.21$ \\
Cora-CA & $79.57 \pm 2.79$ & $80.01 \pm 2.37$ & $84.59 \pm 0.45$ & $0.44 \pm 0.53$ & $4.58 \pm 2.13$ \\
DBLP-CA & $89.83 \pm 0.15$ & $89.91 \pm 0.15$ & $91.19 \pm 0.35$ & $0.09 \pm 0.05$ & $1.28 \pm 0.25$ \\
Cora & $77.73 \pm 2.23$ & $78.17 \pm 2.45$ & $78.52 \pm 1.73$ & $0.44 \pm 0.29$ & $0.34 \pm 1.63$ \\
Pubmed & $86.39 \pm 0.21$ & $86.48 \pm 0.26$ & $86.98 \pm 0.28$ & $0.09 \pm 0.05$ & $0.51 \pm 0.11$ \\
Gamers & $51.34 \pm 1.29$ & $51.30 \pm 1.34$ & $51.75 \pm 1.38$ & $-0.03 \pm 0.09$ & $0.44 \pm 0.13$ \\
Pokec & $57.19 \pm 0.66$ & $57.47 \pm 1.51$ & $58.34 \pm 0.67$ & $0.28 \pm 0.99$ & $0.87 \pm 1.08$ \\
Ratings & $26.53 \pm 1.35$ & $26.52 \pm 1.30$ & $25.70 \pm 0.05$ & $-0.01 \pm 0.05$ & $-0.82 \pm 1.29$ \\
Trivago & $47.75 \pm 0.52$ & $50.13 \pm 0.25$ & $44.81 \pm 0.77$ & $2.39 \pm 0.30$ & $-5.32 \pm 0.56$ \\
Walmart & $67.97 \pm 0.47$ & $68.15 \pm 0.42$ & $74.56 \pm 0.21$ & $0.18 \pm 0.06$ & $6.41 \pm 0.29$ \\
Yelp & $26.84 \pm 0.35$ & $27.04 \pm 0.17$ & $29.57 \pm 1.22$ & $0.20 \pm 0.29$ & $2.52 \pm 1.32$ \\
\bottomrule
\end{tabular}
\endgroup
\end{table}

\begin{table}[p]
\centering
\small
\caption{Hypergraph classification. Mean $\pm$ sample SD. The metric is accuracy.}
\label{tab:stronger-hg_cls-scores}
\setlength{\tabcolsep}{4pt}
\begin{tabular}{@{}lrrrrr@{}}
\toprule
Dataset & CEGCN & +O & +D & +L & +O+D+L \\
\midrule
IMDB-Dir-Form & $63.30 \pm 1.92$ & $61.70 \pm 0.53$ & $63.83 \pm 3.72$ & $61.52 \pm 1.11$ & $60.28 \pm 2.01$ \\
IMDB-Dir-Genre & $77.55 \pm 1.11$ & $76.86 \pm 1.45$ & $77.35 \pm 1.35$ & $78.24 \pm 0.51$ & $76.47 \pm 2.35$ \\
RHG-10 & $95.00 \pm 1.80$ & $93.83 \pm 1.89$ & $95.00 \pm 1.32$ & $95.83 \pm 0.76$ & $94.67 \pm 0.29$ \\
RHG-3 & $98.67 \pm 0.67$ & $98.00 \pm 1.76$ & $98.00 \pm 0.67$ & $98.44 \pm 0.77$ & $99.78 \pm 0.38$ \\
Steam-Player & $54.85 \pm 2.43$ & $55.18 \pm 1.01$ & $54.53 \pm 2.67$ & $53.24 \pm 4.04$ & $53.72 \pm 2.67$ \\
Twitter-Friend & $58.02 \pm 1.32$ & $56.23 \pm 3.09$ & $51.40 \pm 1.59$ & $55.73 \pm 3.05$ & $56.74 \pm 1.17$ \\
\bottomrule
\end{tabular}
\end{table}

\begin{table}[p]
\centering
\small
\caption{Hypergraph classification. Changes relative to the official CEGCN, mean $\pm$ sample SD.}
\label{tab:stronger-hg_cls-deltas}
\setlength{\tabcolsep}{4pt}
\begin{tabular}{@{}lrrrr@{}}
\toprule
Dataset & +O & +D & +L & +O+D+L \\
\midrule
IMDB-Dir-Form & $-1.60 \pm 1.60$ & $+0.53 \pm 2.76$ & $-1.77 \pm 1.63$ & $-3.01 \pm 3.74$ \\
IMDB-Dir-Genre & $-0.69 \pm 0.34$ & $-0.20 \pm 1.03$ & $+0.69 \pm 0.90$ & $-1.08 \pm 1.62$ \\
RHG-10 & $-1.17 \pm 1.15$ & $+0.00 \pm 1.50$ & $+0.83 \pm 2.25$ & $-0.33 \pm 1.61$ \\
RHG-3 & $-0.67 \pm 2.00$ & $-0.67 \pm 0.67$ & $-0.22 \pm 0.38$ & $+1.11 \pm 0.38$ \\
Steam-Player & $+0.32 \pm 1.48$ & $-0.32 \pm 0.28$ & $-1.62 \pm 2.84$ & $-1.13 \pm 0.28$ \\
Twitter-Friend & $-1.78 \pm 3.92$ & $-6.62 \pm 2.68$ & $-2.29 \pm 4.25$ & $-1.27 \pm 2.45$ \\
\bottomrule
\end{tabular}
\end{table}

\begin{table}[t]
\centering
\begingroup\small
\setlength{\tabcolsep}{3pt}
\caption{{Hypergraph classification: successive changes to the propagation operator (accuracy). Original-view scores (percent) and paired stepwise changes (percentage points) are mean $\pm$ sample SD. O$_1$ adds pair multiplicities; O$_2$ changes structural self weights and normalization after pair multiplicities have been added by O$_1$. D and L are disabled.}}
\label{tab:hg-cls-o1o2-primary}
\begin{tabular}{@{}lrrrrr@{}}
\toprule
Dataset & CEGCN & +O$_1$ & +O$_1$+O$_2$ & $\delta_1$ & $\delta_2$ \\
\midrule
IMDB-Dir-Form & $63.30 \pm 1.92$ & $60.28 \pm 2.68$ & $61.70 \pm 0.53$ & $-3.01 \pm 3.02$ & $1.42 \pm 3.02$ \\
IMDB-Dir-Genre & $77.55 \pm 1.11$ & $77.94 \pm 0.51$ & $76.86 \pm 1.45$ & $0.39 \pm 1.03$ & $-1.08 \pm 1.33$ \\
RHG-10 & $95.00 \pm 1.80$ & $94.67 \pm 0.58$ & $93.83 \pm 1.89$ & $-0.33 \pm 2.02$ & $-0.83 \pm 1.76$ \\
RHG-3 & $98.67 \pm 0.67$ & $97.78 \pm 1.39$ & $98.00 \pm 1.76$ & $-0.89 \pm 0.77$ & $0.22 \pm 2.69$ \\
Steam-Player & $54.85 \pm 2.43$ & $54.05 \pm 2.93$ & $55.18 \pm 1.01$ & $-0.81 \pm 0.56$ & $1.13 \pm 2.02$ \\
Twitter-Friend & $58.02 \pm 1.32$ & $55.22 \pm 5.93$ & $56.23 \pm 3.09$ & $-2.80 \pm 7.21$ & $1.02 \pm 5.93$ \\
\bottomrule
\end{tabular}
\endgroup
\end{table}

\begin{table}[p]
\centering
\small
\caption{Hyperedge prediction using the identity policy for cold nodes. Mean $\pm$ sample SD. The metric is AP.}
\label{tab:stronger-edge_pred_identity-scores}
\setlength{\tabcolsep}{4pt}
\begin{tabular}{@{}lrrrrr@{}}
\toprule
Dataset & CEGCN & +O & +D & +L & +O+D+L \\
\midrule
Actor & $72.11 \pm 0.57$ & $74.91 \pm 1.75$ & $71.39 \pm 1.76$ & $69.39 \pm 0.90$ & $75.39 \pm 0.79$ \\
Cora-CA & $59.57 \pm 1.67$ & $58.87 \pm 1.25$ & $59.95 \pm 2.46$ & $57.75 \pm 1.30$ & $59.55 \pm 2.00$ \\
DBLP-CA & $63.80 \pm 1.34$ & $73.03 \pm 2.23$ & $63.34 \pm 2.39$ & $63.16 \pm 0.50$ & $71.59 \pm 1.00$ \\
Cora & $59.27 \pm 1.18$ & $57.99 \pm 1.82$ & $59.02 \pm 1.49$ & $56.45 \pm 1.35$ & $57.99 \pm 1.18$ \\
Pubmed & $71.12 \pm 0.70$ & $71.49 \pm 0.72$ & $71.89 \pm 0.27$ & $70.42 \pm 0.51$ & $71.75 \pm 1.68$ \\
Pokec & $59.51 \pm 3.01$ & $55.17 \pm 1.40$ & $60.11 \pm 3.73$ & $59.56 \pm 2.45$ & $59.74 \pm 0.74$ \\
\bottomrule
\end{tabular}
\end{table}

\begin{table}[p]
\centering
\small
\caption{Hyperedge prediction using the identity policy for cold nodes: changes relative to the official CEGCN using the same policy for cold nodes, mean $\pm$ sample SD.}
\label{tab:stronger-edge_pred_identity-deltas}
\setlength{\tabcolsep}{4pt}
\begin{tabular}{@{}lrrrr@{}}
\toprule
Dataset & +O & +D & +L & +O+D+L \\
\midrule
Actor & $+2.80 \pm 2.14$ & $-0.72 \pm 2.01$ & $-2.72 \pm 0.73$ & $+3.28 \pm 1.29$ \\
Cora-CA & $-0.70 \pm 1.50$ & $+0.38 \pm 1.70$ & $-1.82 \pm 0.37$ & $-0.02 \pm 3.13$ \\
DBLP-CA & $+9.23 \pm 1.43$ & $-0.46 \pm 1.29$ & $-0.64 \pm 1.21$ & $+7.79 \pm 0.69$ \\
Cora & $-1.28 \pm 1.49$ & $-0.25 \pm 0.83$ & $-2.82 \pm 0.95$ & $-1.28 \pm 1.05$ \\
Pubmed & $+0.37 \pm 0.31$ & $+0.77 \pm 0.85$ & $-0.70 \pm 1.21$ & $+0.63 \pm 1.02$ \\
Pokec & $-4.34 \pm 1.76$ & $+0.60 \pm 1.83$ & $+0.05 \pm 3.49$ & $+0.23 \pm 3.52$ \\
\bottomrule
\end{tabular}
\end{table}

\begin{table}[t]
\centering
\begingroup\small
\setlength{\tabcolsep}{3pt}
\caption{{Hyperedge prediction: successive changes to the propagation operator (AP; identity policy for cold nodes). Original-view scores (percent) and paired stepwise changes (percentage points) are mean $\pm$ sample SD. O$_1$ adds pair multiplicities; O$_2$ changes structural self weights and normalization after pair multiplicities have been added by O$_1$. D and L are disabled.}}
\label{tab:operator-edge-pred-identity-primary}
\begin{tabular}{@{}lrrrrr@{}}
\toprule
Dataset & CEGCN & +O$_1$ & +O$_1$+O$_2$ & $\delta_1$ & $\delta_2$ \\
\midrule
Actor & $72.11 \pm 0.57$ & $71.08 \pm 2.16$ & $74.91 \pm 1.75$ & $-1.03 \pm 1.63$ & $3.83 \pm 3.73$ \\
Cora-CA & $59.57 \pm 1.67$ & $59.17 \pm 1.32$ & $58.87 \pm 1.25$ & $-0.40 \pm 0.87$ & $-0.30 \pm 1.83$ \\
DBLP-CA & $63.80 \pm 1.34$ & $64.18 \pm 2.10$ & $73.03 \pm 2.23$ & $0.38 \pm 0.97$ & $8.85 \pm 0.66$ \\
Cora & $59.27 \pm 1.18$ & $58.85 \pm 1.01$ & $57.99 \pm 1.82$ & $-0.42 \pm 0.19$ & $-0.85 \pm 1.39$ \\
Pubmed & $71.12 \pm 0.70$ & $71.85 \pm 0.60$ & $71.49 \pm 0.72$ & $0.73 \pm 0.19$ & $-0.36 \pm 0.17$ \\
Pokec & $59.51 \pm 3.01$ & $59.29 \pm 1.15$ & $55.17 \pm 1.40$ & $-0.22 \pm 3.37$ & $-4.12 \pm 2.20$ \\
\bottomrule
\end{tabular}
\endgroup
\end{table}

\begin{table}[p]
\centering
\small
\caption{Hyperedge prediction using the zero-message policy for cold nodes. Mean $\pm$ sample SD. The metric is AP.}
\label{tab:stronger-edge_pred_zero-scores}
\setlength{\tabcolsep}{4pt}
\begin{tabular}{@{}lrrrrr@{}}
\toprule
Dataset & CEGCN & +O & +D & +L & +O+D+L \\
\midrule
Actor & $64.97 \pm 1.39$ & $67.46 \pm 1.85$ & $64.09 \pm 0.36$ & $65.14 \pm 1.33$ & $67.87 \pm 2.50$ \\
Cora-CA & $60.85 \pm 7.06$ & $56.01 \pm 2.27$ & $60.89 \pm 5.55$ & $60.77 \pm 3.60$ & $60.30 \pm 0.51$ \\
DBLP-CA & $65.19 \pm 3.46$ & $60.91 \pm 0.34$ & $65.73 \pm 3.48$ & $69.85 \pm 1.42$ & $71.49 \pm 0.25$ \\
Cora & $57.52 \pm 0.79$ & $56.24 \pm 0.80$ & $57.13 \pm 0.97$ & $57.09 \pm 1.09$ & $57.52 \pm 0.75$ \\
Pubmed & $69.38 \pm 1.14$ & $68.99 \pm 1.08$ & $69.84 \pm 0.57$ & $68.74 \pm 1.23$ & $69.21 \pm 0.58$ \\
Pokec & $78.37 \pm 2.05$ & $75.81 \pm 2.99$ & $77.93 \pm 1.01$ & $79.37 \pm 0.81$ & $78.34 \pm 1.71$ \\
\bottomrule
\end{tabular}
\end{table}

\begin{table}[p]
\centering
\small
\caption{Hyperedge prediction using the zero-message policy for cold nodes. Changes relative to the official CEGCN using the same policy for cold nodes, mean $\pm$ sample SD.}
\label{tab:stronger-edge_pred_zero-deltas}
\setlength{\tabcolsep}{4pt}
\begin{tabular}{@{}lrrrr@{}}
\toprule
Dataset & +O & +D & +L & +O+D+L \\
\midrule
Actor & $+2.49 \pm 3.21$ & $-0.87 \pm 1.65$ & $+0.17 \pm 2.69$ & $+2.90 \pm 1.24$ \\
Cora-CA & $-4.84 \pm 4.94$ & $+0.04 \pm 1.79$ & $-0.08 \pm 7.11$ & $-0.55 \pm 6.55$ \\
DBLP-CA & $-4.28 \pm 3.61$ & $+0.54 \pm 1.15$ & $+4.65 \pm 2.16$ & $+6.30 \pm 3.21$ \\
Cora & $-1.28 \pm 0.31$ & $-0.39 \pm 0.60$ & $-0.43 \pm 1.18$ & $-0.00 \pm 1.24$ \\
Pubmed & $-0.39 \pm 0.09$ & $+0.46 \pm 1.06$ & $-0.64 \pm 0.85$ & $-0.17 \pm 1.12$ \\
Pokec & $-2.56 \pm 2.85$ & $-0.44 \pm 1.64$ & $+1.00 \pm 1.30$ & $-0.03 \pm 0.34$ \\
\bottomrule
\end{tabular}
\end{table}

\begin{table}[t]
\centering
\begingroup\small
\setlength{\tabcolsep}{3pt}
\caption{{Hyperedge prediction: successive changes to the propagation operator (AP; zero-message policy for cold nodes). Original-view scores (percent) and paired stepwise changes (percentage points) are mean $\pm$ sample SD. O$_1$ adds pair multiplicities; O$_2$ changes structural self weights and normalization after pair multiplicities have been added by O$_1$. D and L are disabled.}}
\label{tab:operator-edge-pred-zero-primary}
\begin{tabular}{@{}lrrrrr@{}}
\toprule
Dataset & CEGCN & +O$_1$ & +O$_1$+O$_2$ & $\delta_1$ & $\delta_2$ \\
\midrule
Actor & $64.97 \pm 1.39$ & $63.42 \pm 0.65$ & $67.46 \pm 1.85$ & $-1.55 \pm 1.02$ & $4.04 \pm 2.45$ \\
Cora-CA & $60.85 \pm 7.06$ & $59.66 \pm 5.67$ & $56.01 \pm 2.27$ & $-1.18 \pm 1.99$ & $-3.65 \pm 3.89$ \\
DBLP-CA & $65.19 \pm 3.46$ & $61.70 \pm 2.29$ & $60.91 \pm 0.34$ & $-3.50 \pm 2.92$ & $-0.79 \pm 2.63$ \\
Cora & $57.52 \pm 0.79$ & $57.17 \pm 1.01$ & $56.24 \pm 0.80$ & $-0.35 \pm 0.83$ & $-0.92 \pm 1.12$ \\
Pubmed & $69.38 \pm 1.14$ & $69.72 \pm 1.16$ & $68.99 \pm 1.08$ & $0.34 \pm 1.03$ & $-0.73 \pm 0.95$ \\
Pokec & $78.37 \pm 2.05$ & $75.35 \pm 0.62$ & $75.81 \pm 2.99$ & $-3.01 \pm 1.53$ & $0.46 \pm 2.52$ \\
\bottomrule
\end{tabular}
\endgroup
\end{table}

\begin{table}[p]
\centering
\small
\caption{Hyperedge prediction: AP under the zero-message versus identity policy for cold nodes, mean $\pm$ sample SD.}
\label{tab:stronger-edge-policy}
\setlength{\tabcolsep}{4pt}
\begin{tabular}{@{}lrrrrr@{}}
\toprule
Dataset & CEGCN & +O & +D & +L & +O+D+L \\
\midrule
Actor & $-7.14 \pm 1.37$ & $-7.45 \pm 3.03$ & $-7.30 \pm 1.52$ & $-4.25 \pm 1.25$ & $-7.51 \pm 2.55$ \\
Cora-CA & $+1.28 \pm 6.03$ & $-2.87 \pm 3.14$ & $+0.94 \pm 5.72$ & $+3.02 \pm 4.46$ & $+0.75 \pm 2.48$ \\
DBLP-CA & $+1.39 \pm 2.38$ & $-12.12 \pm 2.01$ & $+2.39 \pm 1.29$ & $+6.69 \pm 1.19$ & $-0.10 \pm 0.90$ \\
Cora & $-1.75 \pm 0.89$ & $-1.75 \pm 2.38$ & $-1.89 \pm 0.70$ & $+0.64 \pm 0.43$ & $-0.47 \pm 0.44$ \\
Pubmed & $-1.74 \pm 0.58$ & $-2.51 \pm 0.86$ & $-2.05 \pm 0.35$ & $-1.68 \pm 1.43$ & $-2.54 \pm 1.95$ \\
Pokec & $+18.86 \pm 3.29$ & $+20.64 \pm 3.95$ & $+17.82 \pm 2.72$ & $+19.81 \pm 3.12$ & $+18.60 \pm 2.38$ \\
\bottomrule
\end{tabular}
\end{table}

\paragraph{Stronger-baseline comparisons using secondary metrics.}
\cref{tab:secondary-transfer,tab:secondary-hg_cls_standard-scores,tab:secondary-hg_cls_standard-deltas,tab:operator-hg-cls-standard-secondary,tab:secondary-edge_pred_identity-scores,tab:secondary-edge_pred_identity-deltas,tab:operator-edge-pred-identity-secondary,tab:secondary-edge_pred_zero-scores,tab:secondary-edge_pred_zero-deltas,tab:operator-edge-pred-zero-secondary} report secondary-metric results.

\begin{table}[t!]
\centering
\small
\caption{Secondary-metric gains on all benchmarks. Edge-Pred denotes hyperedge prediction; HG-CLS denotes hypergraph classification. Mean paired gain over standard CEGCN (points), weighting benchmarks equally and using the same policy for cold nodes in hyperedge prediction; followed by the number of benchmarks with a positive mean gain.}
\label{tab:secondary-transfer}
\setlength{\tabcolsep}{4pt}
\begin{tabular}{@{}lrrrr@{}}
\toprule
Task / policy & +O & +D & +L & +O+D+L \\
\midrule
Edge-Pred: identity (AUROC) & $+1.69$ (3/6) & $-0.04$ (4/6) & $-1.58$ (1/6) & $+2.31$ (4/6) \\
Edge-Pred: zero (AUROC) & $-2.73$ (0/6) & $-0.55$ (2/6) & $+0.34$ (1/6) & $+1.07$ (4/6) \\
HG-CLS (Macro-F1) & $-2.62$ (0/6) & $-1.29$ (2/6) & $-0.58$ (3/6) & $-0.78$ (1/6) \\
\bottomrule
\end{tabular}
\end{table}

\begin{table}[p]
\centering
\small
\caption{Hypergraph classification (Macro-F1): original-view scores (percent), mean $\pm$ sample SD.}
\label{tab:secondary-hg_cls_standard-scores}
\setlength{\tabcolsep}{4pt}
\begin{tabular}{@{}lrrrrr@{}}
\toprule
Dataset & CEGCN & +O & +D & +L & +O+D+L \\
\midrule
IMDB-Dir-Form & $57.19 \pm 2.34$ & $56.45 \pm 3.18$ & $58.09 \pm 5.91$ & $57.66 \pm 1.48$ & $56.15 \pm 4.49$ \\
IMDB-Dir-Genre & $72.65 \pm 2.55$ & $70.60 \pm 4.08$ & $72.58 \pm 2.79$ & $73.07 \pm 2.08$ & $71.28 \pm 2.93$ \\
RHG-10 & $94.81 \pm 1.88$ & $93.85 \pm 2.00$ & $94.83 \pm 1.32$ & $95.68 \pm 0.64$ & $94.55 \pm 0.39$ \\
RHG-3 & $98.72 \pm 0.64$ & $98.00 \pm 1.78$ & $98.10 \pm 0.54$ & $98.52 \pm 0.70$ & $99.78 \pm 0.39$ \\
Steam-Player & $54.03 \pm 2.57$ & $47.68 \pm 1.94$ & $52.81 \pm 3.64$ & $52.24 \pm 3.03$ & $51.43 \pm 1.52$ \\
Twitter-Friend & $54.44 \pm 5.46$ & $49.56 \pm 2.25$ & $47.72 \pm 4.07$ & $51.22 \pm 1.92$ & $53.98 \pm 2.62$ \\
\bottomrule
\end{tabular}
\end{table}

\begin{table}[p]
\centering
\small
\caption{Hypergraph classification (Macro-F1): paired gains over standard CEGCN (points), mean $\pm$ sample SD.}
\label{tab:secondary-hg_cls_standard-deltas}
\setlength{\tabcolsep}{4pt}
\begin{tabular}{@{}lrrrr@{}}
\toprule
Dataset & +O & +D & +L & +O+D+L \\
\midrule
IMDB-Dir-Form & $-0.73 \pm 0.84$ & $+0.90 \pm 3.66$ & $+0.48 \pm 0.89$ & $-1.03 \pm 2.94$ \\
IMDB-Dir-Genre & $-2.05 \pm 1.55$ & $-0.07 \pm 1.54$ & $+0.41 \pm 1.19$ & $-1.37 \pm 1.87$ \\
RHG-10 & $-0.96 \pm 1.18$ & $+0.02 \pm 1.47$ & $+0.87 \pm 2.28$ & $-0.26 \pm 1.61$ \\
RHG-3 & $-0.72 \pm 1.97$ & $-0.62 \pm 0.61$ & $-0.20 \pm 0.35$ & $+1.05 \pm 0.33$ \\
Steam-Player & $-6.36 \pm 2.17$ & $-1.23 \pm 1.75$ & $-1.79 \pm 1.50$ & $-2.61 \pm 1.84$ \\
Twitter-Friend & $-4.89 \pm 3.30$ & $-6.72 \pm 3.48$ & $-3.23 \pm 4.49$ & $-0.47 \pm 7.38$ \\
\bottomrule
\end{tabular}
\end{table}

\begin{table}[t]
\centering
\begingroup\small
\setlength{\tabcolsep}{3pt}
\caption{{Hypergraph classification: successive changes to the propagation operator (Macro-F1). Original-view scores (percent) and paired stepwise changes (percentage points) are mean $\pm$ sample SD. 
O$_1$ adds pair multiplicities; O$_2$ changes structural self weights and normalization after pair multiplicities have been added by O$_1$. D and L are disabled.}}
\label{tab:operator-hg-cls-standard-secondary}
\begin{tabular}{@{}lrrrrr@{}}
\toprule
Dataset & CEGCN & +O$_1$ & +O$_1$+O$_2$ & $\delta_1$ & $\delta_2$ \\
\midrule
IMDB-Dir-Form & $57.19 \pm 2.34$ & $54.61 \pm 1.08$ & $56.45 \pm 3.18$ & $-2.58 \pm 2.06$ & $1.85 \pm 2.83$ \\
IMDB-Dir-Genre & $72.65 \pm 2.55$ & $73.20 \pm 1.38$ & $70.60 \pm 4.08$ & $0.54 \pm 1.59$ & $-2.60 \pm 2.94$ \\
RHG-10 & $94.81 \pm 1.88$ & $94.57 \pm 0.63$ & $93.85 \pm 2.00$ & $-0.24 \pm 2.14$ & $-0.73 \pm 1.88$ \\
RHG-3 & $98.72 \pm 0.64$ & $97.82 \pm 1.35$ & $98.00 \pm 1.78$ & $-0.90 \pm 0.78$ & $0.18 \pm 2.68$ \\
Steam-Player & $54.03 \pm 2.57$ & $52.60 \pm 3.32$ & $47.68 \pm 1.94$ & $-1.44 \pm 1.10$ & $-4.92 \pm 2.20$ \\
Twitter-Friend & $54.44 \pm 5.46$ & $51.61 \pm 3.51$ & $49.56 \pm 2.25$ & $-2.83 \pm 7.24$ & $-2.06 \pm 4.26$ \\
\bottomrule
\end{tabular}
\endgroup
\end{table}

\begin{table}[p]
\centering
\small
\caption{Hyperedge prediction using the identity policy for cold nodes (AUROC): original-view scores (percent), mean $\pm$ sample SD over the same three seeds as the primary-metric table.}
\label{tab:secondary-edge_pred_identity-scores}
\setlength{\tabcolsep}{4pt}
\begin{tabular}{@{}lrrrrr@{}}
\toprule
Dataset & CEGCN & +O & +D & +L & +O+D+L \\
\midrule
Actor & $71.57 \pm 0.93$ & $74.54 \pm 1.77$ & $71.58 \pm 1.89$ & $68.80 \pm 1.46$ & $75.14 \pm 0.18$ \\
Cora-CA & $60.08 \pm 1.87$ & $59.24 \pm 2.11$ & $60.21 \pm 1.71$ & $58.47 \pm 1.58$ & $60.92 \pm 2.39$ \\
DBLP-CA & $63.58 \pm 2.01$ & $73.87 \pm 2.05$ & $62.81 \pm 2.60$ & $63.49 \pm 0.92$ & $73.54 \pm 0.88$ \\
Cora & $61.82 \pm 1.04$ & $60.72 \pm 3.32$ & $61.26 \pm 0.81$ & $57.70 \pm 2.19$ & $60.71 \pm 1.90$ \\
Pubmed & $74.38 \pm 0.33$ & $74.57 \pm 0.69$ & $74.60 \pm 0.29$ & $73.47 \pm 0.18$ & $75.30 \pm 1.24$ \\
Pokec & $55.92 \pm 1.60$ & $54.56 \pm 0.78$ & $56.62 \pm 0.93$ & $55.93 \pm 2.17$ & $55.60 \pm 3.13$ \\
\bottomrule
\end{tabular}
\end{table}

\begin{table}[p]
\centering
\small
\caption{Hyperedge prediction using the identity policy for cold nodes (AUROC): paired gains over CEGCN using the same policy for cold nodes (points), mean $\pm$ sample SD over the same three seeds as the primary-metric table.}
\label{tab:secondary-edge_pred_identity-deltas}
\setlength{\tabcolsep}{4pt}
\begin{tabular}{@{}lrrrr@{}}
\toprule
Dataset & +O & +D & +L & +O+D+L \\
\midrule
Actor & $+2.97 \pm 2.42$ & $+0.01 \pm 1.62$ & $-2.77 \pm 0.97$ & $+3.57 \pm 1.07$ \\
Cora-CA & $-0.84 \pm 2.97$ & $+0.13 \pm 0.54$ & $-1.61 \pm 0.63$ & $+0.84 \pm 3.70$ \\
DBLP-CA & $+10.29 \pm 1.03$ & $-0.77 \pm 0.96$ & $-0.09 \pm 1.89$ & $+9.96 \pm 1.16$ \\
Cora & $-1.09 \pm 2.68$ & $-0.56 \pm 1.01$ & $-4.11 \pm 1.43$ & $-1.10 \pm 1.28$ \\
Pubmed & $+0.19 \pm 0.37$ & $+0.23 \pm 0.17$ & $-0.90 \pm 0.51$ & $+0.92 \pm 0.96$ \\
Pokec & $-1.36 \pm 1.12$ & $+0.70 \pm 2.44$ & $+0.01 \pm 1.22$ & $-0.32 \pm 1.96$ \\
\bottomrule
\end{tabular}
\end{table}

\begin{table}[t]
\centering
\begingroup\small
\setlength{\tabcolsep}{3pt}
\caption{{Hyperedge prediction: successive changes to the propagation operator (AUROC; identity policy for cold nodes). Original-view scores (percent) and paired stepwise changes (percentage points) are mean $\pm$ sample SD. O$_1$ adds pair multiplicities; O$_2$ changes structural self weights and normalization after pair multiplicities have been added by O$_1$. D and L are disabled.}}
\label{tab:operator-edge-pred-identity-secondary}
\begin{tabular}{@{}lrrrrr@{}}
\toprule
Dataset & CEGCN & +O$_1$ & +O$_1$+O$_2$ & $\delta_1$ & $\delta_2$ \\
\midrule
Actor & $71.57 \pm 0.93$ & $71.35 \pm 1.83$ & $74.54 \pm 1.77$ & $-0.22 \pm 0.98$ & $3.19 \pm 2.91$ \\
Cora-CA & $60.08 \pm 1.87$ & $59.64 \pm 1.53$ & $59.24 \pm 2.11$ & $-0.44 \pm 0.69$ & $-0.40 \pm 3.11$ \\
DBLP-CA & $63.58 \pm 2.01$ & $63.95 \pm 2.18$ & $73.87 \pm 2.05$ & $0.37 \pm 0.40$ & $9.92 \pm 0.73$ \\
Cora & $61.82 \pm 1.04$ & $61.08 \pm 1.14$ & $60.72 \pm 3.32$ & $-0.73 \pm 1.31$ & $-0.36 \pm 2.40$ \\
Pubmed & $74.38 \pm 0.33$ & $74.97 \pm 0.17$ & $74.57 \pm 0.69$ & $0.59 \pm 0.17$ & $-0.40 \pm 0.52$ \\
Pokec & $55.92 \pm 1.60$ & $56.53 \pm 0.80$ & $54.56 \pm 0.78$ & $0.61 \pm 2.40$ & $-1.97 \pm 1.47$ \\
\bottomrule
\end{tabular}
\endgroup
\end{table}

\begin{table}[p]
\centering
\small
\caption{Hyperedge prediction using the zero-message policy for cold nodes (AUROC): original-view scores (percent), mean $\pm$ sample SD.}
\label{tab:secondary-edge_pred_zero-scores}
\setlength{\tabcolsep}{4pt}
\begin{tabular}{@{}lrrrrr@{}}
\toprule
Dataset & CEGCN & +O & +D & +L & +O+D+L \\
\midrule
Actor & $65.44 \pm 2.43$ & $65.12 \pm 0.75$ & $63.48 \pm 0.66$ & $65.09 \pm 2.54$ & $67.08 \pm 4.28$ \\
Cora-CA & $61.17 \pm 6.87$ & $53.74 \pm 1.44$ & $62.03 \pm 6.82$ & $60.59 \pm 2.12$ & $62.88 \pm 1.11$ \\
DBLP-CA & $65.13 \pm 3.35$ & $60.45 \pm 1.38$ & $64.86 \pm 3.43$ & $69.36 \pm 1.03$ & $69.40 \pm 0.37$ \\
Cora & $59.89 \pm 0.89$ & $58.58 \pm 0.57$ & $59.40 \pm 1.22$ & $58.84 \pm 1.36$ & $58.87 \pm 1.32$ \\
Pubmed & $72.26 \pm 1.02$ & $72.17 \pm 0.45$ & $72.47 \pm 0.76$ & $72.04 \pm 1.05$ & $73.16 \pm 0.32$ \\
Pokec & $75.22 \pm 3.07$ & $72.69 \pm 2.70$ & $73.56 \pm 0.87$ & $75.21 \pm 1.26$ & $74.13 \pm 1.34$ \\
\bottomrule
\end{tabular}
\end{table}

\begin{table}[p]
\centering
\small
\caption{Hyperedge prediction using the zero-message policy for cold nodes (AUROC): paired gains over CEGCN using the same policy for cold nodes (points), mean $\pm$ sample SD.}
\label{tab:secondary-edge_pred_zero-deltas}
\setlength{\tabcolsep}{4pt}
\begin{tabular}{@{}lrrrr@{}}
\toprule
Dataset & +O & +D & +L & +O+D+L \\
\midrule
Actor & $-0.32 \pm 2.07$ & $-1.97 \pm 1.94$ & $-0.35 \pm 4.96$ & $+1.64 \pm 2.91$ \\
Cora-CA & $-7.43 \pm 6.47$ & $+0.86 \pm 0.28$ & $-0.57 \pm 6.57$ & $+1.71 \pm 6.66$ \\
DBLP-CA & $-4.68 \pm 3.48$ & $-0.27 \pm 0.13$ & $+4.23 \pm 2.41$ & $+4.26 \pm 2.98$ \\
Cora & $-1.31 \pm 1.31$ & $-0.49 \pm 0.80$ & $-1.05 \pm 1.19$ & $-1.02 \pm 1.98$ \\
Pubmed & $-0.09 \pm 0.58$ & $+0.22 \pm 0.34$ & $-0.21 \pm 0.41$ & $+0.90 \pm 0.93$ \\
Pokec & $-2.54 \pm 4.66$ & $-1.66 \pm 2.21$ & $-0.01 \pm 2.08$ & $-1.09 \pm 1.75$ \\
\bottomrule
\end{tabular}
\end{table}

\begin{table}[t]
\centering
\begingroup\small
\setlength{\tabcolsep}{3pt}
\caption{{Hyperedge prediction: successive changes to the propagation operator (AUROC; zero-message policy for cold nodes). Original-view scores (percent) and paired stepwise changes (percentage points) are mean $\pm$ sample SD. O$_1$ adds pair multiplicities; O$_2$ changes structural self weights and normalization after pair multiplicities have been added by O$_1$. D and L are disabled.}}
\label{tab:operator-edge-pred-zero-secondary}
\begin{tabular}{@{}lrrrrr@{}}
\toprule
Dataset & CEGCN & +O$_1$ & +O$_1$+O$_2$ & $\delta_1$ & $\delta_2$ \\
\midrule
Actor & $65.44 \pm 2.43$ & $63.07 \pm 1.14$ & $65.12 \pm 0.75$ & $-2.37 \pm 1.51$ & $2.05 \pm 1.25$ \\
Cora-CA & $61.17 \pm 6.87$ & $60.03 \pm 6.36$ & $53.74 \pm 1.44$ & $-1.13 \pm 0.81$ & $-6.30 \pm 5.83$ \\
DBLP-CA & $65.13 \pm 3.35$ & $62.03 \pm 2.07$ & $60.45 \pm 1.38$ & $-3.10 \pm 1.96$ & $-1.58 \pm 2.96$ \\
Cora & $59.89 \pm 0.89$ & $59.66 \pm 1.01$ & $58.58 \pm 0.57$ & $-0.23 \pm 0.88$ & $-1.08 \pm 1.58$ \\
Pubmed & $72.26 \pm 1.02$ & $72.86 \pm 1.08$ & $72.17 \pm 0.45$ & $0.61 \pm 0.59$ & $-0.70 \pm 0.77$ \\
Pokec & $75.22 \pm 3.07$ & $74.71 \pm 1.12$ & $72.69 \pm 2.70$ & $-0.51 \pm 3.07$ & $-2.02 \pm 1.79$ \\
\bottomrule
\end{tabular}
\endgroup
\end{table}

\clearpage

\end{document}